\documentclass{article}

\usepackage{arxiv}
\usepackage[utf8]{inputenc} 
\usepackage[T1]{fontenc}    
\usepackage{hyperref}       
\usepackage{url}            
\usepackage{booktabs}       
\usepackage{amsfonts}       
\usepackage{nicefrac}       
\usepackage{tabularx}
\usepackage{graphicx}
\usepackage{doi}
\usepackage{cite}
\usepackage{caption}
\usepackage{subcaption}

\title{Sign Language Recognition Using Original and Synthetic Depth Image Based Point Cloud Data Models}

\newif\ifuniqueAffiliation
\uniqueAffiliationtrue

\ifuniqueAffiliation 
\author{ \href{https://orcid.org/0000-0002-7724-6848}{\includegraphics[scale=0.06]{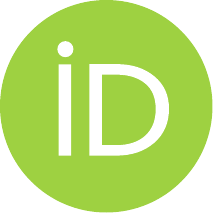}\hspace{1mm}Rüstem Özakar}\\
	Department of Computer Engineering\\
	Erzurum Technical University\\
    Erzurum, Turkey \\
	\texttt{rustem.ozakar@erzurum.edu.tr} \\
	\And
	\href{https://orcid.org/0000-0002-7212-5457}{\includegraphics[scale=0.06]{orcid.pdf}\hspace{1mm}Eyüp Gedikli} \\
	Department of Computer Engineering\\
	Trabzon University\\
	Trabzon, Turkey \\
	\texttt{eyupgedikli@trabzon.edu.tr} \\
}
\fi

\renewcommand{\headeright}{ }
\renewcommand{\undertitle}{ }

\begin{document}
\maketitle

\begin{abstract}
	Research regarding the sign language recognition mostly relies on RGB images, whileas sign language datasets that provide depth images are limited. Point clouds obtained from depth images can be used for sign language recognition with neural networks like PointNet. In recent years, various neural networks are used for generating realistic depth images from monocular RGB images. In this work, synthetic depth images were created from RGB images using Depth Anything V2 network. For this purpose, three sign language datasets (Real-time ASL Fingerspelling, KArSL, AUTSL) which contain both RGB and depth images were used. Classification accuracies of the point cloud data created from both original and synthetic depth images using various PointNet architectures were measured for sign language recognition. From the original and synthetic point clouds, frame based, Point Gesture Map and Long Short Term Memory data models were used for classification and their performances were compared. In the results, both original and synthetic based data achieved acceptable performance in most models. In general, original depth based point cloud models performed better than synthetic ones, however in some models synthetic depth based models performed better than the originals.
\end{abstract}


\section{Introduction}

Sign language is a communication method used by deaf people that includes hand, arm, body movements and facial gestures. Sign language can be used for words and sentences as well as individual letters (fingerspelling). Different nations have their own sign languages. 

Sign Language Recognition (SLR) has been an intensive research topic in computer science for many years. Significant success can be achieved especially with the advent of the advanced vision based artificial neural networks. SLR can be considered as a variant of hand gesture recognition. Thus, all challenges of the hand gesture recognition also applies to the SLR. Different sign languages, variety of hand gestures, gestures being performed by different signers in different environments, real time processing of the gestures in a continous speech, ensuring semantic correctness, dataset scarcity can be listed as some of the challenges of the SLR.

Vision based artificial neural networks is the common method of choice for gesture recognition. With the continous improvements in neural network architectures, they became able to solve more and more complex problems. For gesture recognition, image data, sensor data or radar data can be used. In vision based models, depth images and point cloud data can be used together with the RGB images. Point clouds provide three dimensional data obtained from depth images, which opens possibilites for new techniques. Various operations like classification \cite{zhang2023deep}, segmentation \cite{zhang2019review}, mathematical measurements \cite{chang2017object}, spatio-temporal examination \cite{wang2019space}, gesture recognition \cite{mirsu2020pointnet} can be done using point clouds.

Being the prevalent data type in SLR, RGB data has some disadvantages. It's most significant disadvantage is that it can be affected by light or color changes in the objects/persons. Depth images on the other hand, depending on their sensor type, may not be affected by these at all. Depth images are acquired with the dedicated cameras. These cameras are also called time-of-flight cameras, they use various techniques to map the scene they are viewing in 3d. Because of monocular RGB images being the most common image type in daily life, unavailability of the depth information of the most RGB images and limits of the depth sensors regarding resolution and effective distance, researchers are using artificial neural networks to create depth images from RGB images. Thus, many useful methods regarding depth images become applicable to RGB data.

In this work, SLR success of the synthetic depth images created from RGB images is compared to the original depth images. For this purpose, three different datasets (Real-time ASL Fingerspelling, KArSL, AUTSL) which contain both RGB and depth images were used. Point cloud data models created from both original and synthetic depth images were used to train PointNet \cite{qi2017pointnet} models and their performances were measured. Datasets are selected from the publicly available SLR datasets, two of these datasets contain high amount of data, making them suitable for comparison.


\section{Literature Review}

SLR research can be evaluated in different categories. They can be categorized as isolated, meaning words or expressions are separated individually or, continuous; like the gestures from an actual live speech, unseparated. They can be categorized based on dynamic or static nature of the data. Or, they can be categorized according to the data type, sensor based or vision based. In the literature, RGB data is the prevalent data type for SLR. For detailed summary of the literature regarding the SLR, \cite{cheok2019review}, \cite{koller2020quantitative}, \cite{sarhan2023unraveling}, \cite{tao2024sign}, \cite{al2024advancements}, \cite{khan2025deep} surveys can be examined.

When the recent literature is reviewed, it can be seen that depth images and point clouds are less used than RGB images. A brief summary of some depth based SLR works in the literature is as follows; Wang et al. \cite{wang2016large} developed three different representations of the depth images (Dynamic Depth Image, Dynamic Depth Normal Image, Dynamic Depth Motion Normal Image) using temporal dimension. Recognition was performed using ConvNet classifiers. They worked on Chalern LAP IsoGD dataset. This dataset contains 249 gestures performed by 21 persons. Warchol et al. \cite{warchol2019recognition} worked on point cloud data of the Polish fingerspelling language. To extract features, they used viewpoint feature histogram, eigenvalues-based, ensemble of shape functions and global radius based surface descriptors. For classification, Hidden Markov Models (HMM) were used. In the gestures representing 16 letters, arm region is separated using depth values, then hand region is obtained from this region. They used Kinect V2 camera. Aly et al. \cite{aly2019user} worked on American sign language dataset for fingerspelling. To determine the hand regions, they used the depth value. After this step, they detected the wrist region to obtain hands. For feature extraction, PCANet was used. Support Vector Machine (SVM) was used for classification. Adaloglou et al. \cite{adaloglou2021comprehensive} examined computer vision and deep learning methods for SLR. A Greek sign language dataset with RGB-D images containing isolated and continous signs was created. On this dataset, various Convolutional Neural Network (CNN) architectures were tested including spatio-temporal models. Depth images were also tested for their performance. Oszust and Krupski \cite{oszust2021isolated}, divided the depth images to cells and extracted feature vectors from them. They used Dynamic Time Warping for various data length. Most relevant cells were identified and gestures were classified using Nearest Neighbour. PSL 3D dataset was created using Kinect. Sarhan et al. \cite{sarhan2023pseudodepth}, similar to this research, created synthetic depth images with Dense Prediction Transformer. RGB and depth images were used with 3DCNN architecture for SLR. They used ChaLearn 249 IsoGD dataset which was created using Kinect.

\section{Methodology}

In this work, to be able to compare the perfomance of the point clouds created from the original and synthetic depth images, SLR datasets containing both RGB and depth data were used. Original depth images can already be used to create point clouds. With the advent of generative neural networks, it has become possible to create depth images from RGB images. These depth images can also be used for creating point clouds. Point cloud data represents the 3d world position of the each pixel in the depth image, they can be calculated given the instrinsic camera matrix and depth value. These unordered points can be used with a neural network like PointNet  \cite{qi2017pointnet}. Point clouds can be created using libraries like Open3d \cite{Zhou2018}. In this research, Open3d is used for point cloud creation.

\begin{figure}[htb]
\centering
\includegraphics[width=\textwidth]{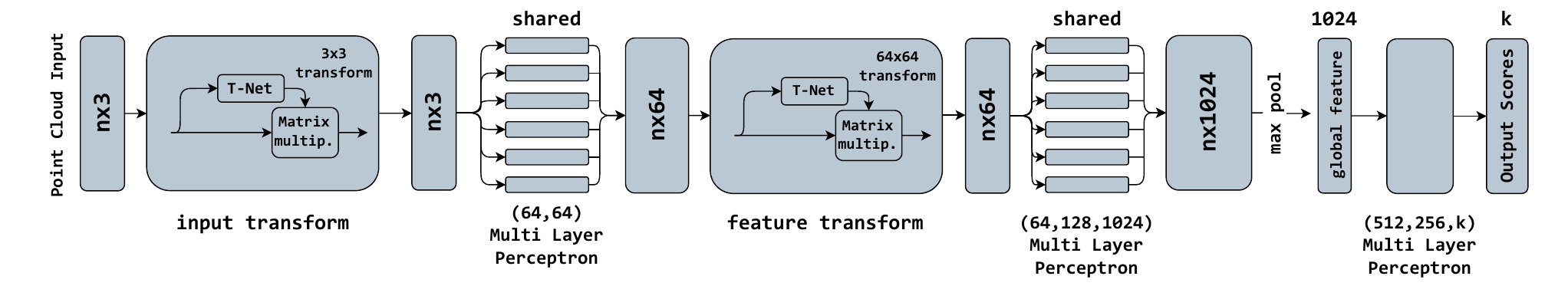}
\caption{\label{pointnet-architecture} PointNet architecture. \cite{qi2017pointnet}.}
\end{figure}

PointNet is a specially designed neural network for point clouds, which can be used for tasks like object classification, gesture recognition using spatio-temporal information, segmentation and more. In it's architecture, T-net modules learns a transformation matrix which ensures transform invariance of the points. Multi-Layer Perceptron (MLP) and max pooling layers ensures performing on unordered points. By extracting features from the point clouds, classification and segmentation is performed. PointNet architecture can be seen in the Figure \ref{pointnet-architecture}. In this work, PointNets are used as a classifier for different point cloud data models. Like image frame based CNNs, PointNets also work with frame based point cloud data. Also, as in the work of \cite{ozakar2023evaluation}, spatio-temporal recognition using the merged point clouds of a gesture similar to 3DCNN is possible. Because of SLR gestures having a spatio-temporal domain, LSTM \cite{hochreiter1997long} networks can be used as a classifier with the aggregated frames belonging to a gesture. In this research, LSTM network is used with the features extracted from a fixed amount of point clouds of the gestures.

Various networks can be used for creating depth images from monocular RGB images. Among the architectures, there are models based on Stable Diffusion \cite{rombach2022high}, as well as discriminative models like Depth Anything V2 \cite{yang2024depth}. In Depth Anything V2 model, a DINO-v2 transformer model with a discriminative generative neural network is trained using synthetic images. This network produced depth images using a large amount of real world data. With these depth images, final models of the Depth Anything V2 model is trained. In this work, Depth Anything V2 model is used for creating synthetic depth images from three SLR datasets. 

\subsection{Datasets}

\begin{figure}[!htb]
\centering
\includegraphics[height=75px]{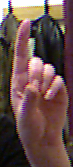}
\includegraphics[height=75px]{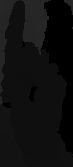}
\includegraphics[height=75px]{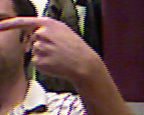}
\includegraphics[height=75px]{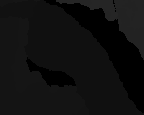}
\includegraphics[height=75px]{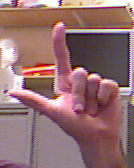}
\includegraphics[height=75px]{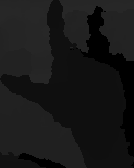}
\caption{\label{datasetA-examples} Example data from the Dataset-A \cite{pugeault2011spelling}.}
\end{figure}

\underline{\textbf{Dataset-A (Real-time ASL Fingerspelling):}} Dataset-A is a dataset of American fingerspelling gestures \cite{pugeault2011spelling}. It contains RGB and depth frame data which was recorded with Microsoft Kinect V1 camera. 24 gestures for the 24 letters of the alphabet were recorded by various signers. It is a static image library with no spatiotemporal dimension. Authors developed a classification system using Gabor filters for feature extraction and Random Forest for classification. Example frames from the dataset can be seen in Figure \ref{datasetA-examples}.

\begin{figure}[!htb]
\centering
\includegraphics[height=75px]{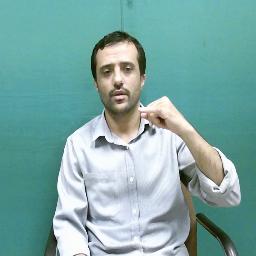}
\includegraphics[height=75px]{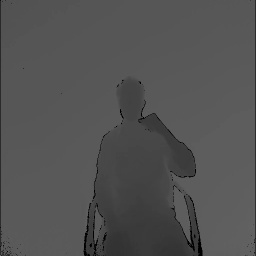}
\includegraphics[height=75px]{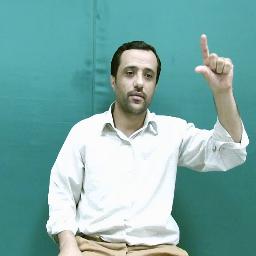}
\includegraphics[height=75px]{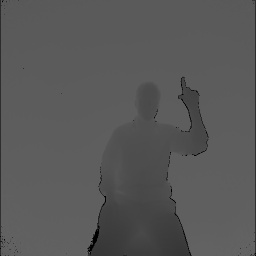}
\caption{\label{datasetB-examples} Example data from the Dataset-B \cite{sidig2021karsl}.}
\end{figure}

\underline{\textbf{Dataset-B (KArSL):}} Dataset-B is a dataset of gestures for the Arabic Sign Language \cite{sidig2021karsl}. It was recorded using Microsoft Kinect V2 camera and contains RGB, depth and skeleton data. Dataset was recorded by three signers, performing 502 different gestures with a green screen background. Each gesture was repeated for 50 times. Dataset contains 75300 samples of gestures in video frame format, having spatiotemporal dimension. Authors developed classification systems using Histogram of Oriented Gradients (HOG) features, HMM, CNN and attention based deep learning classifiers. Classification accuracies varies between \%55 to \%100 for different models/scenarios. Example frames from the dataset can be seen in Figure \ref{datasetB-examples}.

\begin{figure}[!htb]
\centering
\includegraphics[height=75px]{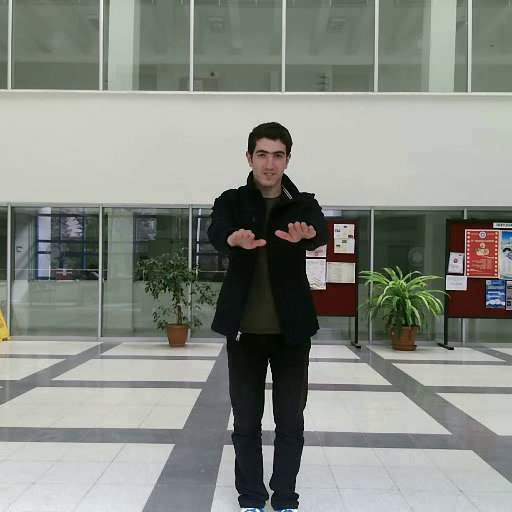}
\includegraphics[height=75px]{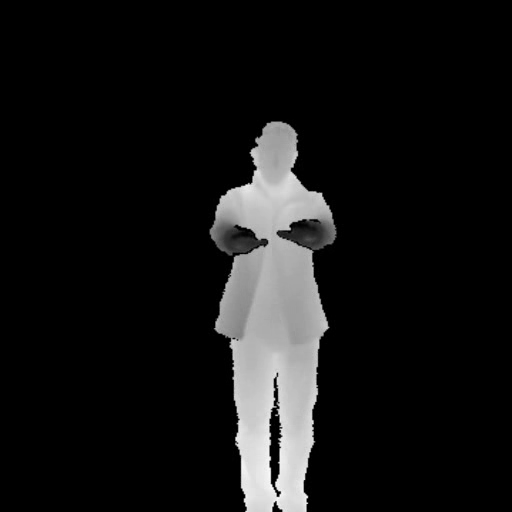}
\includegraphics[height=75px]{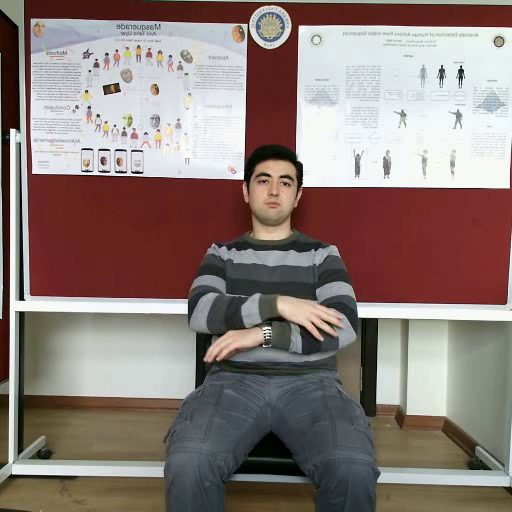}
\includegraphics[height=75px]{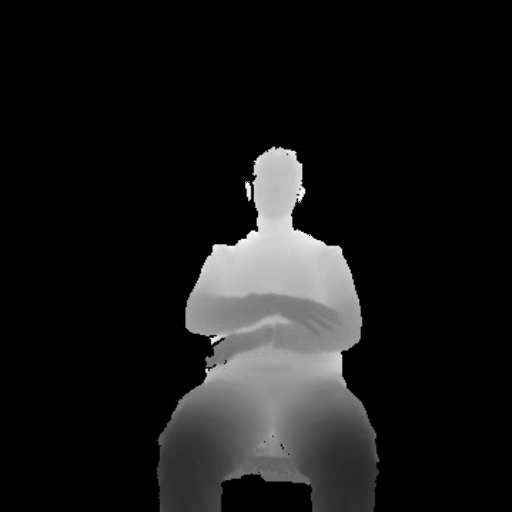}
	\caption{\label{datasetC-examples} Example data from the Dataset-C \cite{sincan2020AUTSL}.}
\end{figure}

\underline{\textbf{Dataset-C (AUTSL):}} Dataset-C is a dataset of gestures for the Turkish Sign Language \cite{sincan2020AUTSL}. It is recorded with the Microsoft Kinect V2 camera and contains RGB, depth and skeleton data. Dataset consists of 226 gestures, performed by 43 signers in different backgrounds. In total, it contains 38336 videos. Authors used CNN, FPM, LSTM and attention based classifiers with different combinations of RGB and depth data for classification. Their models acquired various accuracies ranging from \%37.84 to \%83.93 for depth and \%22.80 to \%75.78 for RGB data. Example frames from the dataset can be seen in Figure \ref{datasetC-examples}. Overall general information about the three datasets is given in Table \ref{table_dataset_info1}.

\begin{table}[htb!]
	\centering
	\caption{General information about datasets.}
	\begin{tabular}{lllll}
		\hline
		\textbf{Dataset} & \textbf{Gesture Amount} & \textbf{Signer Amount} & \textbf{Data Type} & \textbf{Data Amount}\\ 
		\hline
		Real-time ASL Fingerspelling (Dataset-A) & 24 & 5 & Frame & 65890 \\
		\hline
		KArSL (Dataset-B) & 502 & 3 & Video & 75300 \\
		\hline
		AUTSL (Dataset-C) & 226 & 43 & Video & 38336 \\
		\hline
	\end{tabular}
	\label{table_dataset_info1}
\end{table}

\subsection{Data Preparation}

\underline{\textbf{Dataset-A (Real-time ASL Fingerspelling):}} Data from Dataset-A is inside the folders named from a to y, representing each gesture. Inside these folders, RGB and depth images of the each gesture is located. All this folder structure exists inside folders named A, B, C, D, E representing signers. Synthetic depth images were created in a separate location using the same folder structure.

When creating point clouds using Open3d, point clouds were examined and ensured to have an accurate distribution in 3d space. For this reason, following parameters in Open3d were selected for the original depth images; width 110, height 80, ppx 156, ppy 120, fx 1256, fy 960. For the synthetic depth images, width 110, height 80, ppx 157, ppy 120, fx 314, fy 240 parameters were selected. 

Dataset contains 65890 original depth frames, thus the same amount of point clouds. \%25 of these, 16473 frames were reserved for testing. For the synthetic depth, 65774 frames were created. \%25 of these frames (16444) were reserved for testing. Remaining frames were used with five-fold cross validation, 1/5 being validation and 4/5 training. In frame based models, 512 points were sampled from the raw point clouds to be used in PointNet.

In this dataset, gestures do not contain a spatio-temporal dimension. Because of this reason, LSTM networks were not used with this dataset. However, Point Gesture Maps (PGM) were used like in the work of \cite{ozakar2023evaluation}. PGMs are constructed from every point cloud data of a gesture, stacked in an axis with fixed intervals. When the amount of frames of gestures were examined, about 500 frames were observed per gesture. From both original and synthetic point clouds, 49 frames were merged to create PGMs of each to their own. These PGMs were sampled with 6400 points to be used in PointNet.

PGM data of the original depth point clouds consists of 1274 samples in total. \%25 of these (319) were reserved for testing. PGM data of the synthetic depth point clouds consists of 1272 samples in total. \%25 of these (318) were reserved for testing. Remaining samples were used with five-fold cross validation, 1/5 being validation and 4/5 training. During synthetic depth generation and point cloud creation process in general, a frame was skipped if any issue occured. For this reason, different data amounts may be seen in the same modalities.

\begin{figure}[htb!]
	\centering
	\begin{subfigure}{0.1\textwidth}
		\includegraphics[height=60px, width=50px]{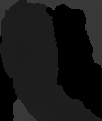}
		\subcaption{}
	\end{subfigure}
	\begin{subfigure}{0.1\textwidth}
		\includegraphics[height=60px, width=50px]{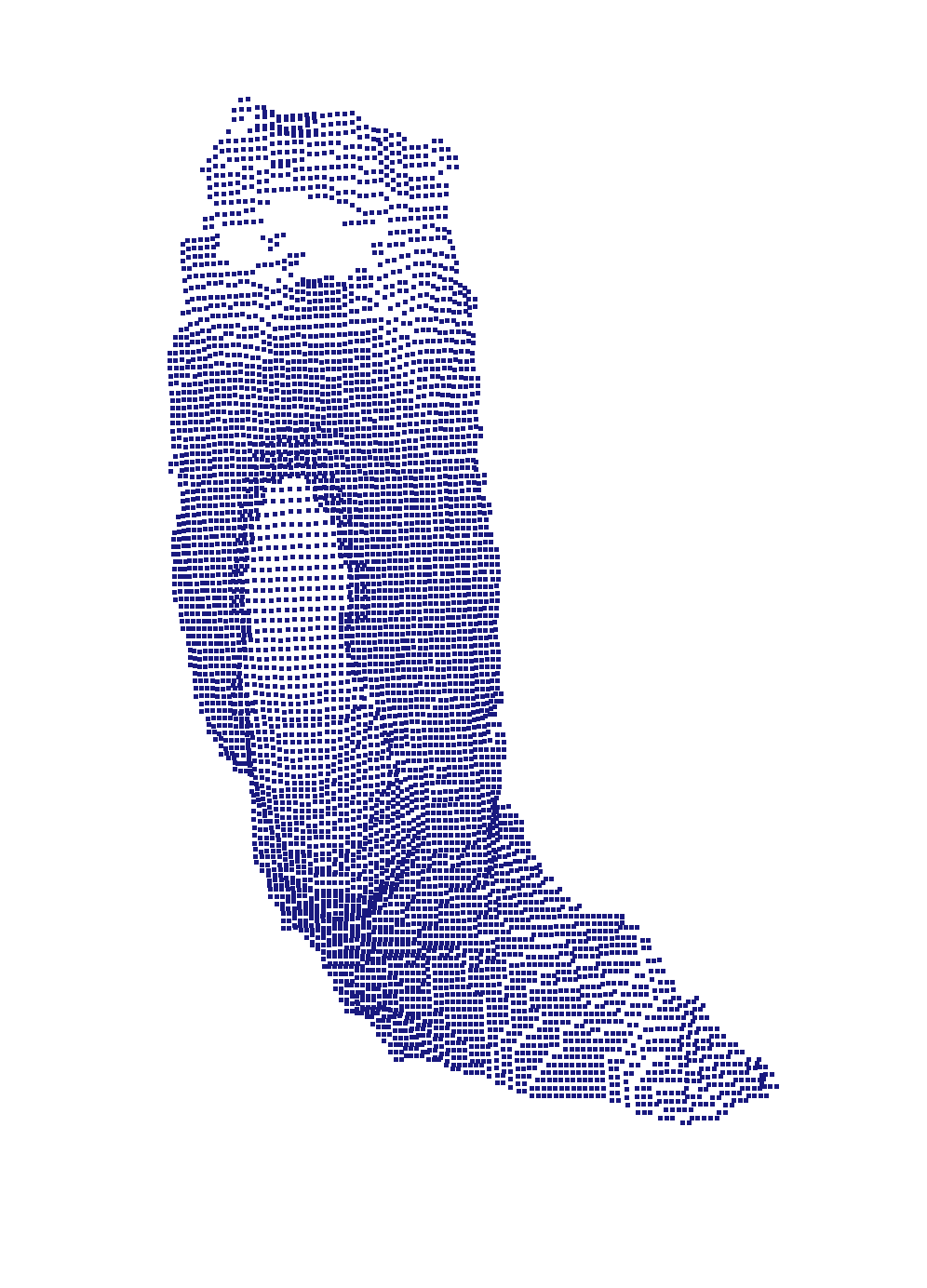}
		\subcaption{}
	\end{subfigure}
	\begin{subfigure}{0.1\textwidth}
		\includegraphics[height=60px, width=50px]{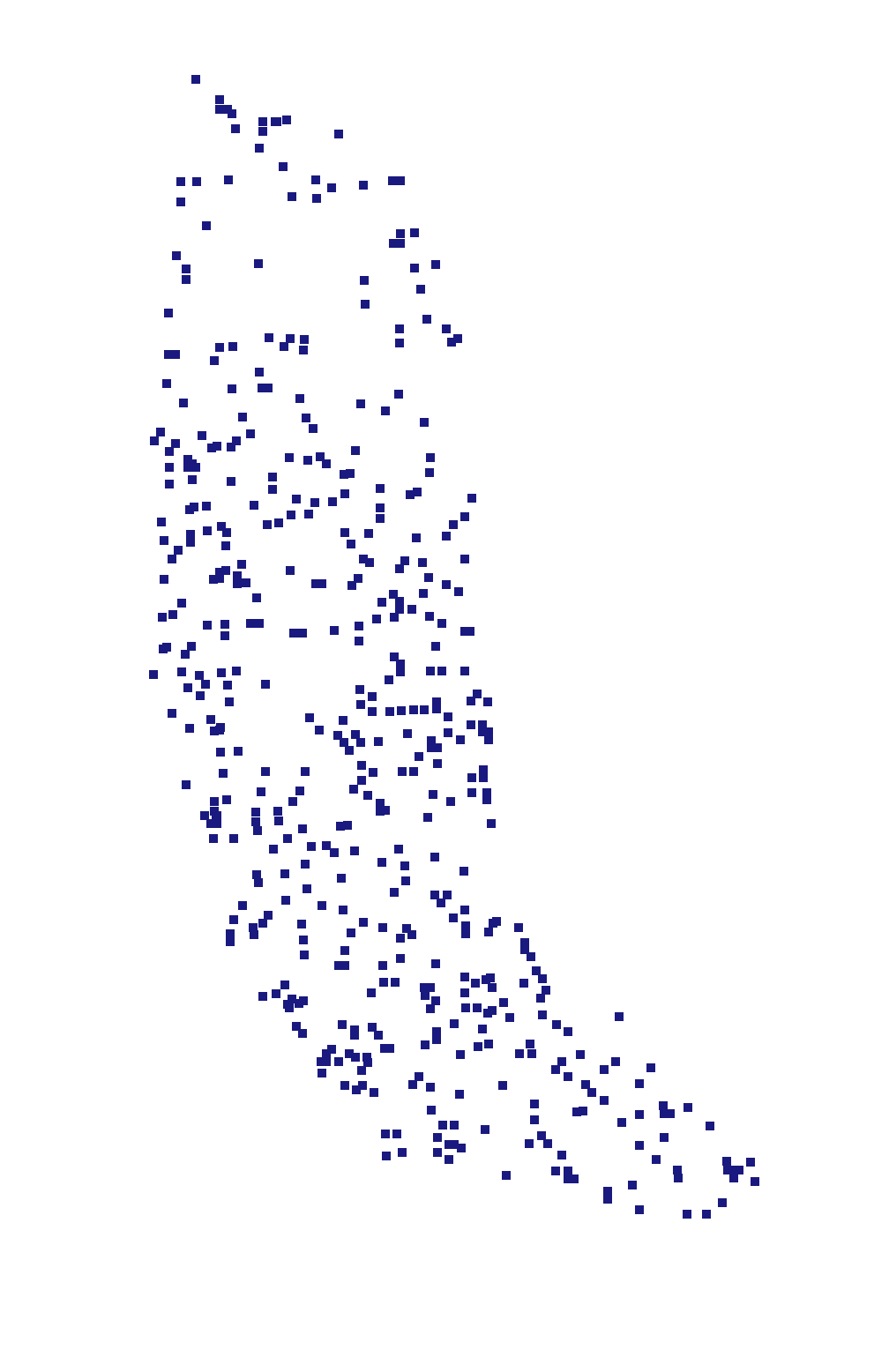}
		\subcaption{}
	\end{subfigure}
	\begin{subfigure}{0.1\textwidth}
		\includegraphics[height=60px, width=50px]{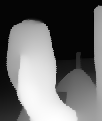}
		\subcaption{}
	\end{subfigure}
	\begin{subfigure}{0.1\textwidth}
		\includegraphics[height=60px, width=50px]{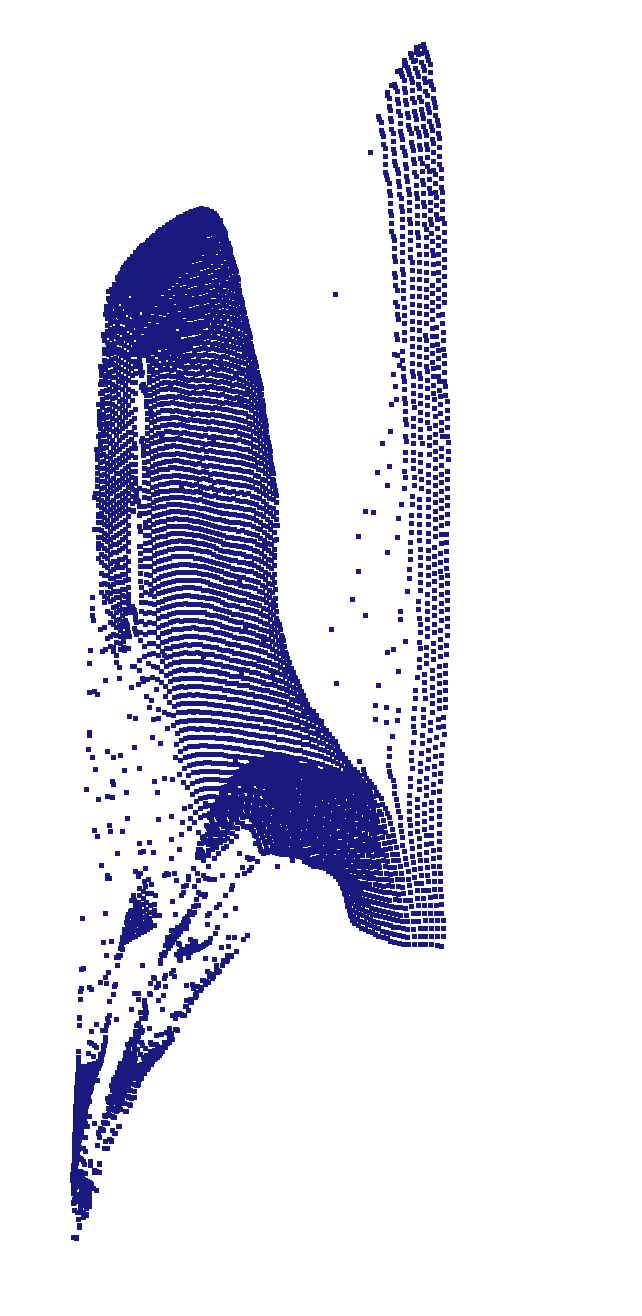}
		\subcaption{}
	\end{subfigure}
	\begin{subfigure}{0.1\textwidth}
		\includegraphics[height=60px, width=50px]{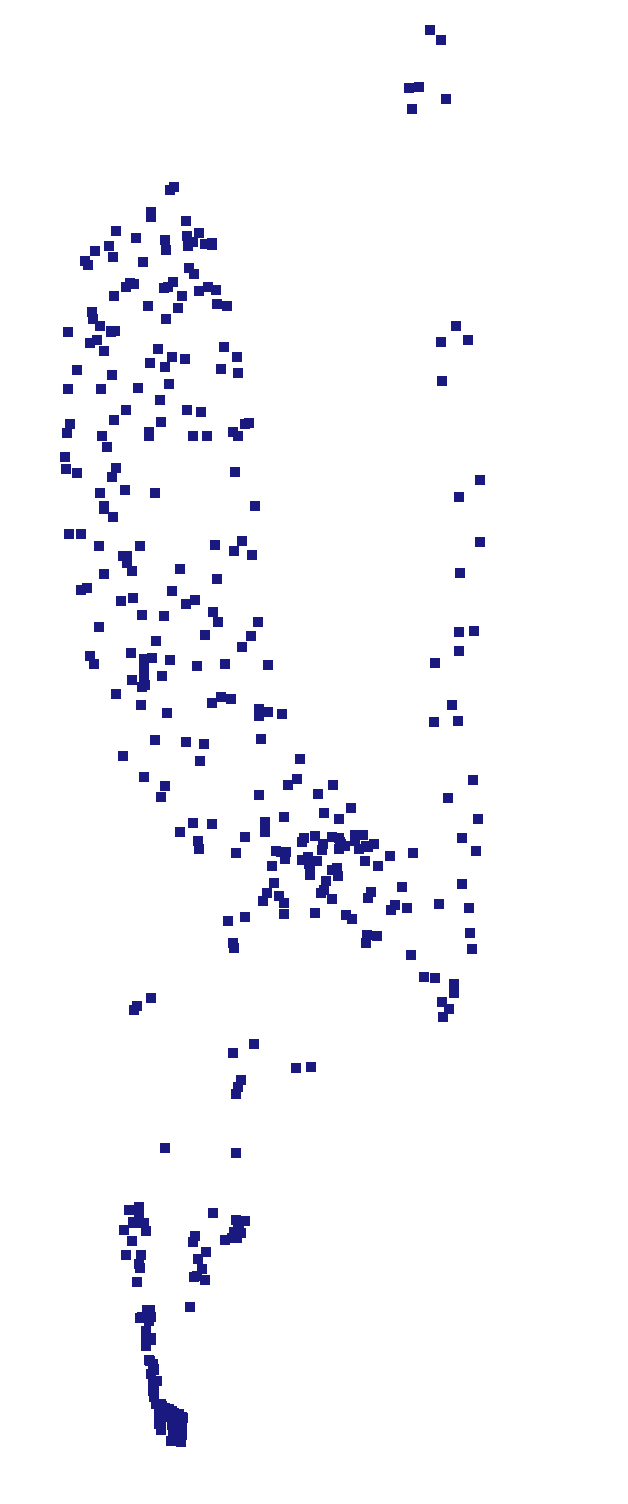}
		\subcaption{}
	\end{subfigure}
	\\
	\begin{subfigure}{0.1\textwidth}
		\includegraphics[height=60px, width=50px]{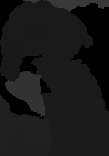}
		\subcaption{}
	\end{subfigure}
	\begin{subfigure}{0.1\textwidth}
		\includegraphics[height=60px, width=50px]{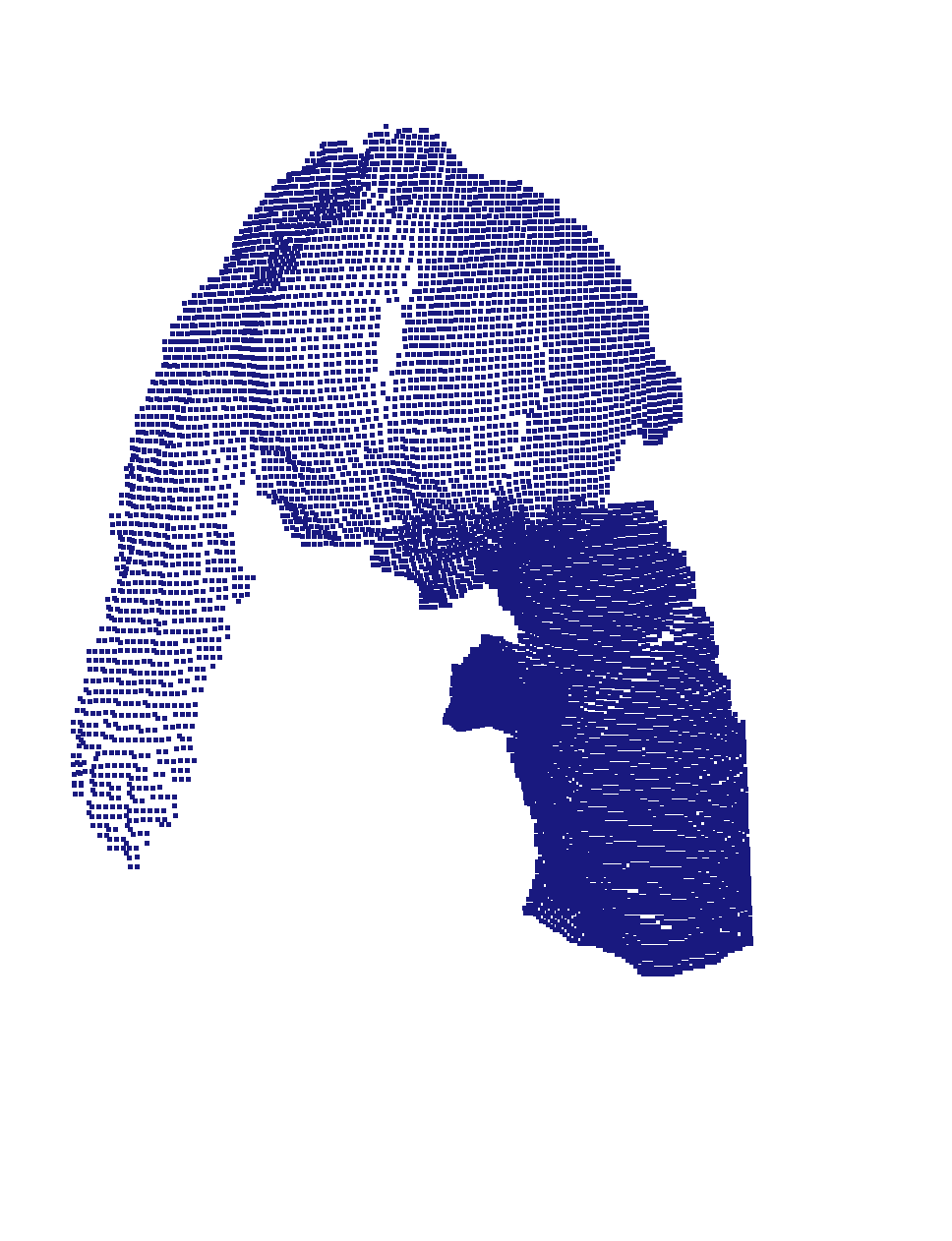}
		\subcaption{}
	\end{subfigure}
	\begin{subfigure}{0.1\textwidth}
		\includegraphics[height=60px, width=50px]{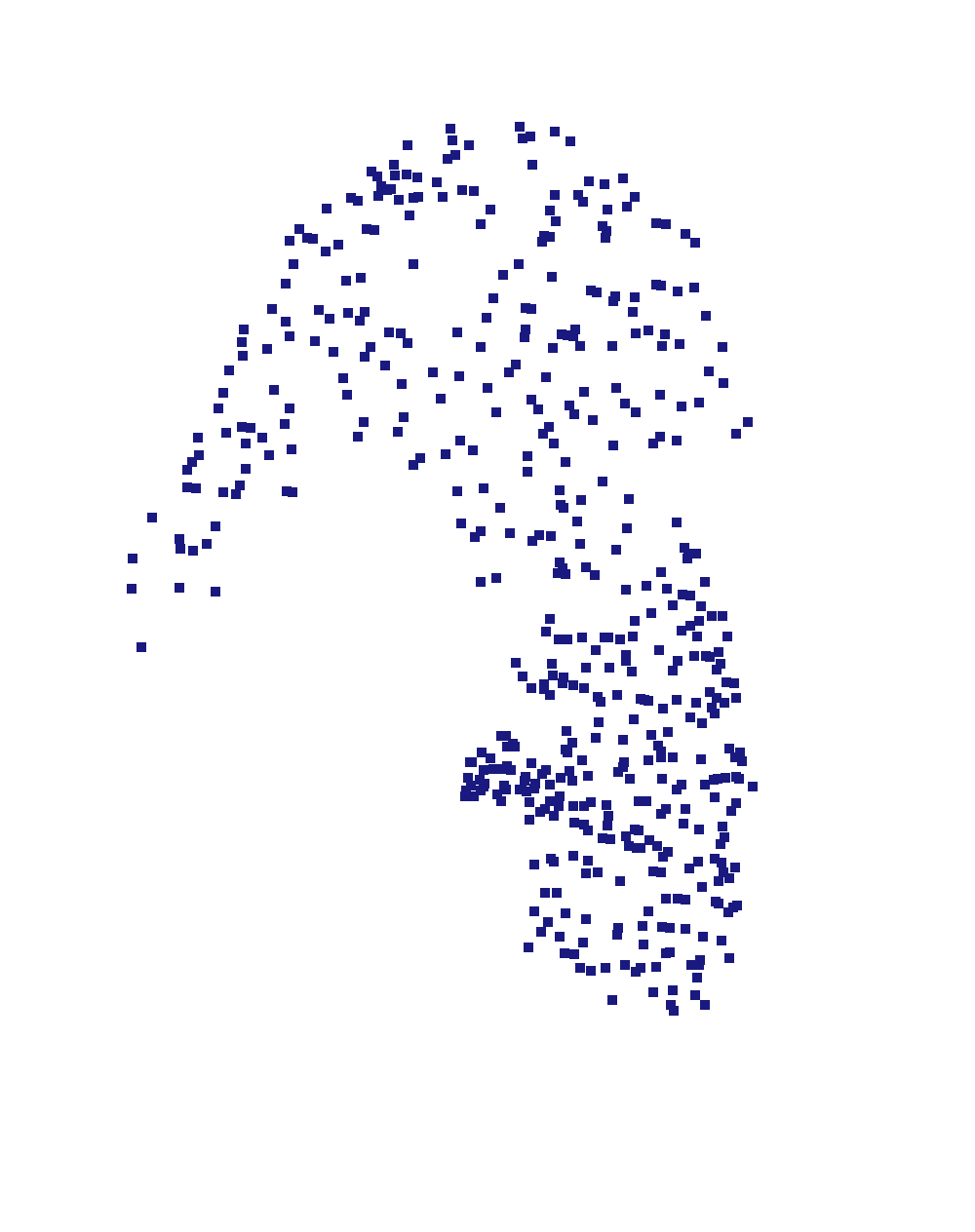}
		\subcaption{}
	\end{subfigure}
	\begin{subfigure}{0.1\textwidth}
		\includegraphics[height=60px, width=50px]{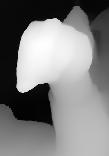}
		\subcaption{}
	\end{subfigure}
	\begin{subfigure}{0.1\textwidth}
		\includegraphics[height=60px, width=50px]{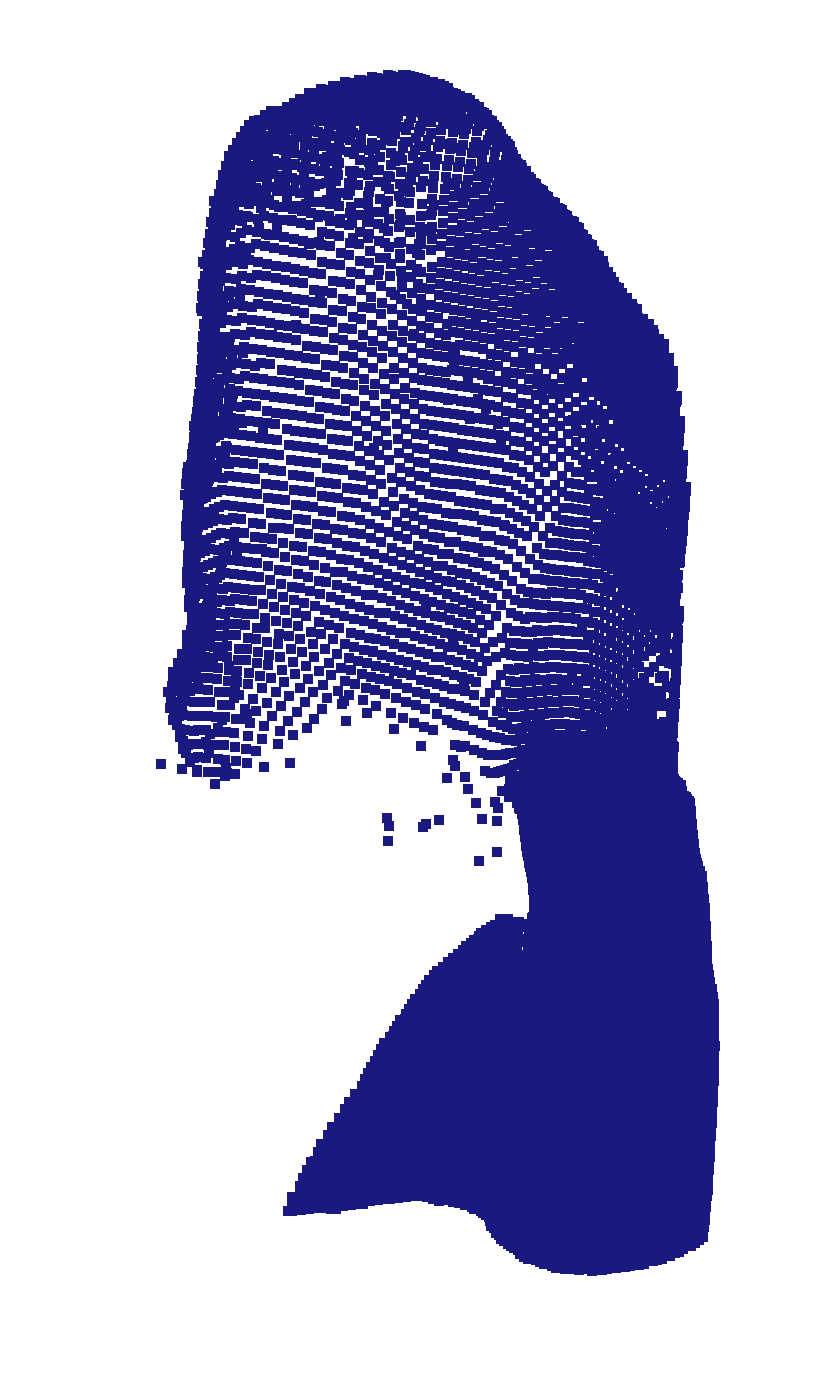}
		\subcaption{}
	\end{subfigure}
	\begin{subfigure}{0.1\textwidth}
		\includegraphics[height=60px, width=50px]{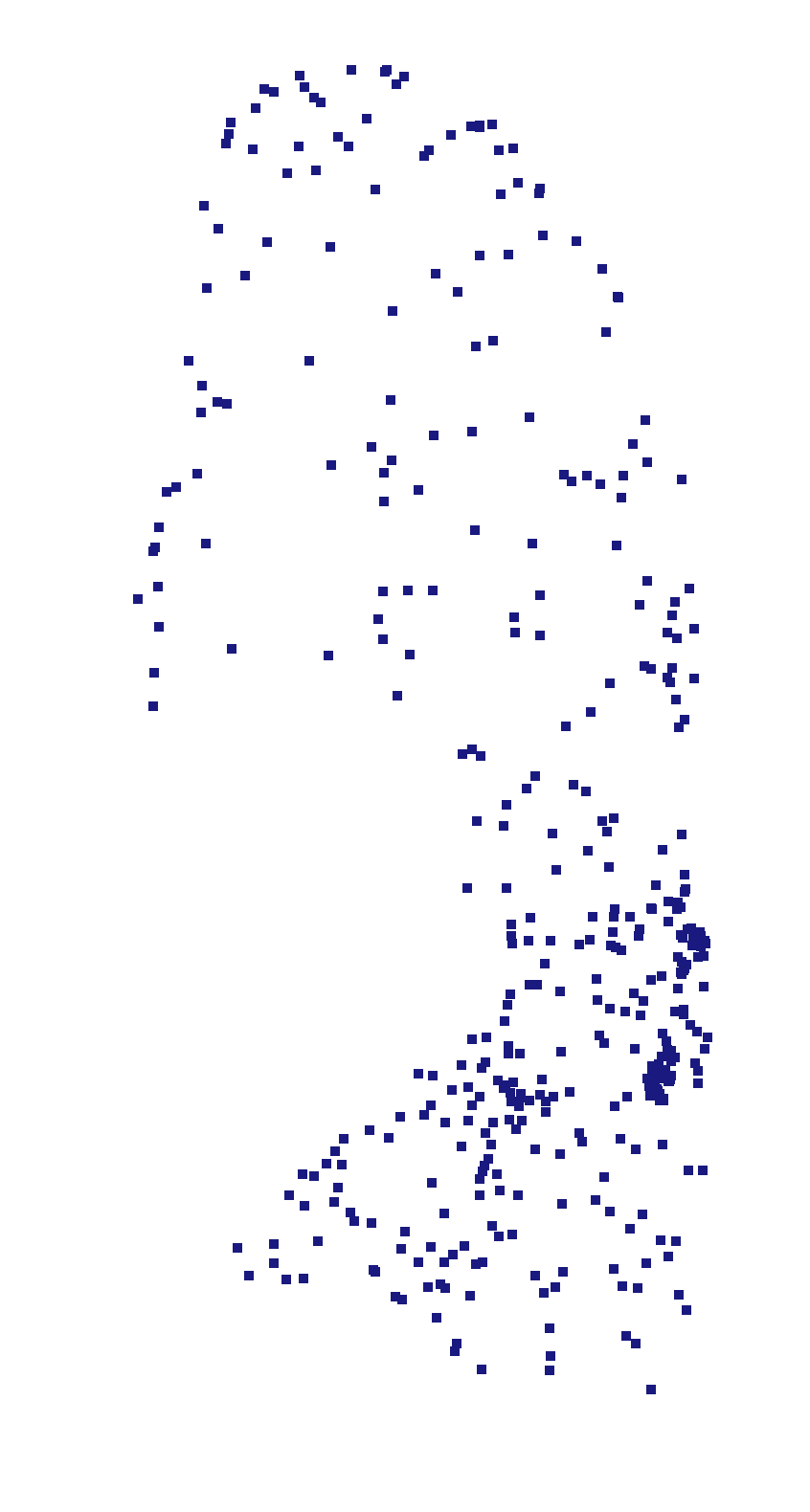}
		\subcaption{}
	\end{subfigure}
	\\
	\begin{subfigure}{0.1\textwidth}
		\includegraphics[height=60px, width=50px]{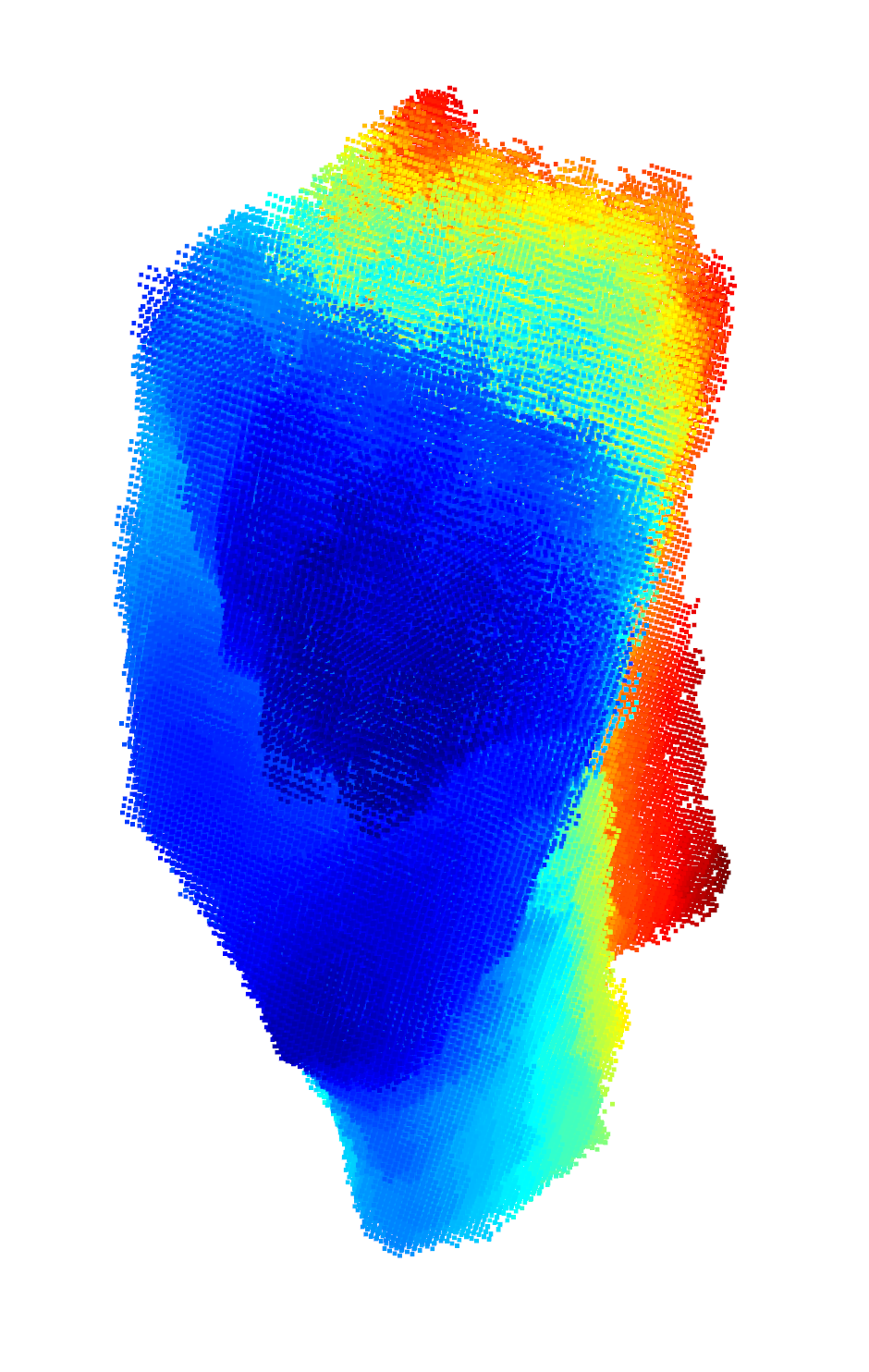}
		\subcaption{}
	\end{subfigure}
	\begin{subfigure}{0.1\textwidth}
		\includegraphics[height=60px, width=50px]{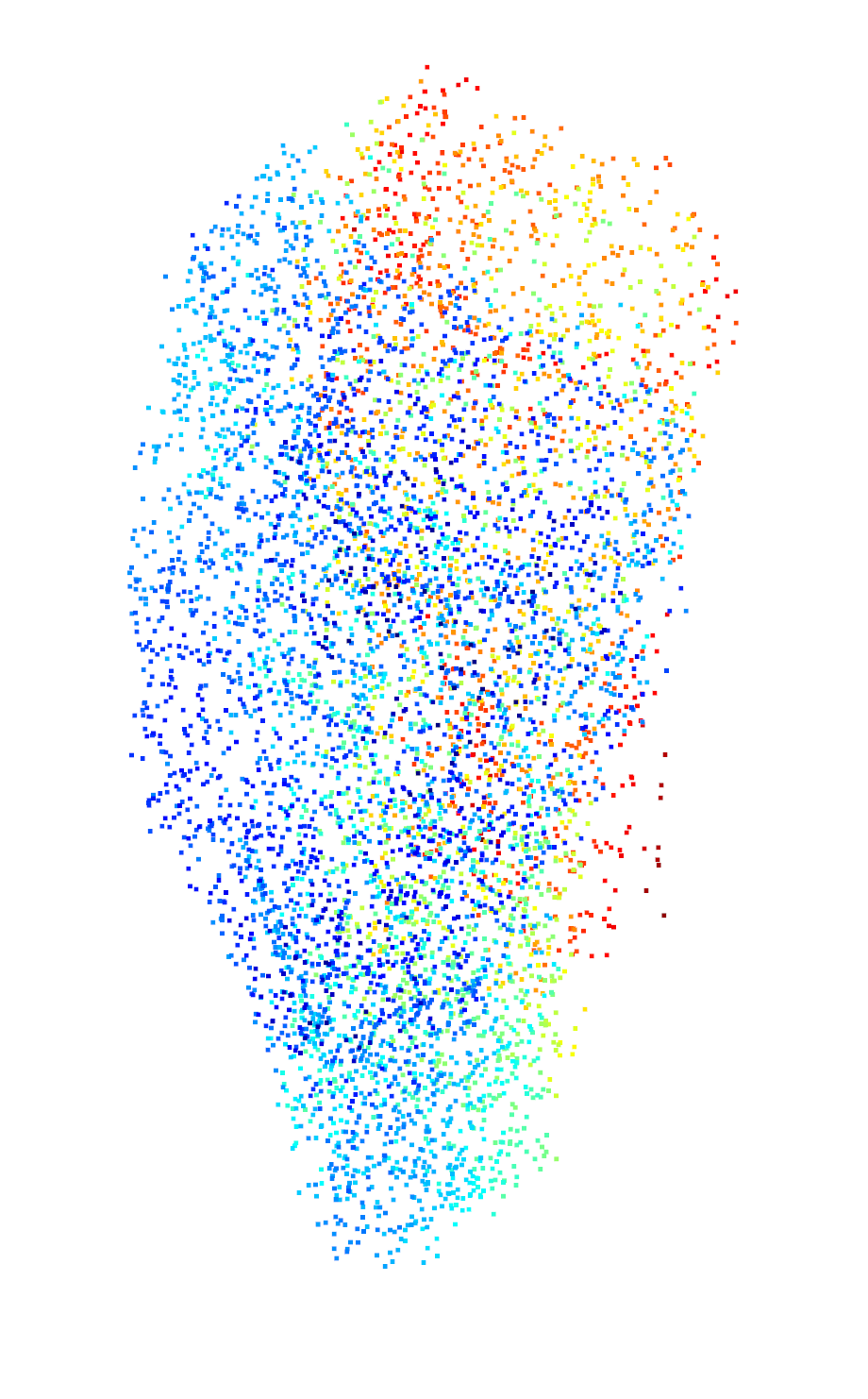}
		\subcaption{}
	\end{subfigure}
	\begin{subfigure}{0.1\textwidth}
		\includegraphics[height=60px, width=50px]{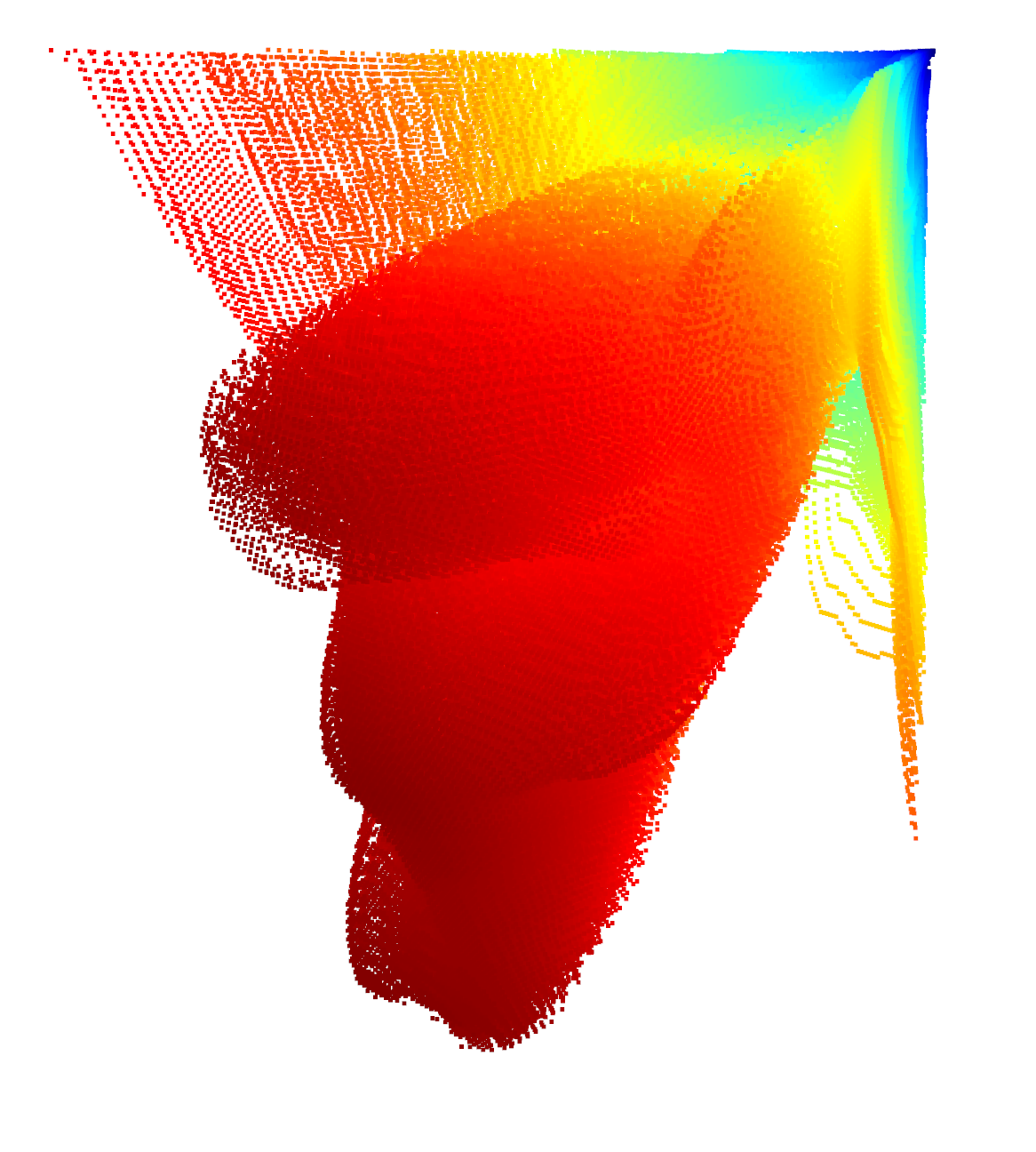}
		\subcaption{}
	\end{subfigure}
	\begin{subfigure}{0.1\textwidth}
		\includegraphics[height=60px, width=50px]{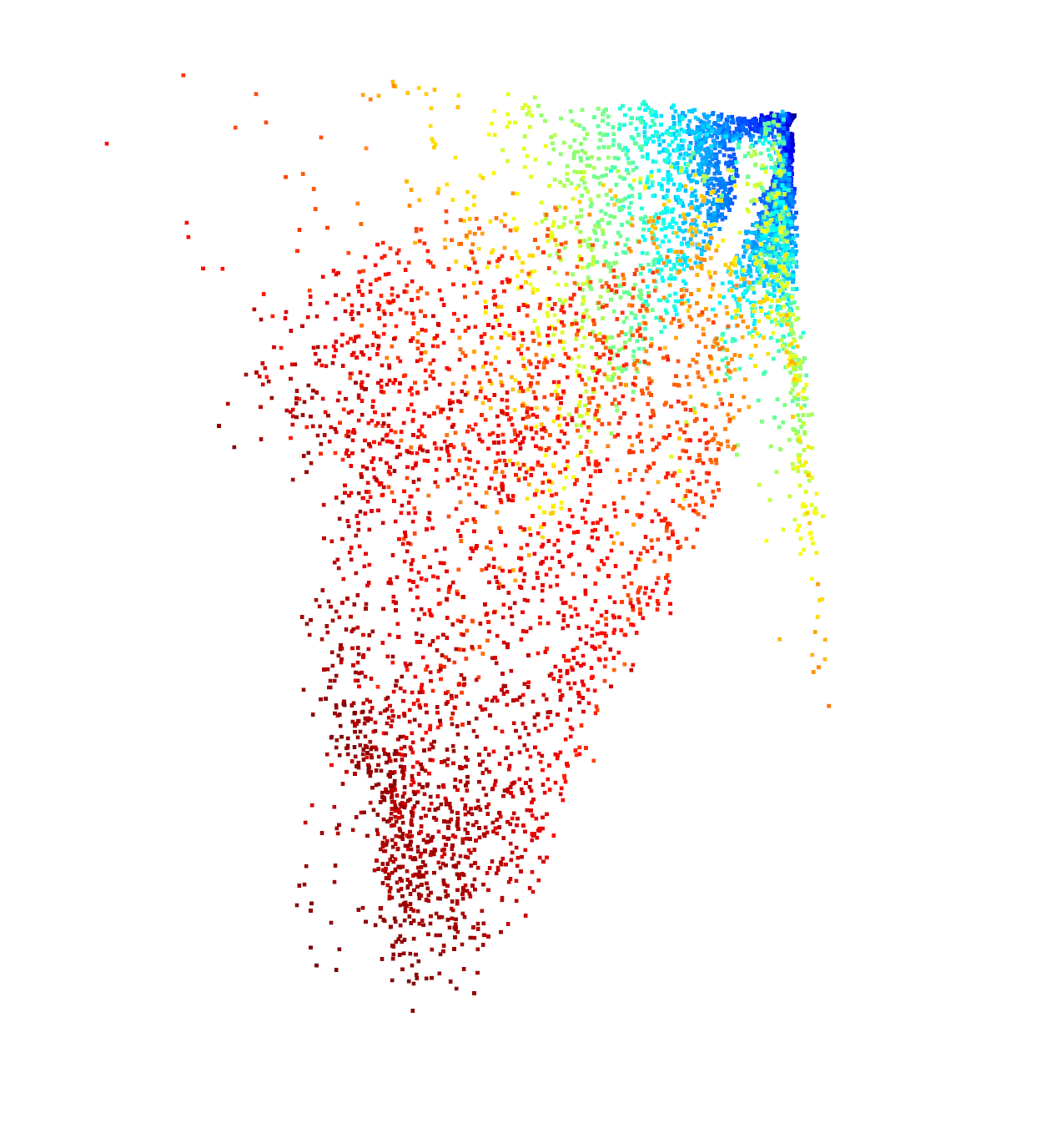}
		\subcaption{}
	\end{subfigure}
	\\
	\begin{subfigure}{0.1\textwidth}
		\includegraphics[height=60px, width=50px]{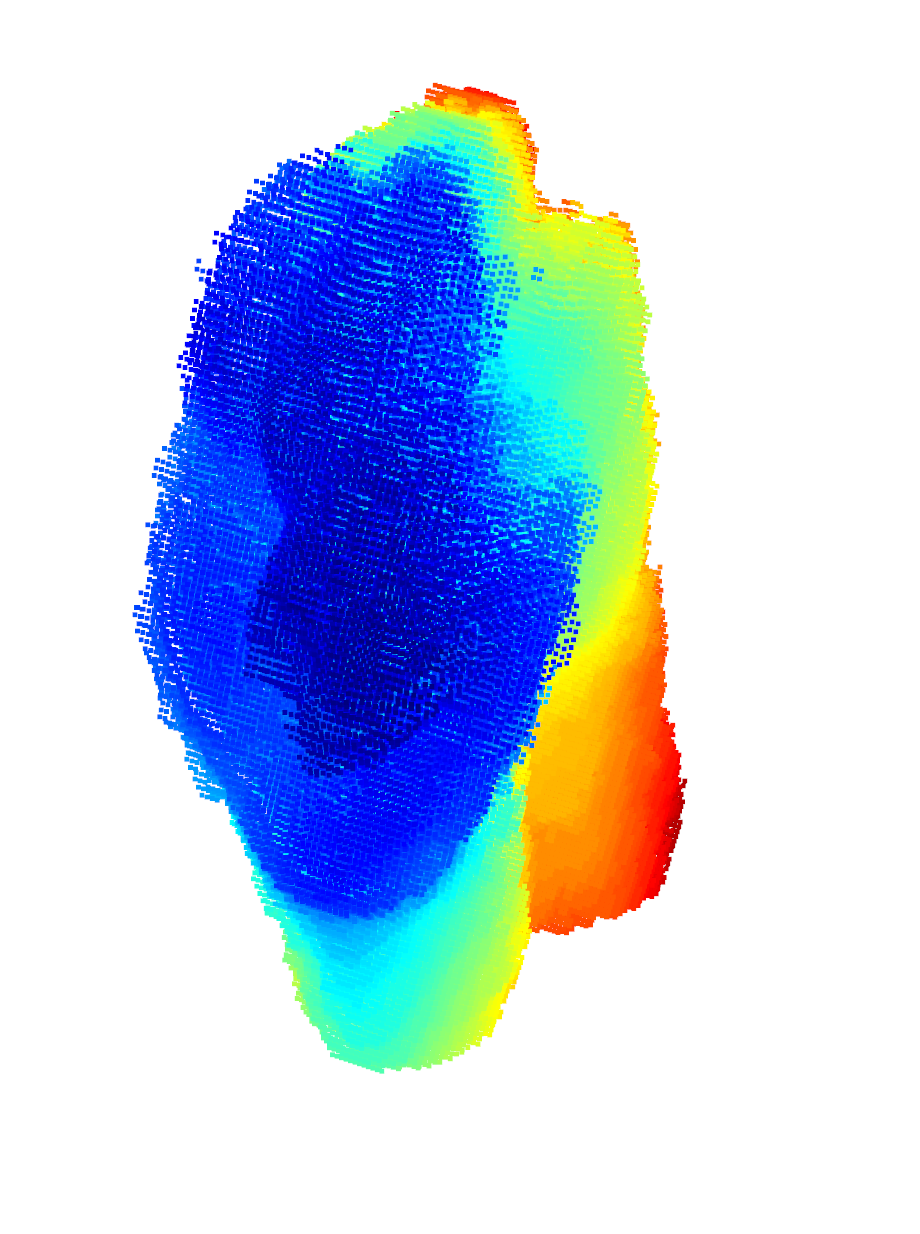}
		\subcaption{}
	\end{subfigure}
	\begin{subfigure}{0.1\textwidth}
		\includegraphics[height=60px, width=50px]{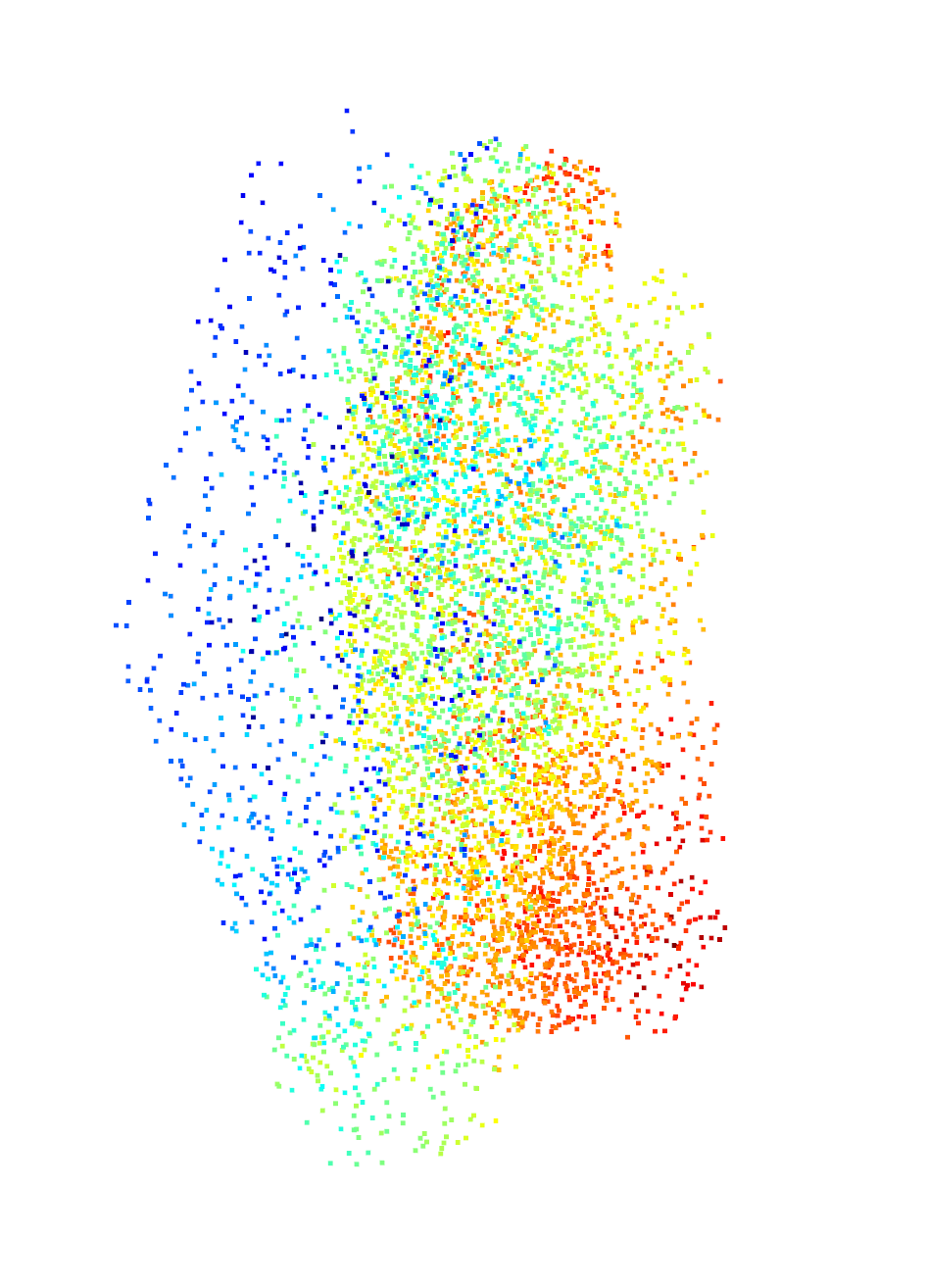}
		\subcaption{}
	\end{subfigure}
	\begin{subfigure}{0.1\textwidth}
		\includegraphics[height=60px, width=50px]{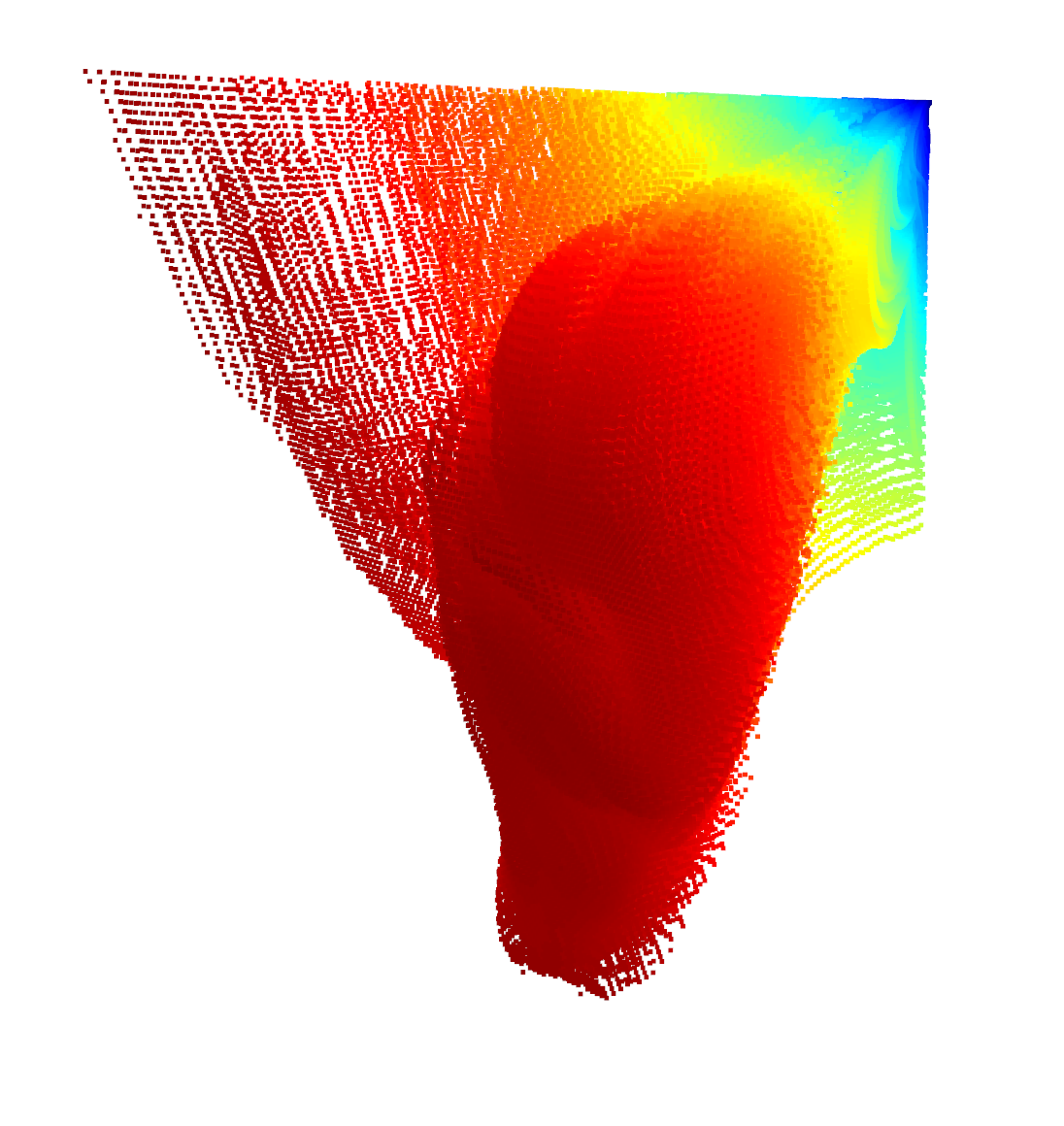}
		\subcaption{}
	\end{subfigure}
	\begin{subfigure}{0.1\textwidth}
		\includegraphics[height=60px, width=50px]{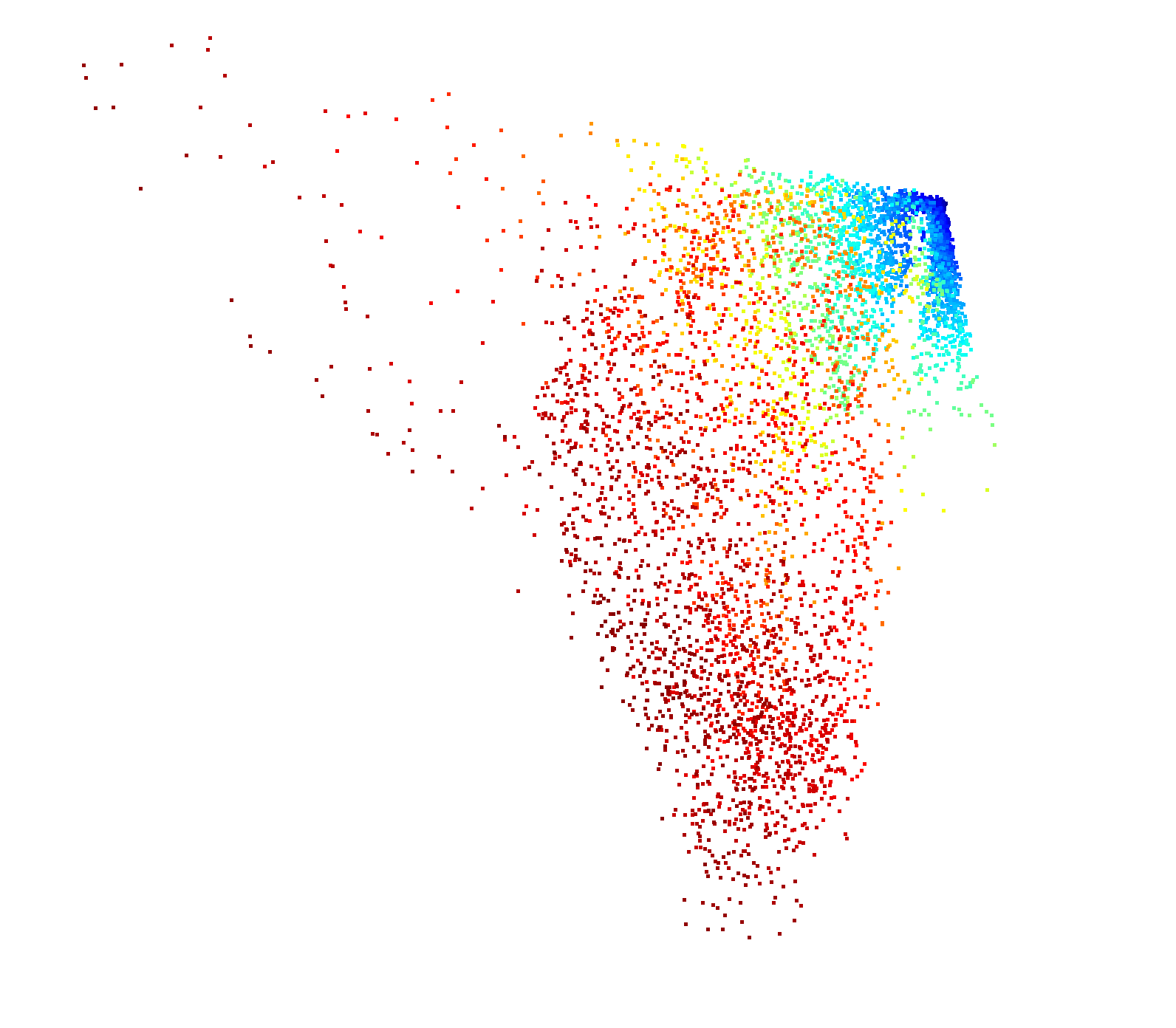}
		\subcaption{}
	\end{subfigure}
	\caption{\label{datasetA-examples2} Examples from the data created from Dataset-A. Top two rows: (a, g) Original depth, (b, h) Raw point cloud, (c, i) 512 sampled point cloud, (d, j) Synthetic depth, (e, k) Raw point cloud (f, l) 512 sampled point cloud. Bottom two row (m, q) Original depth PGM, (n, r) 6400 sampled PGM, (o, s) Synthetic PGM, (p, t) 6400 sampled PGM.}
\end{figure}

For all samples, depth-scale parameter was selected 1.0, depth-trunc was selected 1000 in Open3d. Figure \ref{datasetA-examples2} shows example data created from this dataset.

\underline{\textbf{Dataset-B (KArSL):}} In Arabic sign language dataset, folders start with 01, 02, 03 structure. Each folder also separates as train and test. In these folders, data of each gesture (001 to 502) exists as separate folders. Inside these folders, RGB and depth frames of the videos are stored for different repetitions. Original folder structure of the dataset was used for creating the synthetic depth images.

In Open3d, intrinsic parameters of the Kinect V2 camera was used for creating point clouds. These parameters were as following; width 512, height 424, ppx 220.166, ppy 205.197, fx 367.535, fy 367.353.

From the original depth point clouds in the dataset, \%75 of the frames (1443426 frames) were used for training. Remaining \%25 (481143 frames) were reserved for testing. \%75 of the synthetic depth point clouds (1443489 frames) were used for training. Remaining \%25  (481164 frames) were reserved for testing. Point clouds were sampled using 512 points. Both original and synthetic data was used with five-fold cross validation, using 4/5 for training and 1/5 for validation.

For the original PGM data, 56631 samples which is \%75 of the total samples were used for training. \%25 remaining (18878 samples) were reserved for testing. For the synthetic PGM data, \%75 of the total samples (56636) were used for training. 18879 samples were reserved for testing. Both original and synthetic data required ten-fold cross validation, due to large memory requirements. When creating PGM data, all frames were merged in the subfolder of a gesture to create a sample for the related class. Then, 6400 points were sampled from this raw PGM data.

\begin{figure}[!htb]
	\centering
	\includegraphics[width=0.7\textwidth]{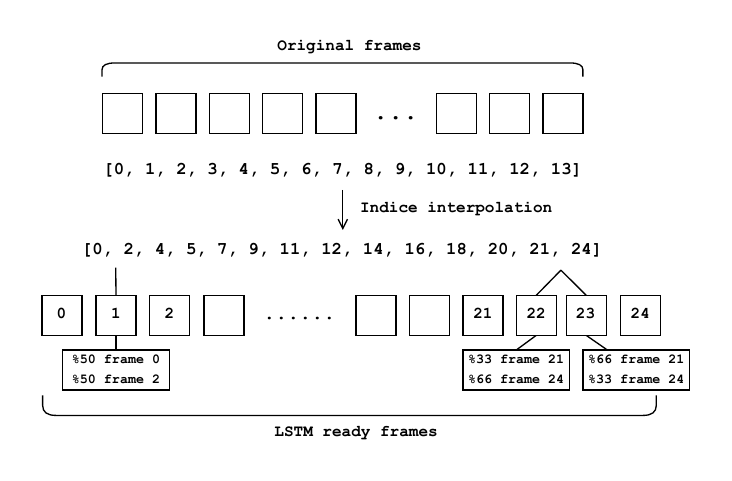}
	\caption{Frame interpolation for the LSTM.}
	\label{lstm-frames}
\end{figure}

Since the gestures of the Arabic dataset contains spatio-temporal information, data for LSTM networks were also created. To have a fixed amount of frames for each gesture, dataset was observed for the average frame amount. 25 frames were decided to be suitable for LSTM networks. In samples where the frame amount is greater than 25, ordered random 25 frames were selected. For the samples having frame amounts between 14 and 25, a special method was followed; indices of the normal frames were mapped to the range 0 - 25. Missing frames were generated by blending the previous and after frames proportionally. In Figure \ref{lstm-frames}, the method of frame generation can be seen.

LSTM data was created using the pretrained frame based PointNet network. Best performing network was selected for feature extraction. Features were extracted from 512 point sampled point clouds. Features were obtained from the GlobalMaxPooling1D layer of the Keras implementation. \%75 of the original point cloud based LSTM data (55445 samples) was used for training, \%25 remaining samples (18482 samples) were reserved for testing. For the synthetic point cloud based LSTM, \%75 of the data (55417 samples) was used for training, \%25 remaining samples (18473 samples) were reserved for testing. Original and synthetic based LSTM training data was used with five-fold cross validation. 

\begin{figure}[htb!]
	\centering
	\begin{subfigure}{0.1\textwidth}
		\includegraphics[height=60px, width=50px]{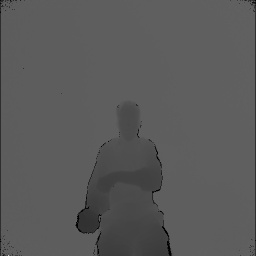}
		\subcaption{}
	\end{subfigure}
	\begin{subfigure}{0.1\textwidth}
		\includegraphics[height=60px, width=50px]{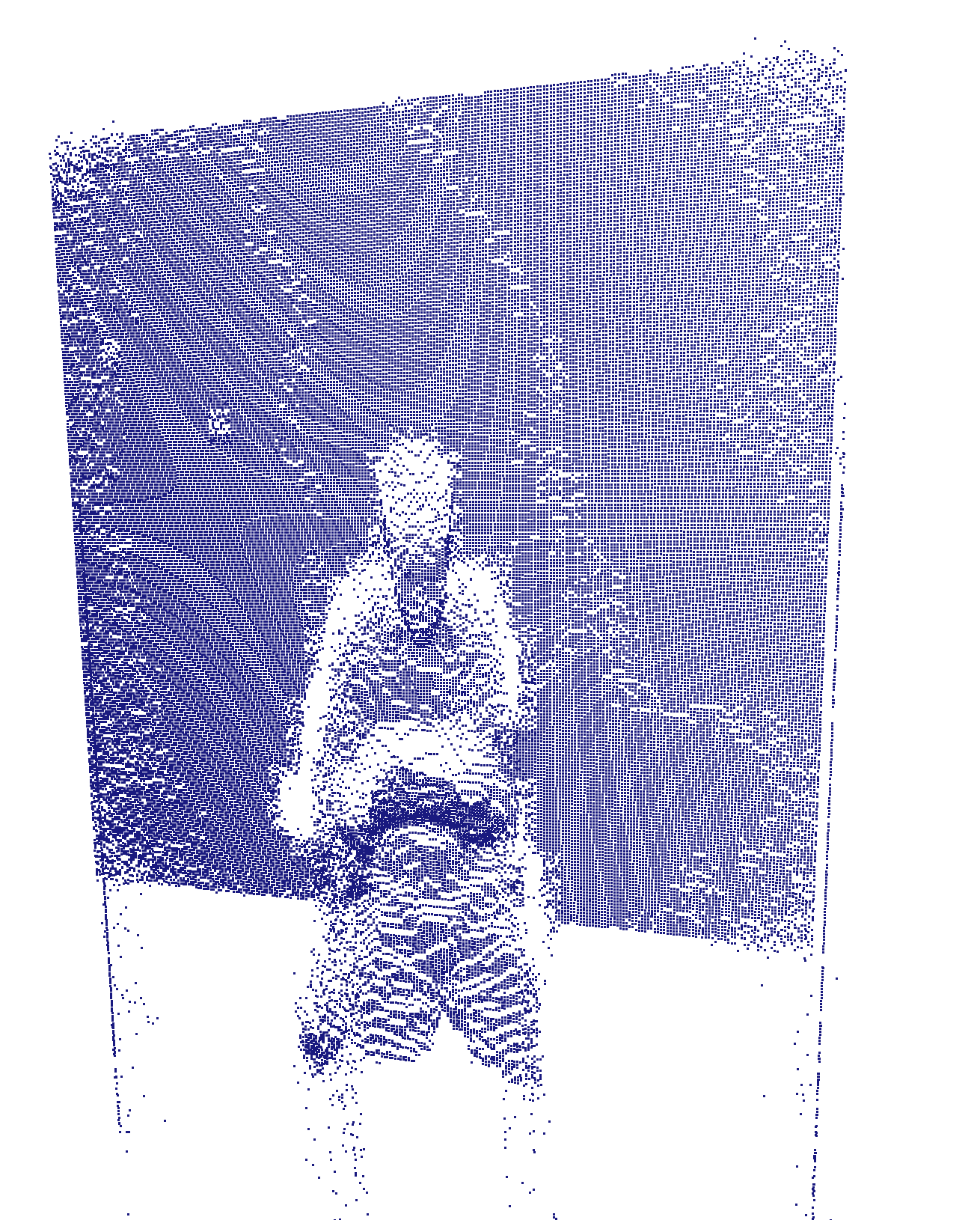}
		\subcaption{}
	\end{subfigure}
	\begin{subfigure}{0.1\textwidth}
		\includegraphics[height=60px, width=50px]{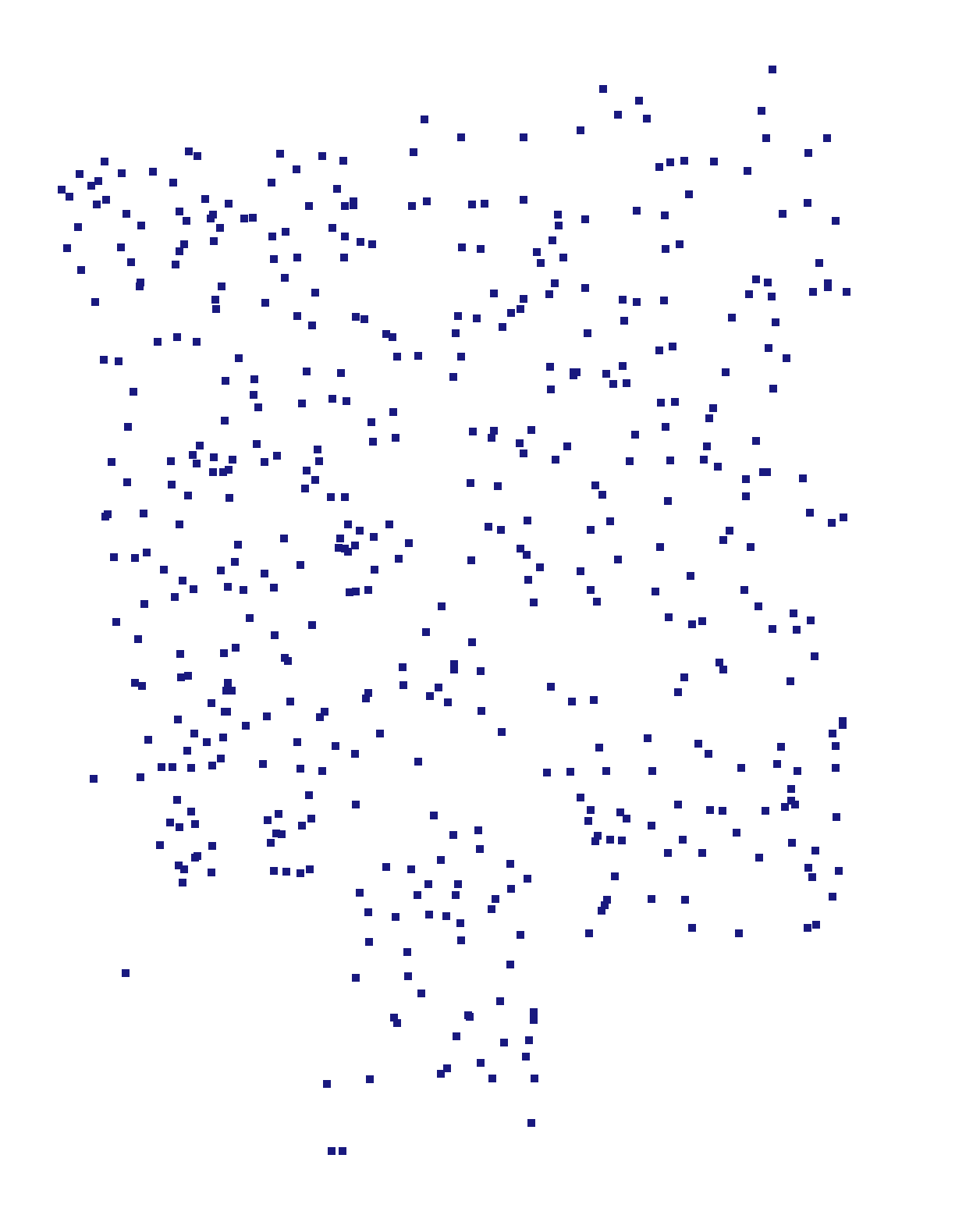}
		\subcaption{}
	\end{subfigure}
	\begin{subfigure}{0.1\textwidth}
		\includegraphics[height=60px, width=50px]{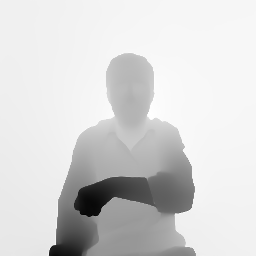}
		\subcaption{}
	\end{subfigure}
	\begin{subfigure}{0.1\textwidth}
		\includegraphics[height=60px, width=50px]{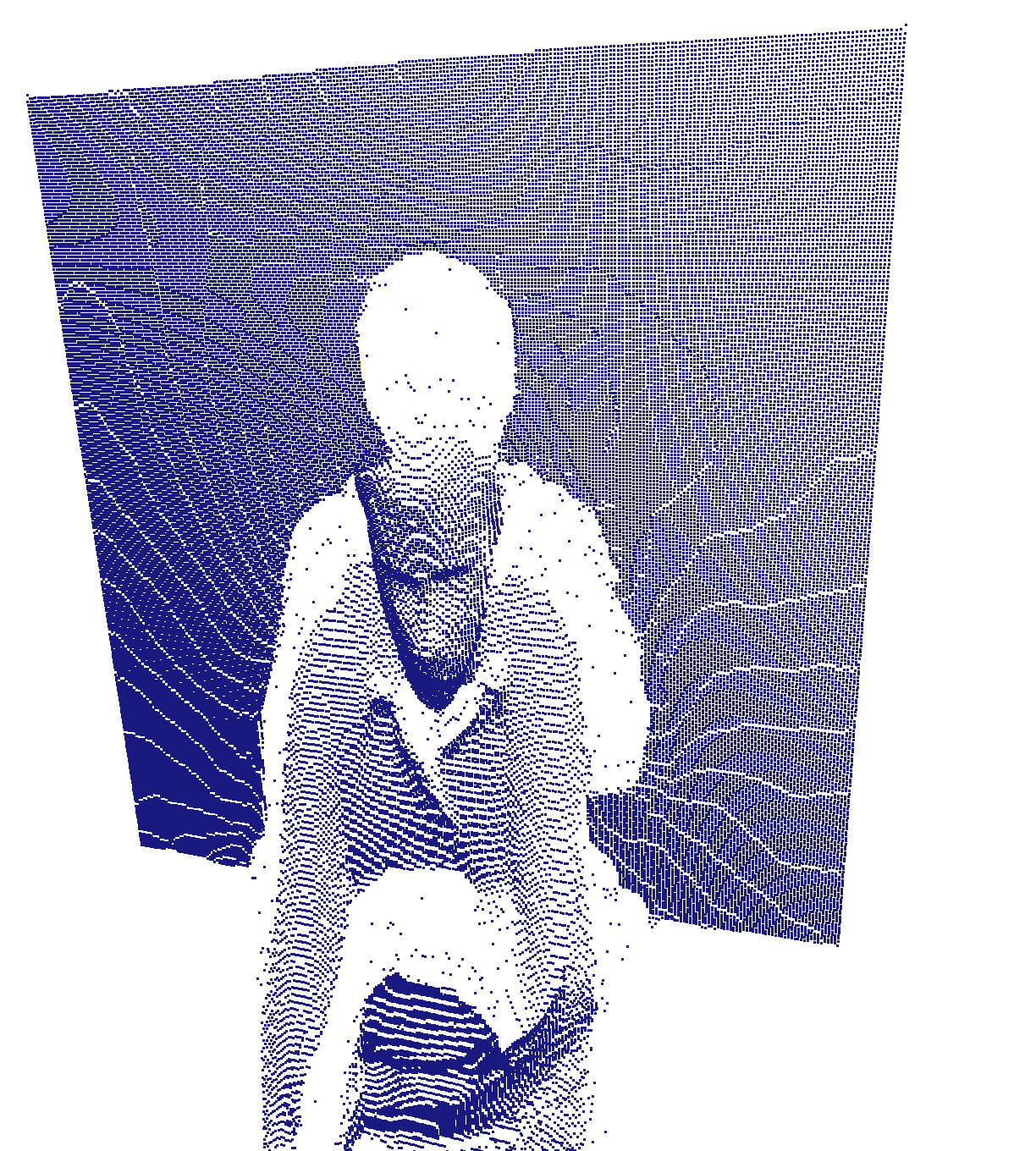}
		\subcaption{}
	\end{subfigure}
	\begin{subfigure}{0.1\textwidth}
		\includegraphics[height=60px, width=50px]{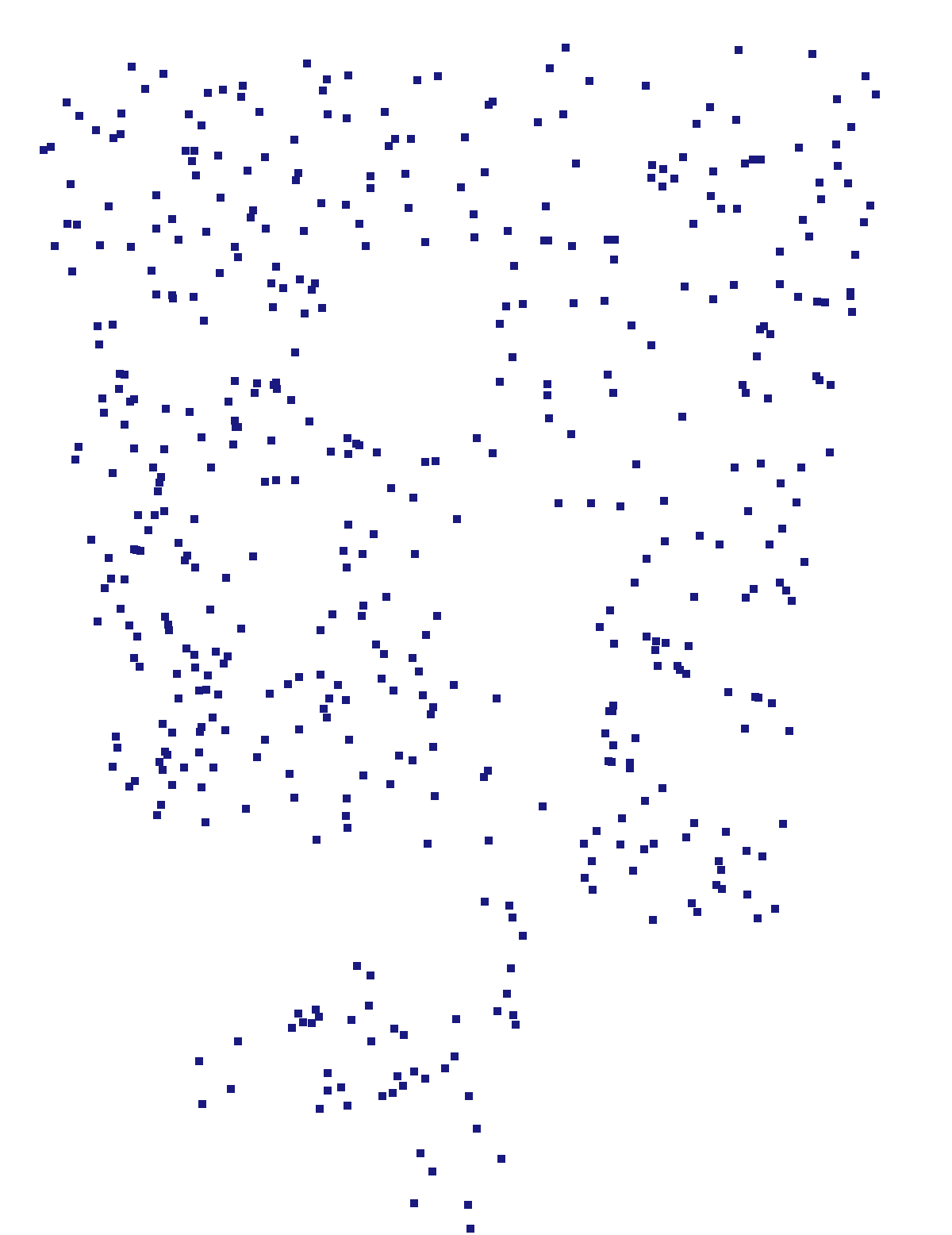}
		\subcaption{}
	\end{subfigure}
	\\
	\begin{subfigure}{0.1\textwidth}
		\includegraphics[height=60px, width=50px]{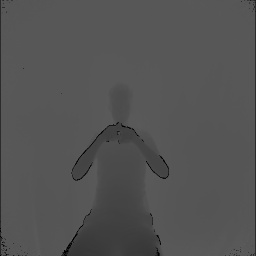}
		\subcaption{}
	\end{subfigure}
	\begin{subfigure}{0.1\textwidth}
		\includegraphics[height=60px, width=50px]{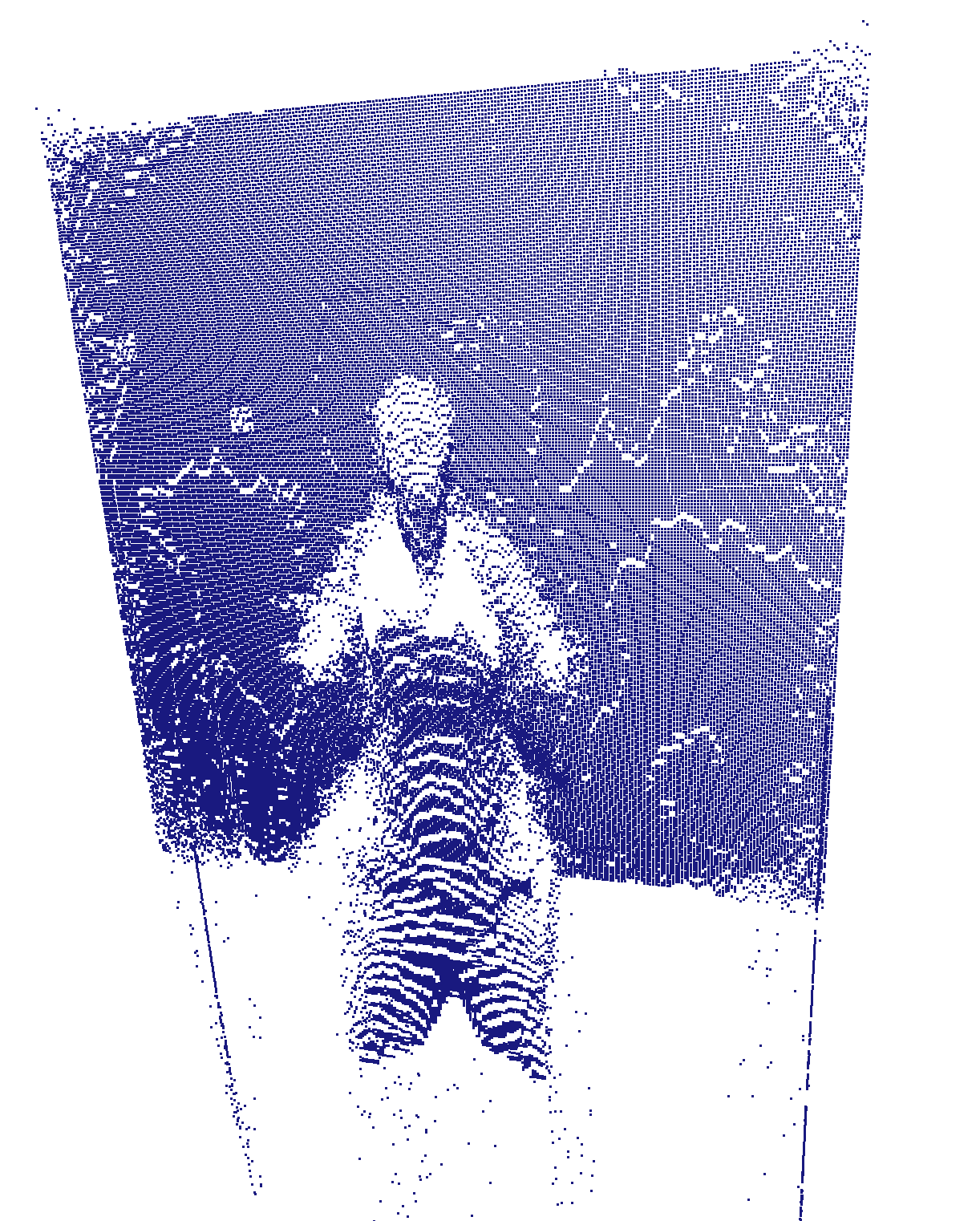}
		\subcaption{}
	\end{subfigure}
	\begin{subfigure}{0.1\textwidth}
		\includegraphics[height=60px, width=50px]{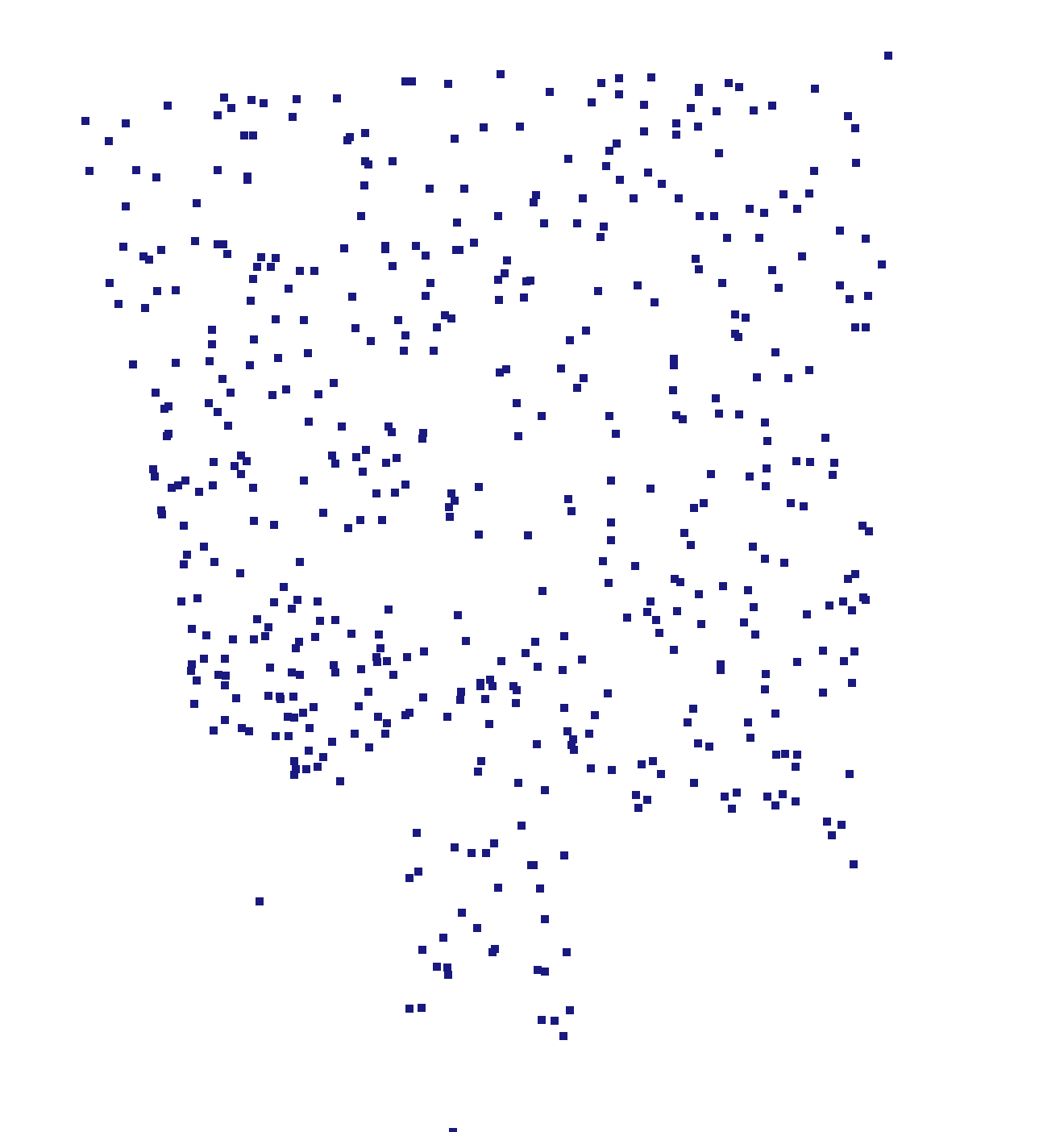}
		\subcaption{}
	\end{subfigure}
	\begin{subfigure}{0.1\textwidth}
		\includegraphics[height=60px, width=50px]{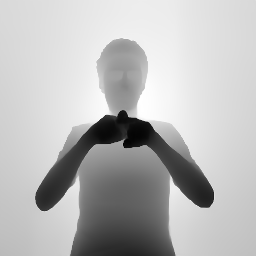}
		\subcaption{}
	\end{subfigure}
	\begin{subfigure}{0.1\textwidth}
		\includegraphics[height=60px, width=50px]{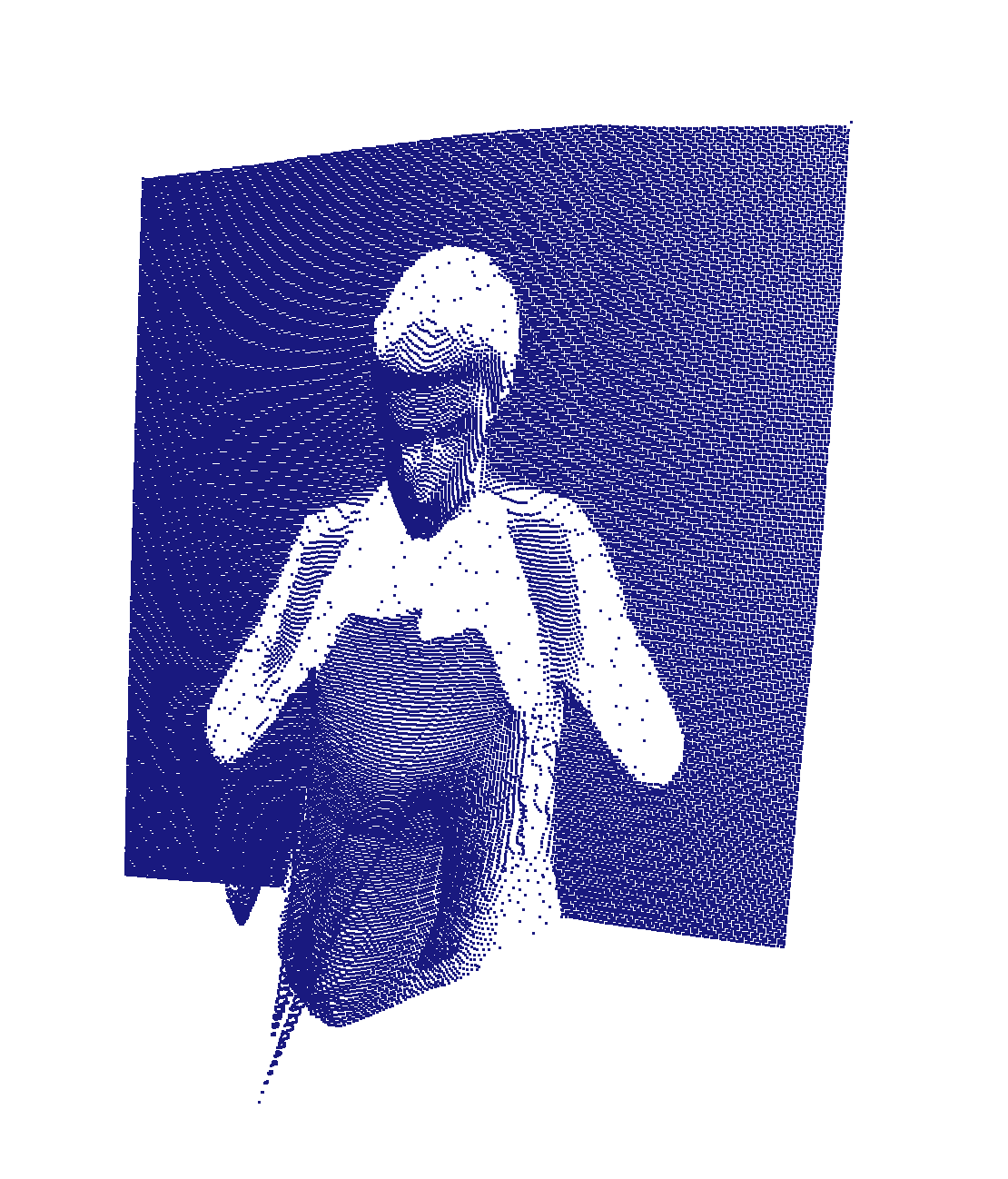}
		\subcaption{}
	\end{subfigure}
	\begin{subfigure}{0.1\textwidth}
		\includegraphics[height=60px, width=50px]{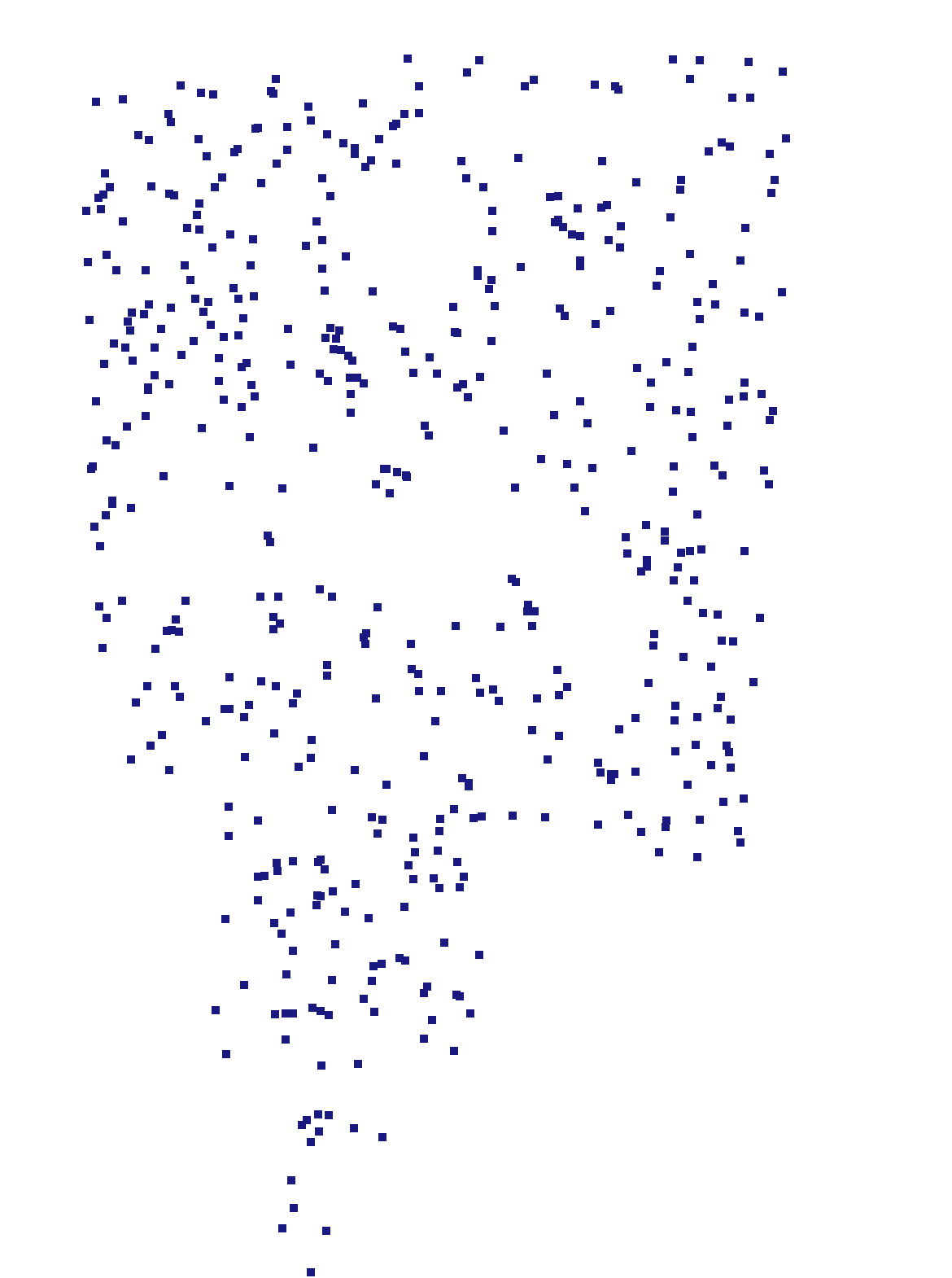}
		\subcaption{}
	\end{subfigure}
	\\
	\begin{subfigure}{0.1\textwidth}
		\includegraphics[height=60px, width=50px]{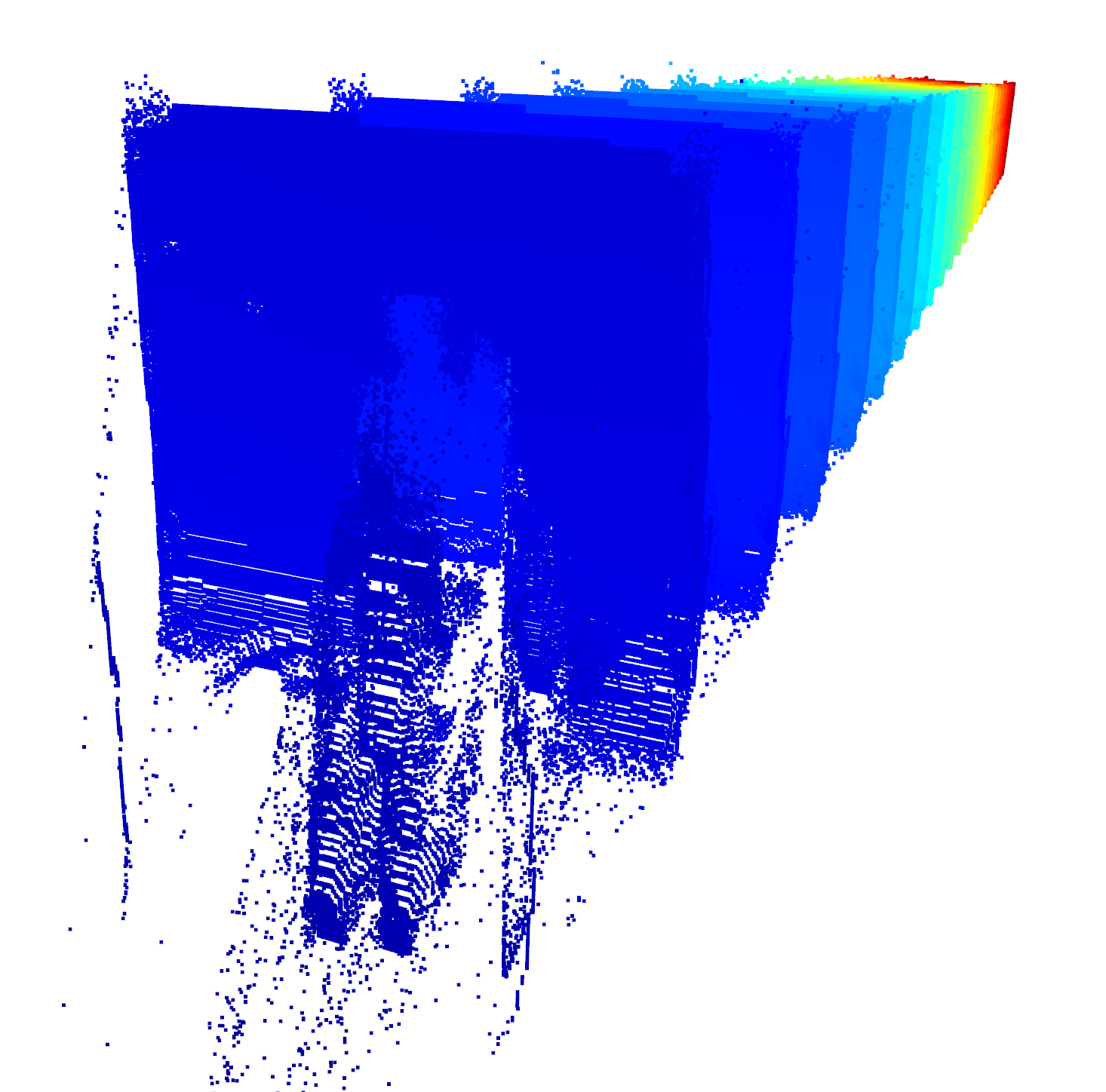}
		\subcaption{}
	\end{subfigure}
	\begin{subfigure}{0.1\textwidth}
		\includegraphics[height=60px, width=50px]{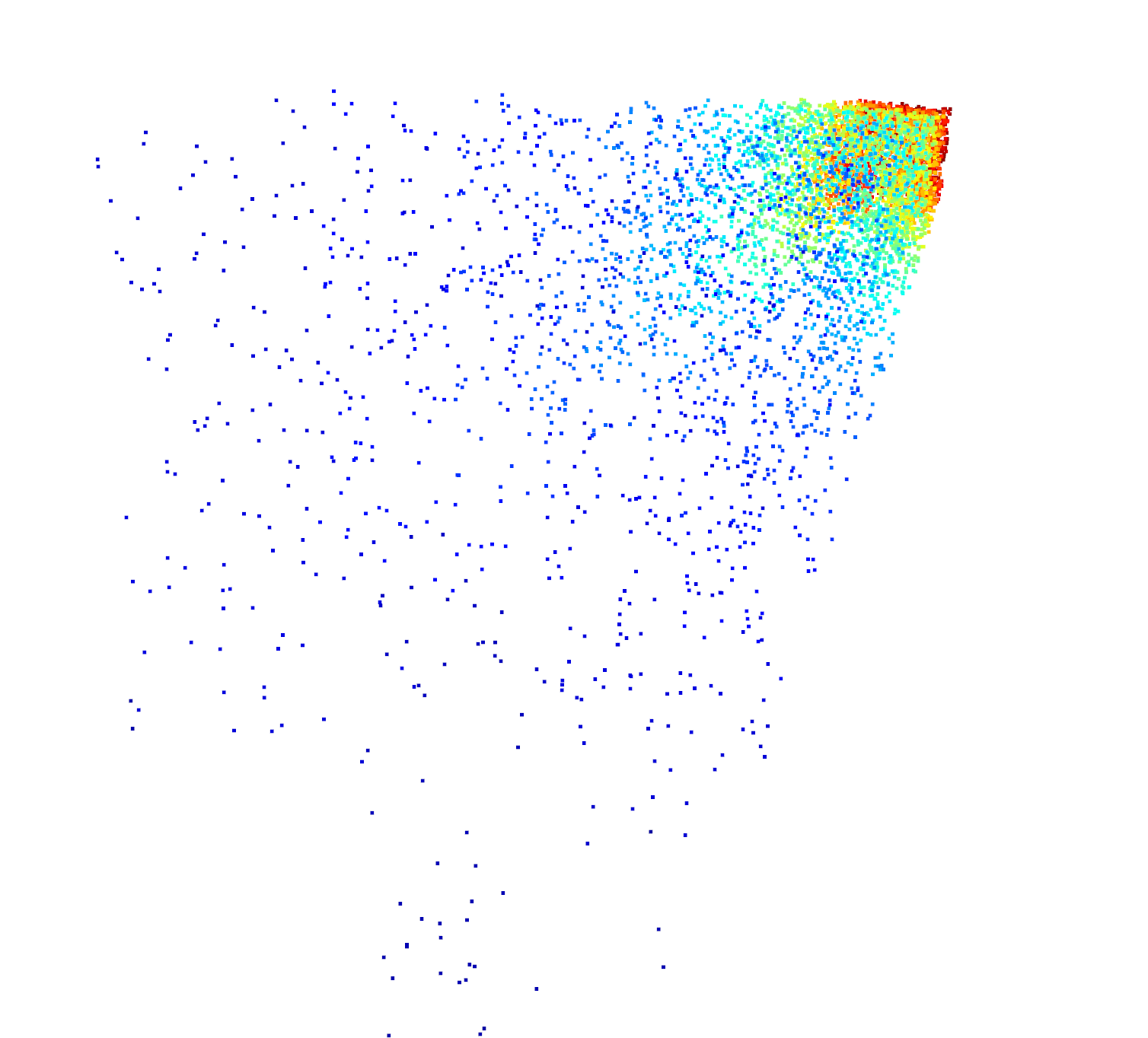}
		\subcaption{}
	\end{subfigure}
	\begin{subfigure}{0.1\textwidth}
		\includegraphics[height=60px, width=50px]{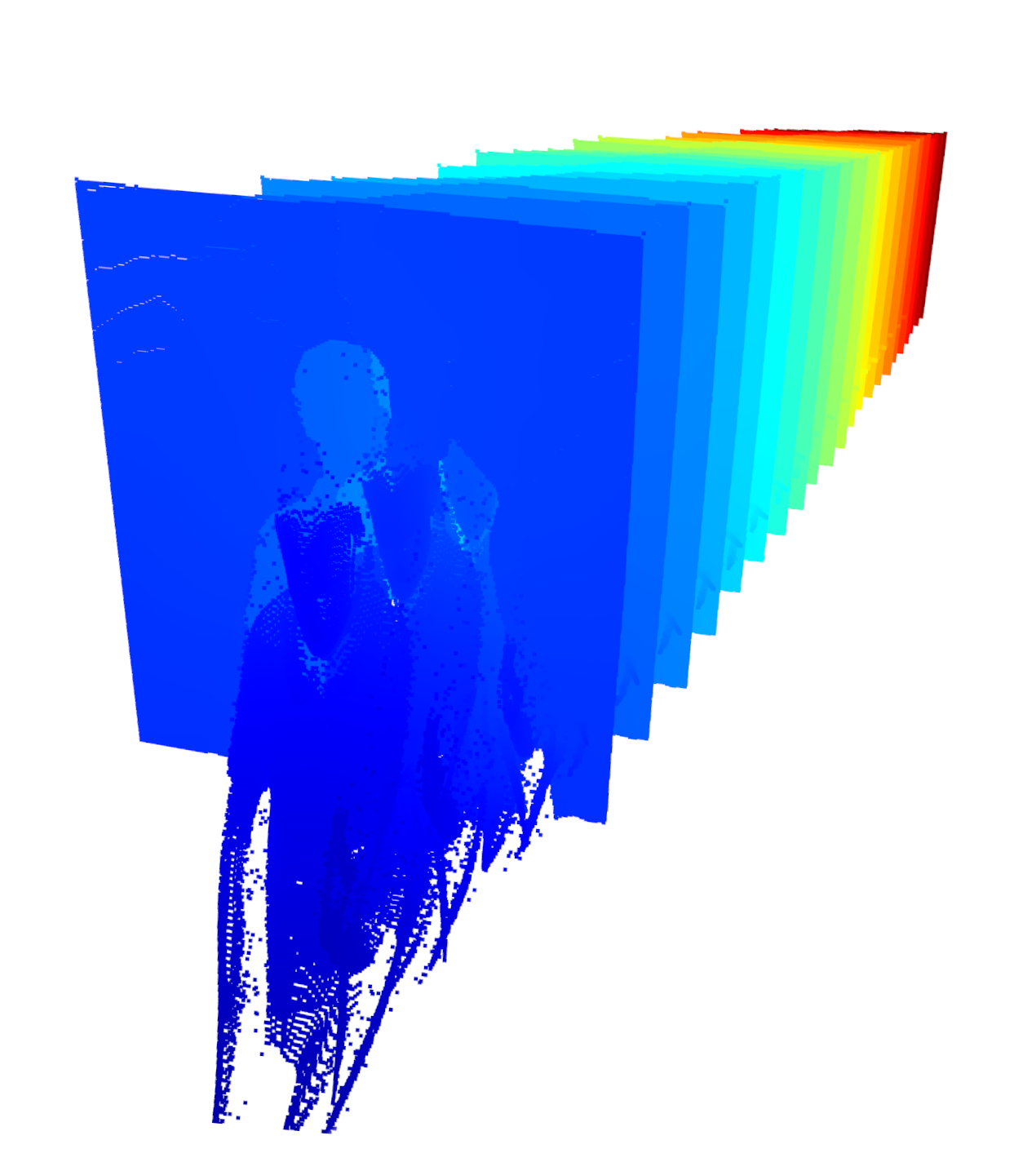}
		\subcaption{}
	\end{subfigure}
	\begin{subfigure}{0.1\textwidth}
		\includegraphics[height=60px, width=50px]{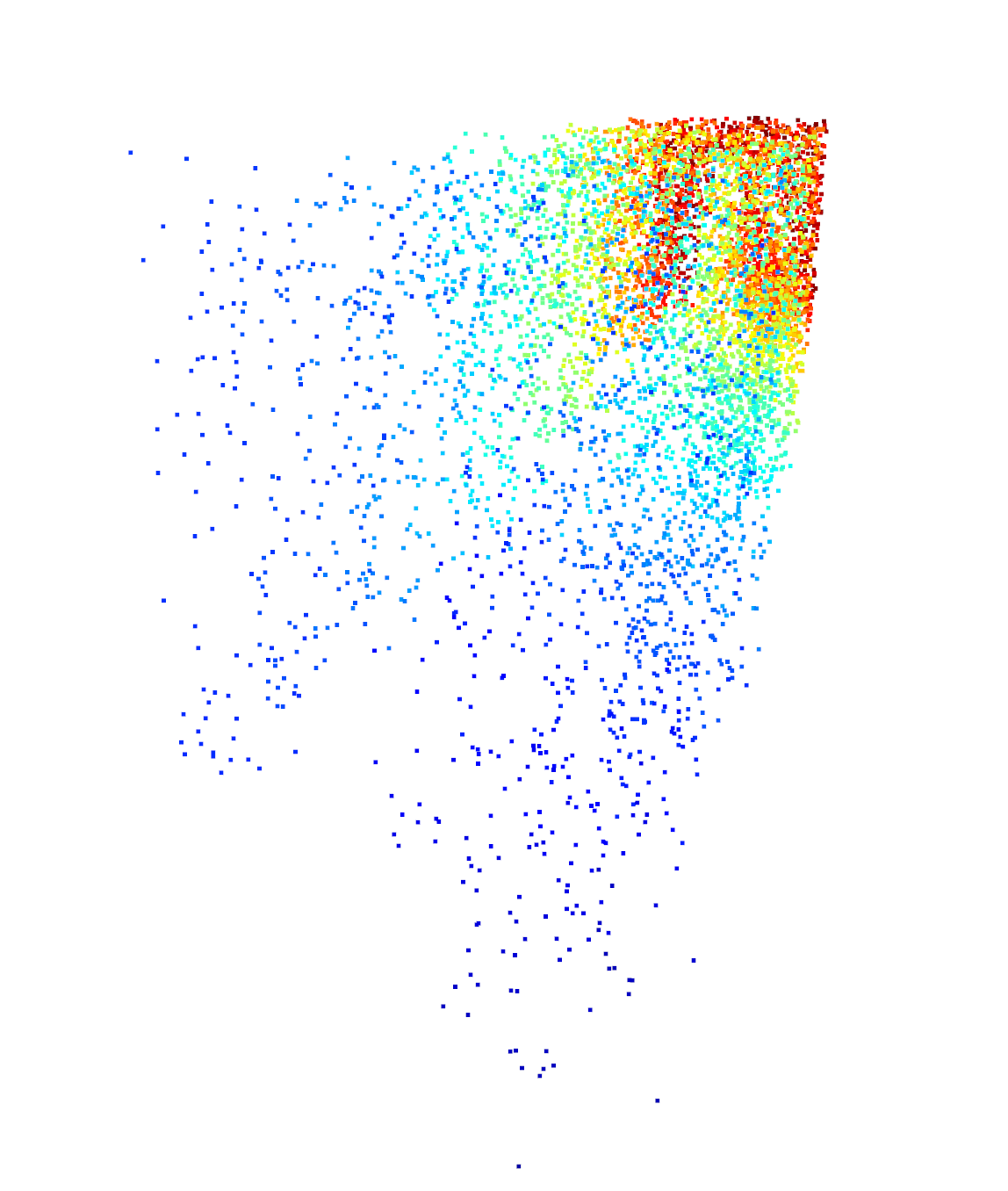}
		\subcaption{}
	\end{subfigure}
	\\
	\begin{subfigure}{0.1\textwidth}
		\includegraphics[height=60px, width=50px]{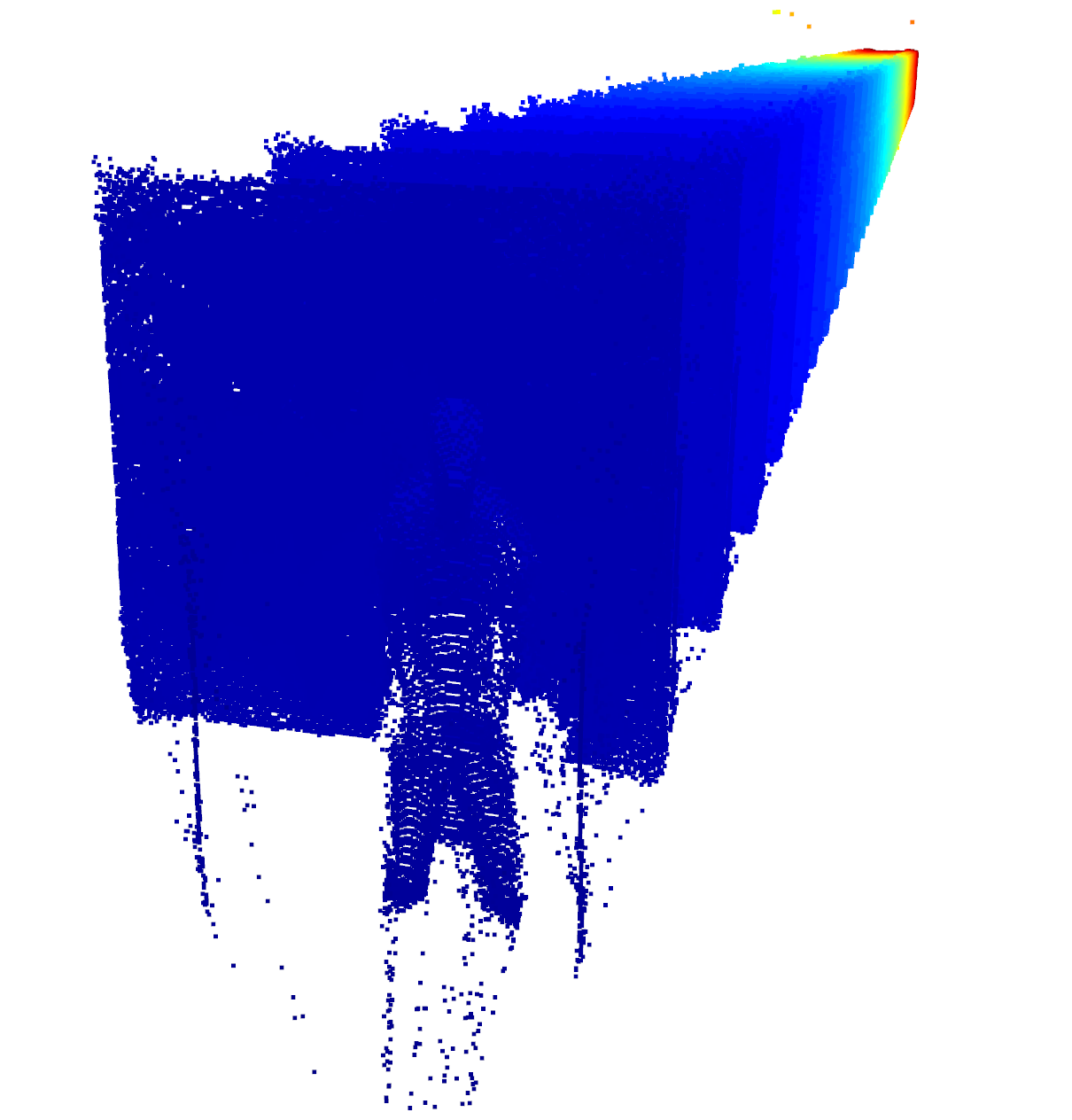}
		\subcaption{}
	\end{subfigure}
	\begin{subfigure}{0.1\textwidth}
		\includegraphics[height=60px, width=50px]{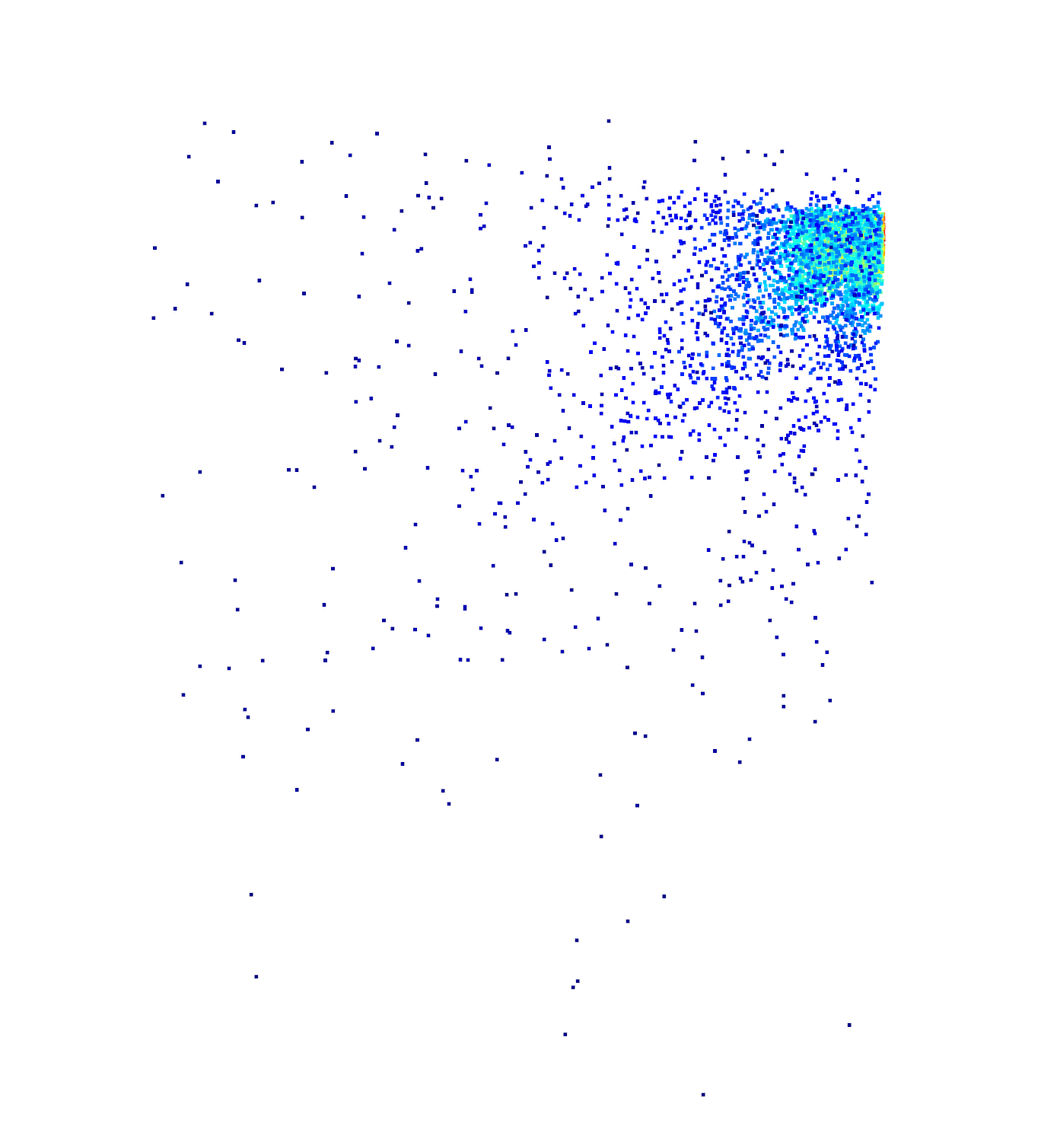}
		\subcaption{}
	\end{subfigure}
	\begin{subfigure}{0.1\textwidth}
		\includegraphics[height=60px, width=50px]{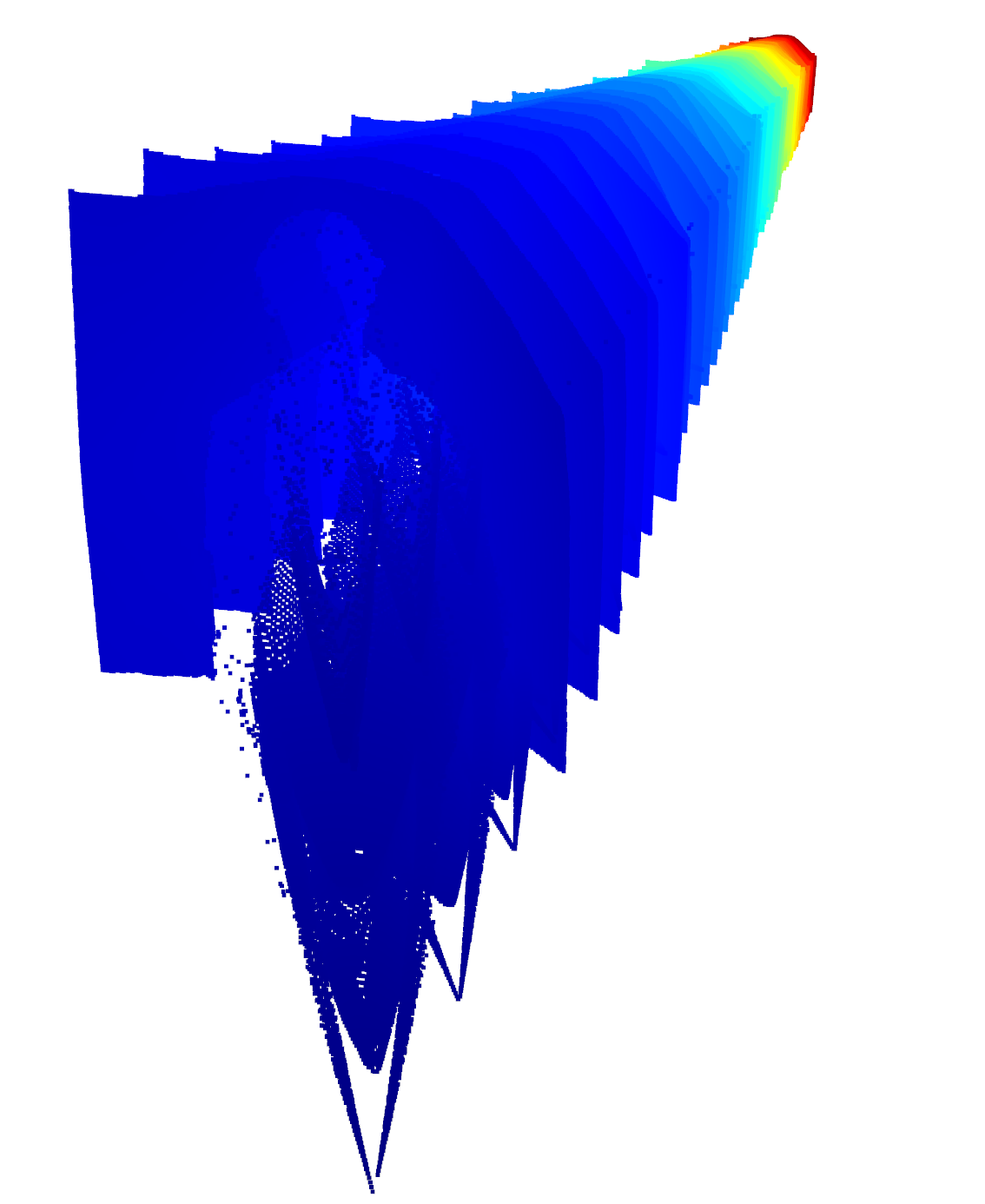}
		\subcaption{}
	\end{subfigure}
	\begin{subfigure}{0.1\textwidth}
		\includegraphics[height=60px, width=50px]{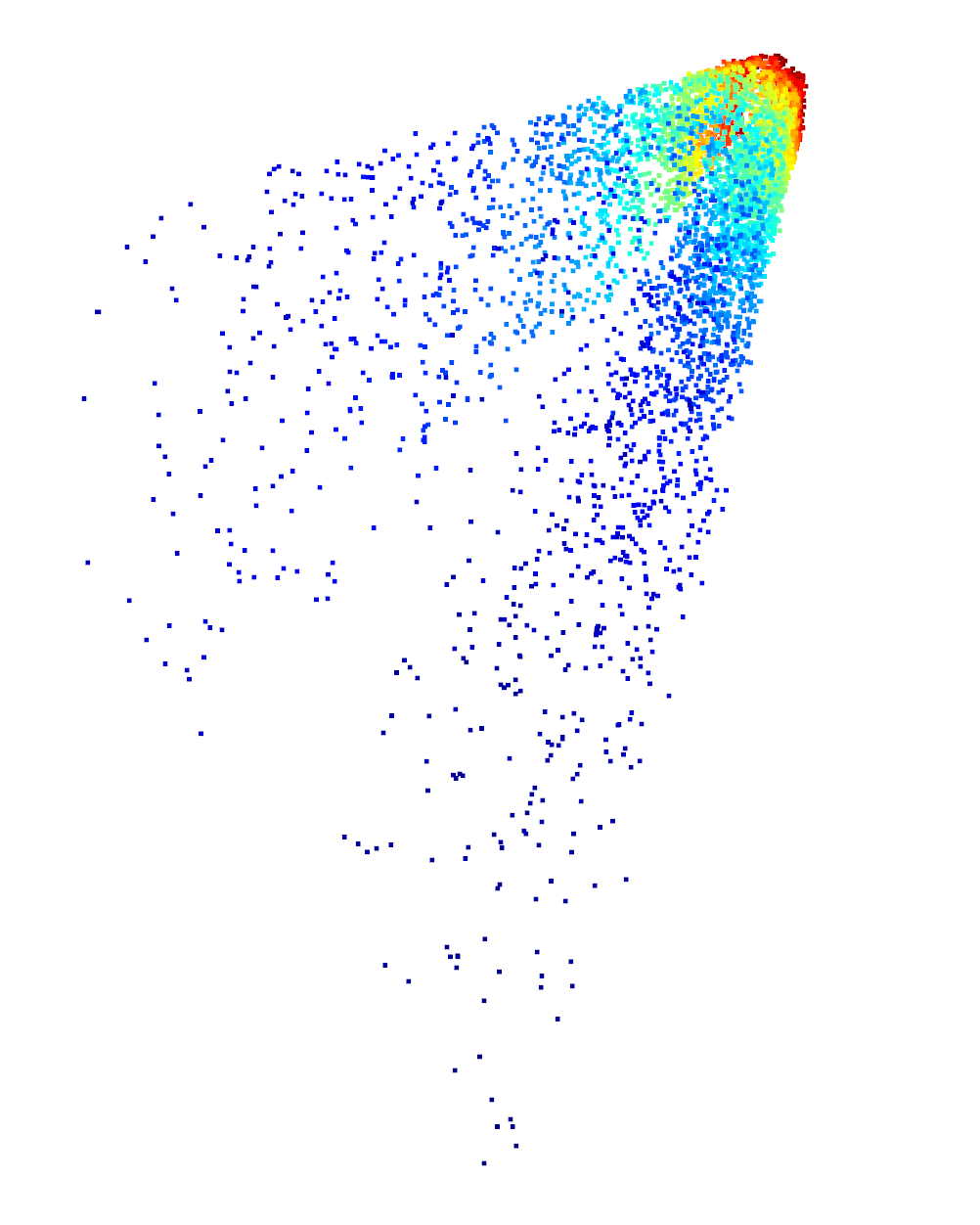}
		\subcaption{}
	\end{subfigure}	
	\caption{\label{datasetB-examples2} Examples from the data created from Dataset-B. Top two rows: (a, g) Original depth, (b, h) Raw point cloud, (c, i) 512 sampled point cloud, (d, j) Synthetic depth, (e, k) Raw point cloud (f, l) 512 sampled point cloud. Bottom two row (m, q) Original depth PGM, (n, r) 6400 sampled PGM, (o, s) Synthetic PGM, (p, t) 6400 sampled PGM.}
\end{figure}

For all data in this dataset, depth-scale parameter was set to 500, depth-trunc was set to 1000 in Open3d. Figure \ref{datasetB-examples2} shows example data created from this dataset.

\underline{\textbf{Dataset-C (AUTSL):}} In Turkish dataset, class labels of the each sample is presented in separate CSV files as train, validation and test. Both RGB and depth videos of a gesture exists with the name listed in the CSV file. Each RGB and depth video of a sample was read frame by frame simultaneously, then original depth and synthetic depth were placed in their separate class folders. Since classes of the videos are unordered, a counter for each class was defined, where it holds the processed total frame amount of every class. File names of the frames of a video sample were given using these counters.

Point clouds were created using Open3d with the same parameters as the Dataset-B, because this dataset was also created using Kinect V2. From the original depth point clouds, \%75 of the total frames (1660229) were used for training. \%25 (553410) were reserved for testing. From the synthetic depth point clouds, also \%75 of the total frames (1660230) were used for training, \%25 were (553410) reserved for testing. Raw point clouds were sampled using 512 points. Both original and synthetic data was used with five-fold cross validation, 4/5 for training and 1/5 for validation.

For the original PGM data, 27226 samples (\%75) were used for training. Remaining 9076 samples were reserved for testing. For the synthetic PGM data, 22690 samples (\%75) were used for training, 7564 were reserved for testing. Five-fold cross validation was applied to both original and synthetic based PGMs. When creating PGMs, all frames of a video sample were used to create a raw PGM for the gesture sample. From this raw PGM, 6400 points were sampled to create the final data.

LSTM data was also created because of dataset having spatio-temporal dimension. Dataset was examined to determine an average frame amount for each gesture sample. 30 frames were determined as a suitable amount. Ordered random frames were selected for the samples having frames more than 30. For the samples with a frame amount between 20 and 30, frame interpolation method mentioned in the Dataset-B was used. 

Best performing pretrained frame based PointNet weights were used to extract features for the LSTM data preparation. Similar to Dataset-B, features were obtained from the GlobalMaxPooling1D layer of the Keras implementation. 26936 (\%75) samples were used for original point cloud based LSTM training. 8979 (\%25) samples were reserved for the testing. For the synthetic point cloud based LSTM, 28173 samples (\%75) were used for training. \%25 (9391) was reserved for testing. LSTM data was used with five-fold cross validation. Features were extracted from the point clouds that were sampled with 512 points.

\begin{figure}[htb!]
	\centering
	\begin{subfigure}{0.1\textwidth}
		\includegraphics[height=60px, width=50px]{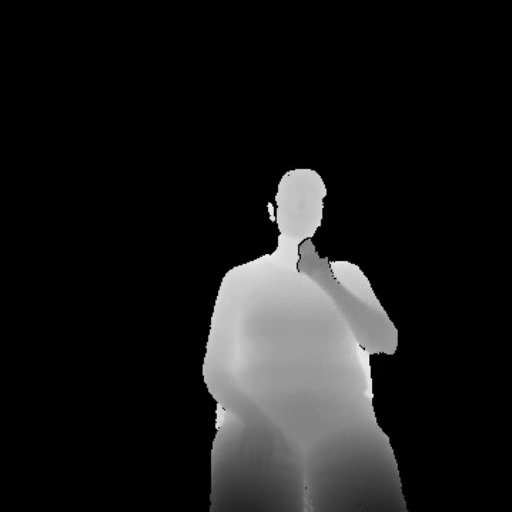}
		\subcaption{}
	\end{subfigure}
	\begin{subfigure}{0.1\textwidth}
		\includegraphics[height=60px, width=50px]{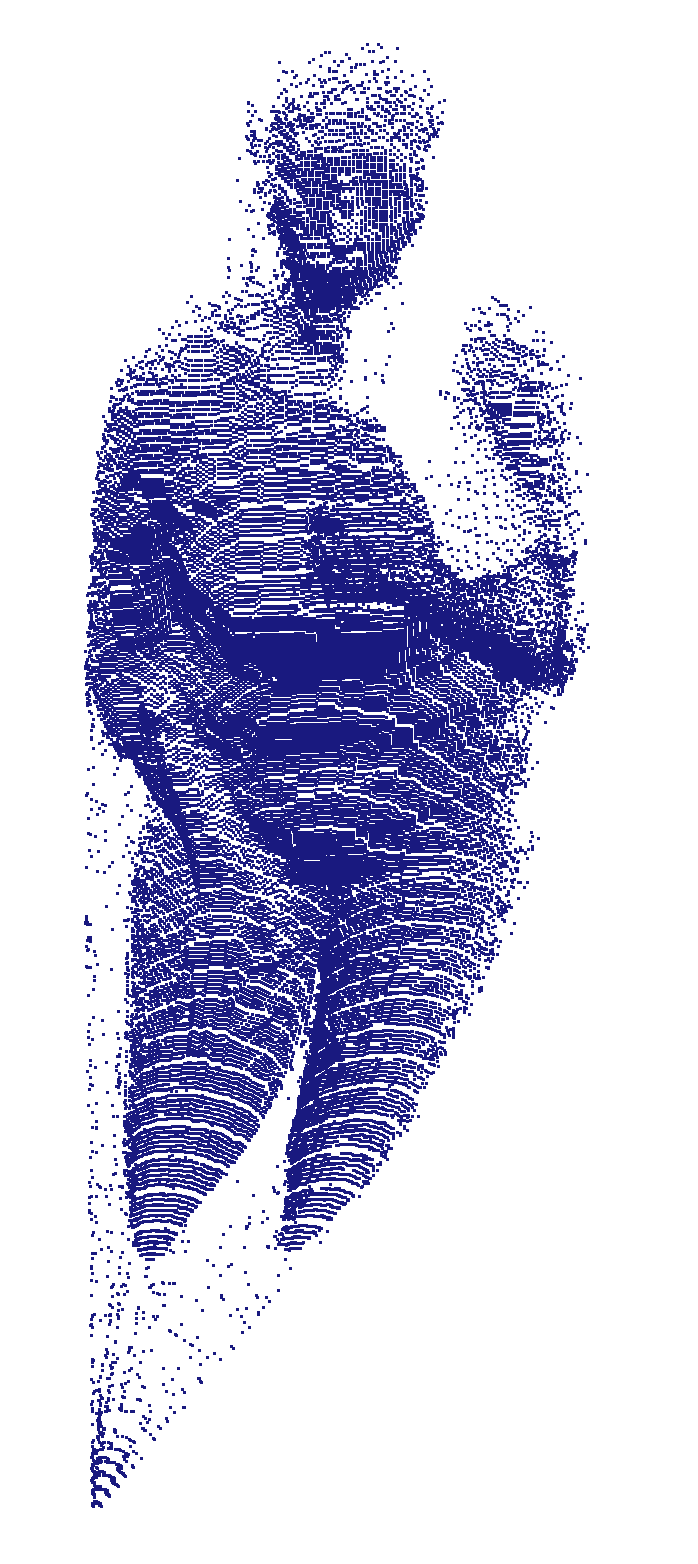}
		\subcaption{}
	\end{subfigure}
	\begin{subfigure}{0.1\textwidth}
		\includegraphics[height=60px, width=50px]{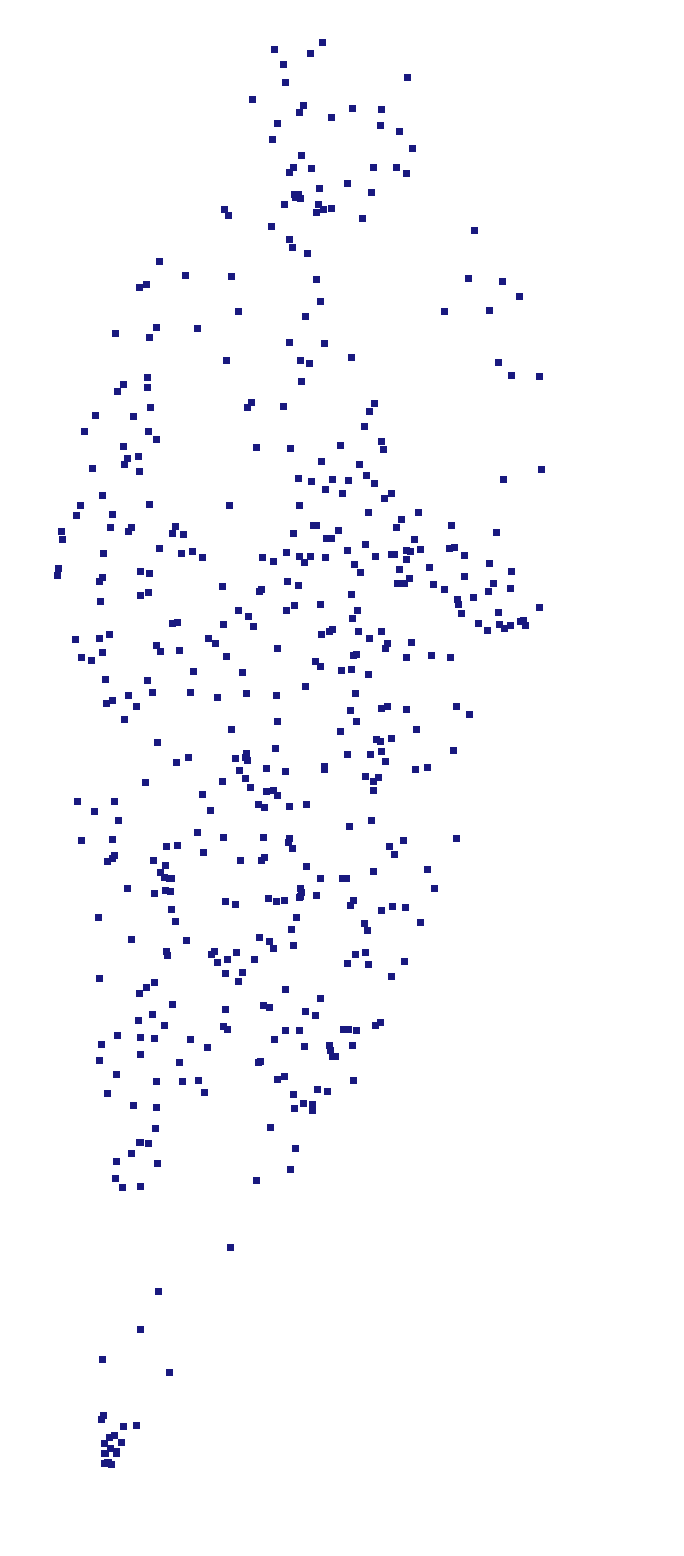}
		\subcaption{}
	\end{subfigure}
	\begin{subfigure}{0.1\textwidth}
		\includegraphics[height=60px, width=50px]{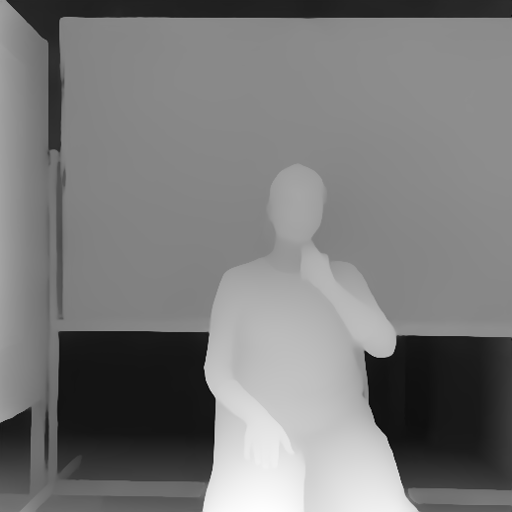}
		\subcaption{}
	\end{subfigure}
	\begin{subfigure}{0.1\textwidth}
		\includegraphics[height=60px, width=50px]{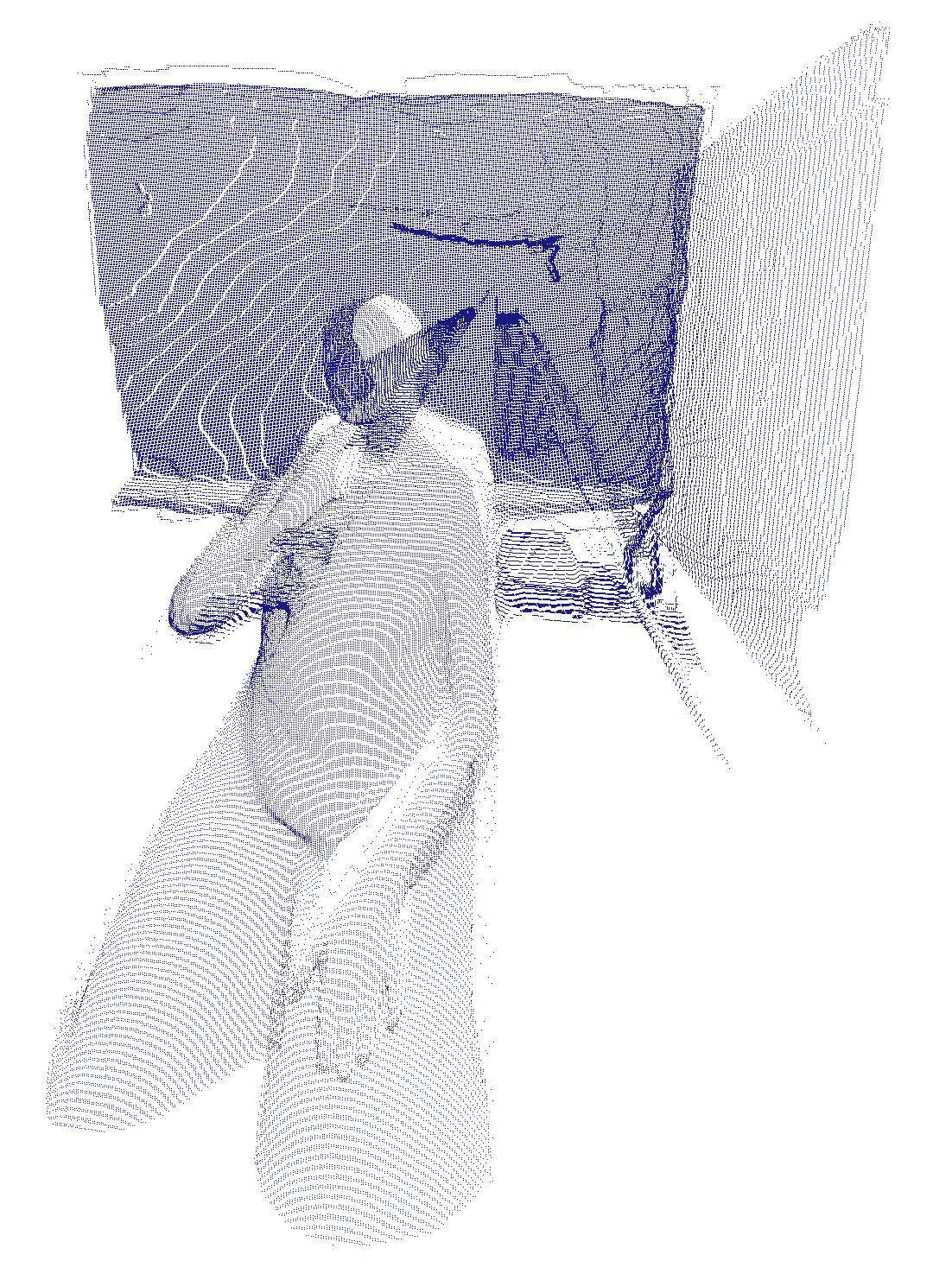}
		\subcaption{}
	\end{subfigure}
	\begin{subfigure}{0.1\textwidth}
		\includegraphics[height=60px, width=50px]{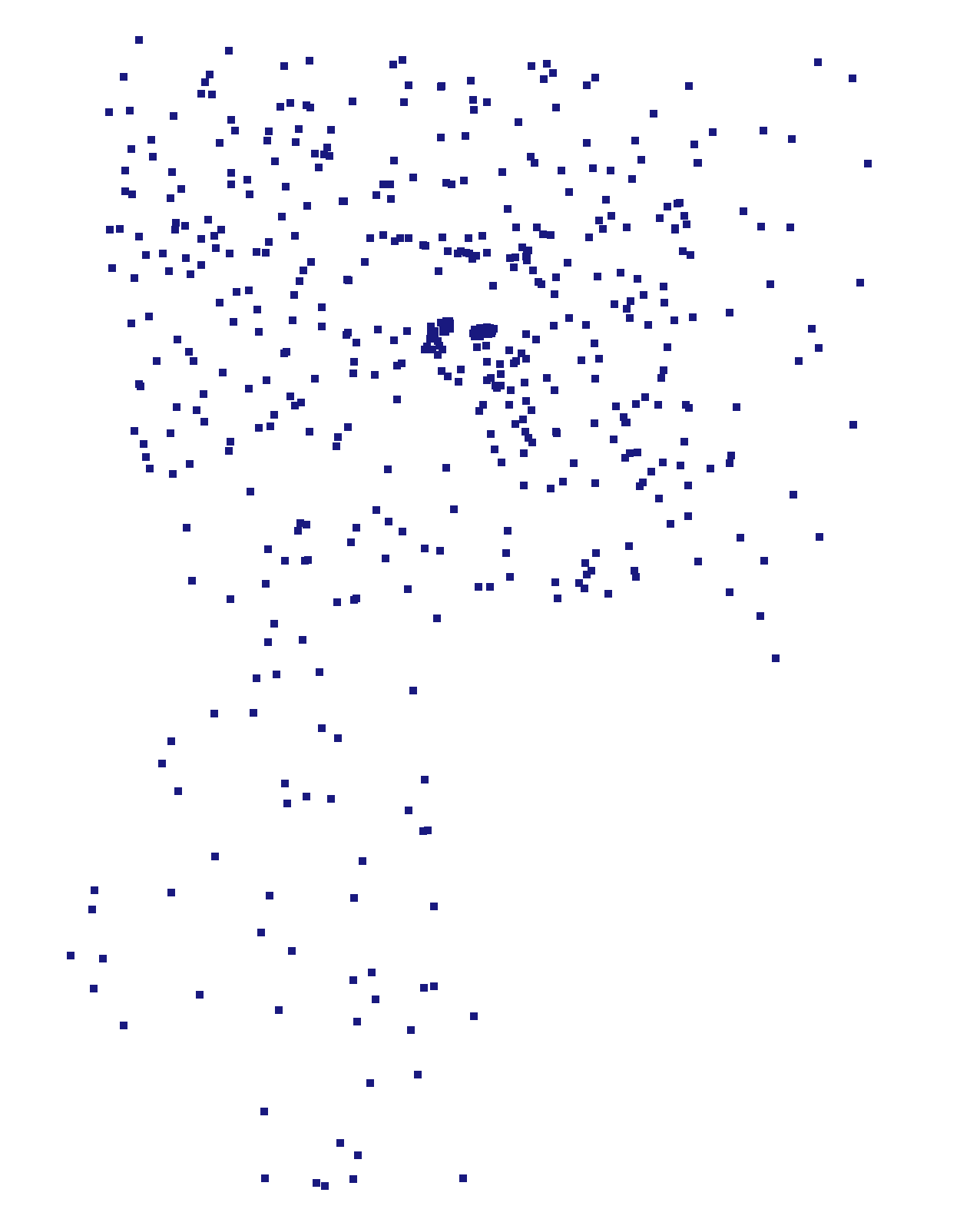}
		\subcaption{}
	\end{subfigure}
	\\
	\begin{subfigure}{0.1\textwidth}
		\includegraphics[height=60px, width=50px]{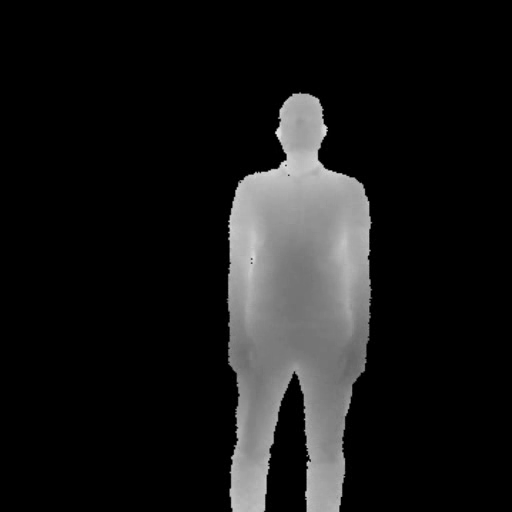}
		\subcaption{}
	\end{subfigure}
	\begin{subfigure}{0.1\textwidth}
		\includegraphics[height=60px, width=50px]{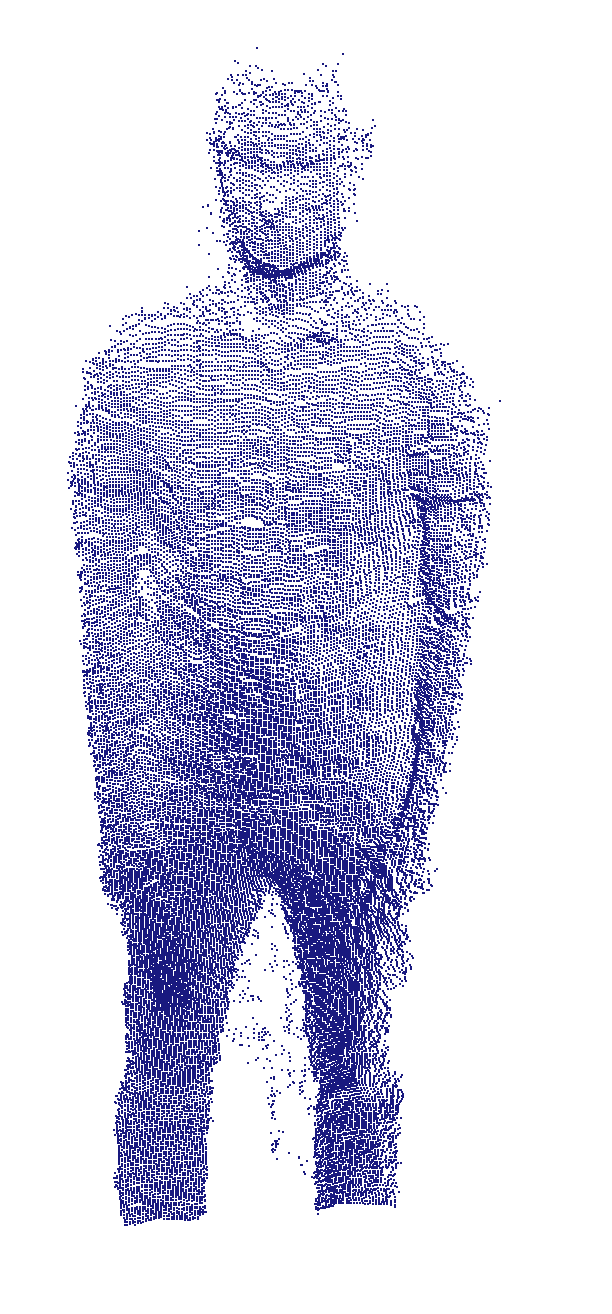}
		\subcaption{}
	\end{subfigure}
	\begin{subfigure}{0.1\textwidth}
		\includegraphics[height=60px, width=50px]{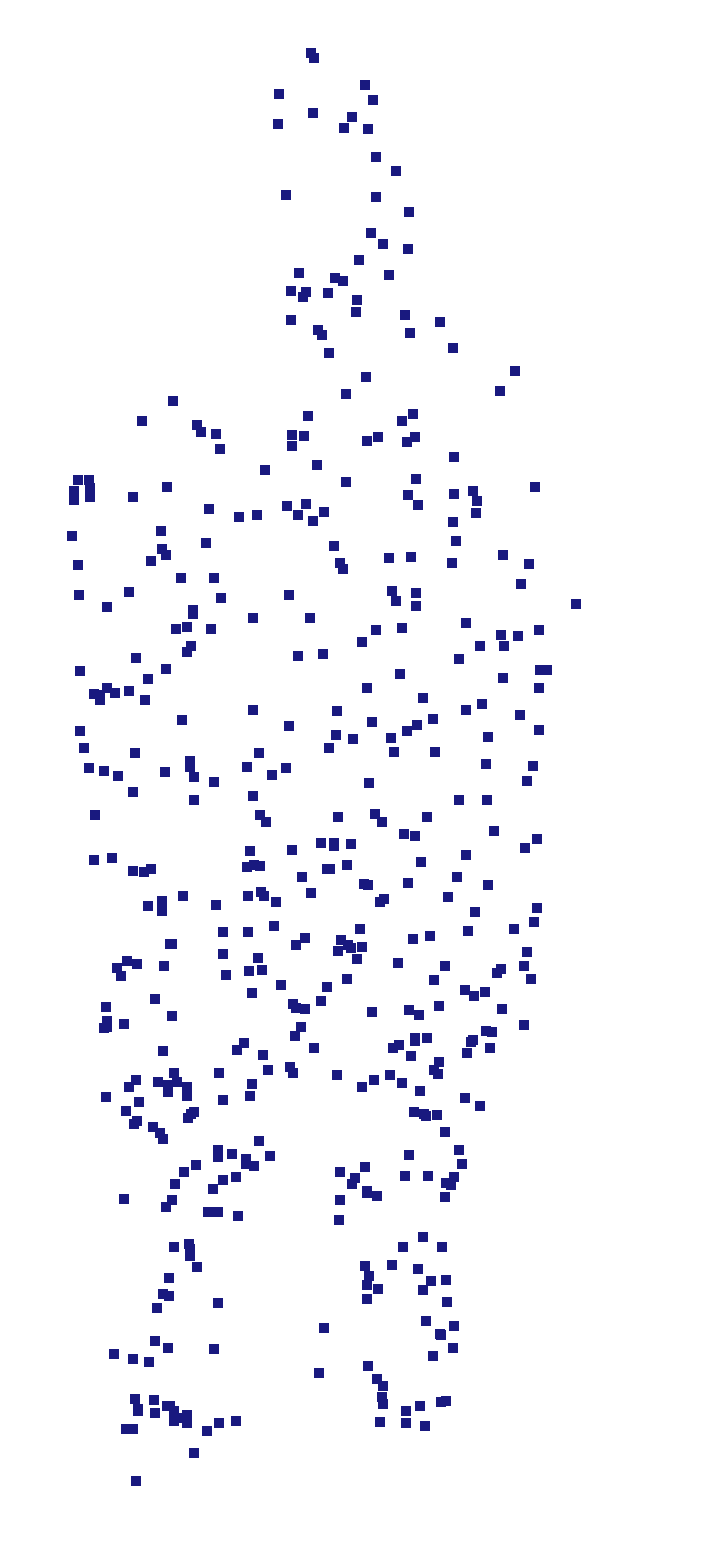}
		\subcaption{}
	\end{subfigure}
	\begin{subfigure}{0.1\textwidth}
		\includegraphics[height=60px, width=50px]{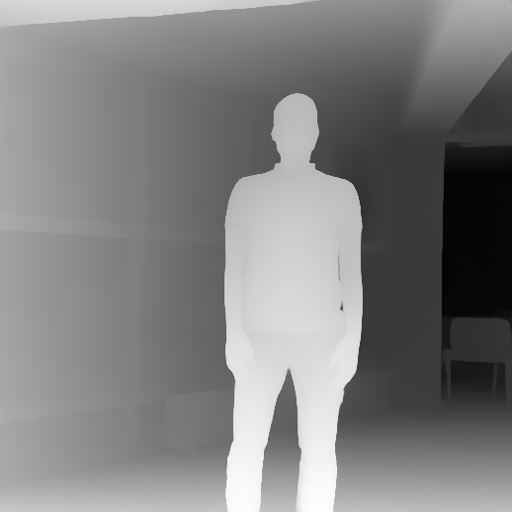}
		\subcaption{}
	\end{subfigure}
	\begin{subfigure}{0.1\textwidth}
		\includegraphics[height=60px, width=50px]{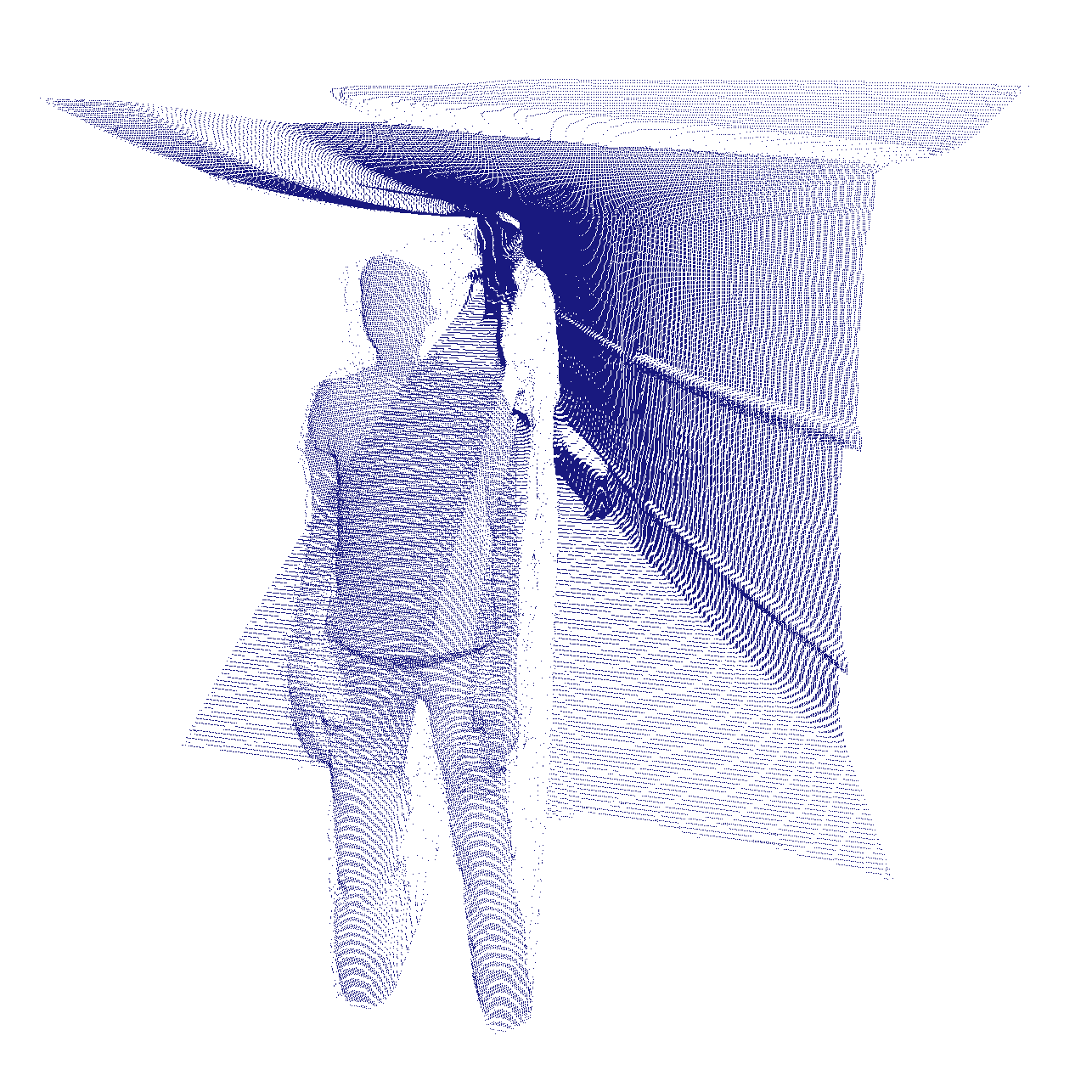}
		\subcaption{}
	\end{subfigure}
	\begin{subfigure}{0.1\textwidth}
		\includegraphics[height=60px, width=50px]{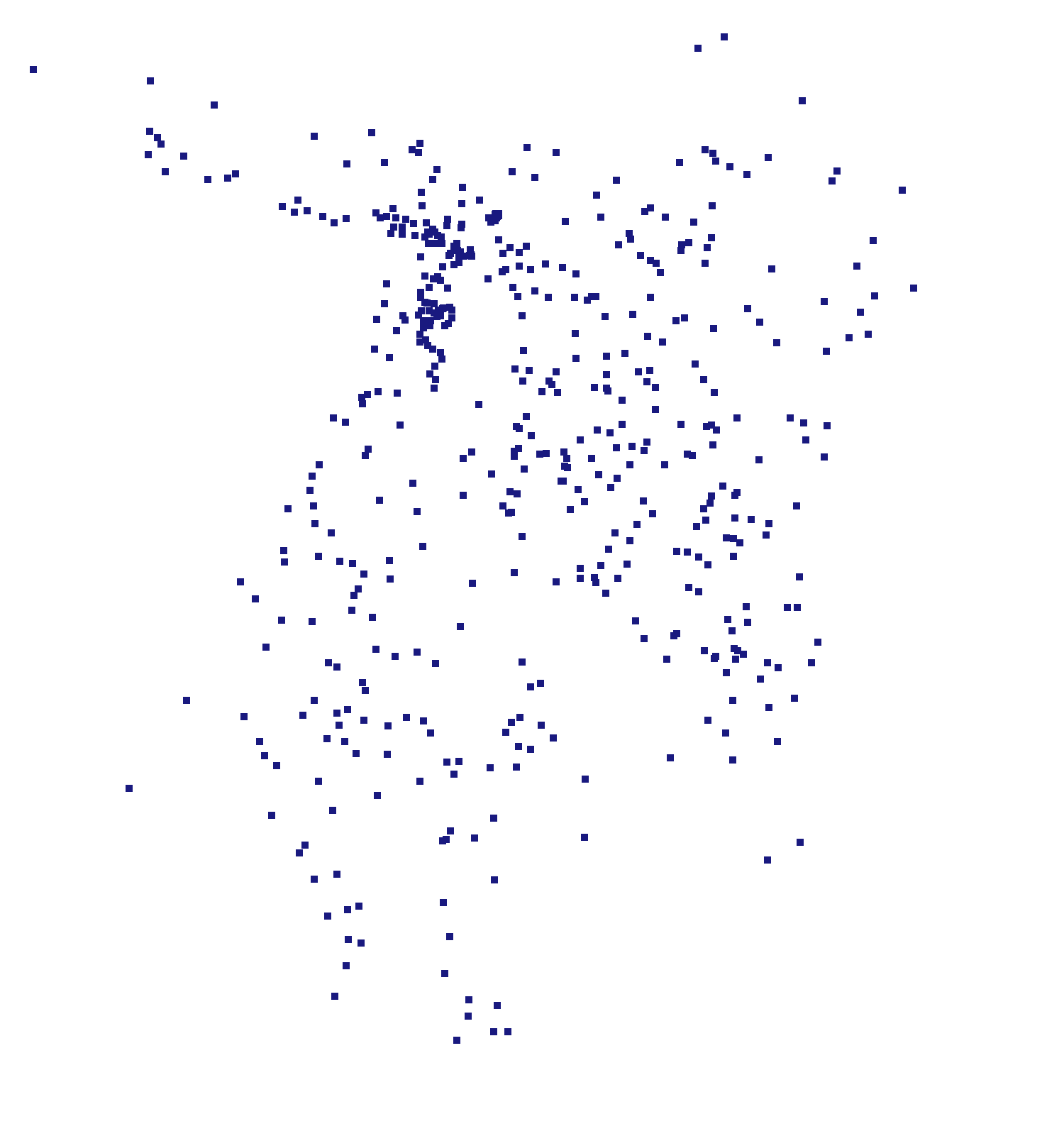}
		\subcaption{}
	\end{subfigure}
	\\
	\begin{subfigure}{0.1\textwidth}
		\includegraphics[height=60px, width=50px]{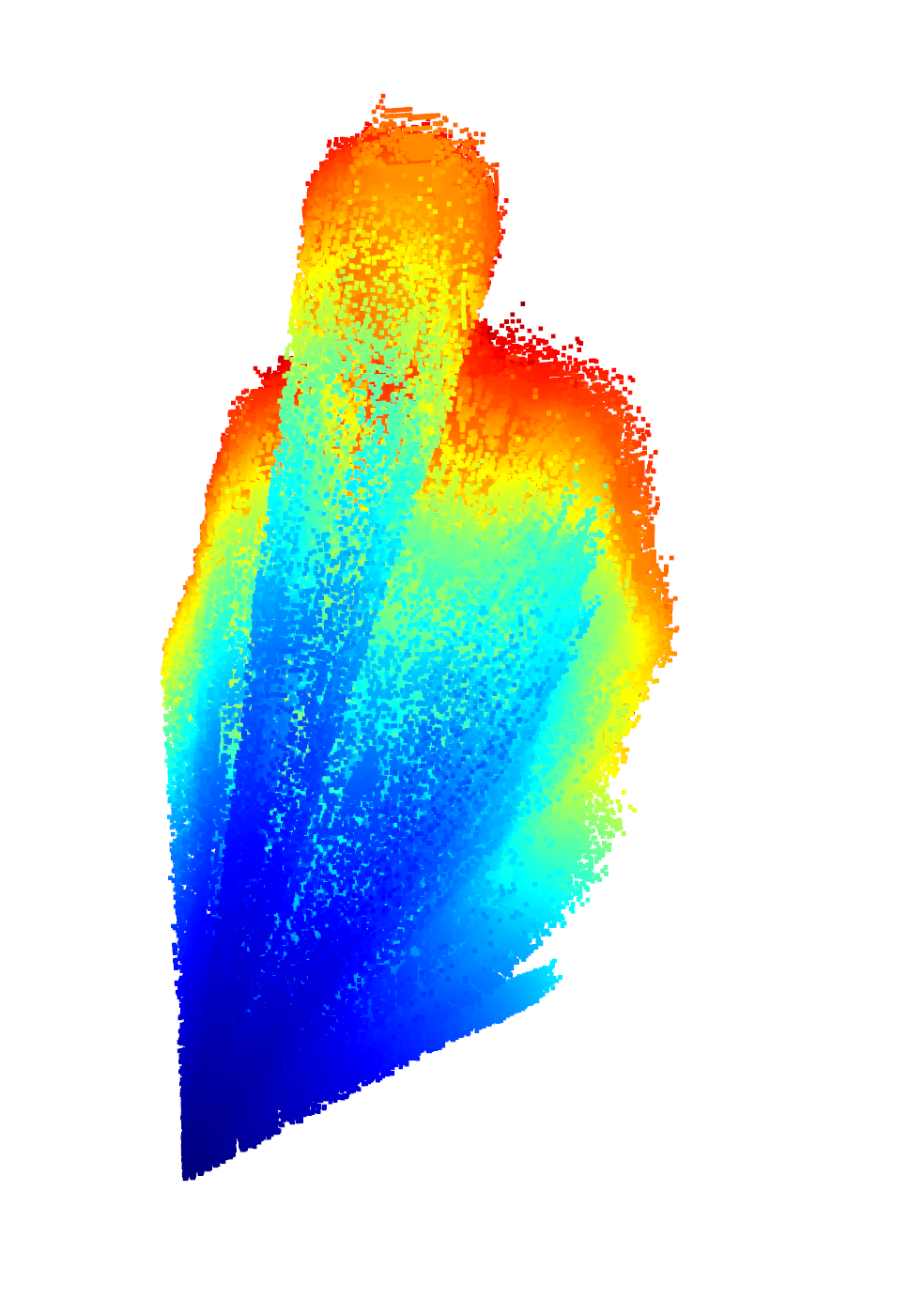}
		\subcaption{}
	\end{subfigure}
	\begin{subfigure}{0.1\textwidth}
		\includegraphics[height=60px, width=50px]{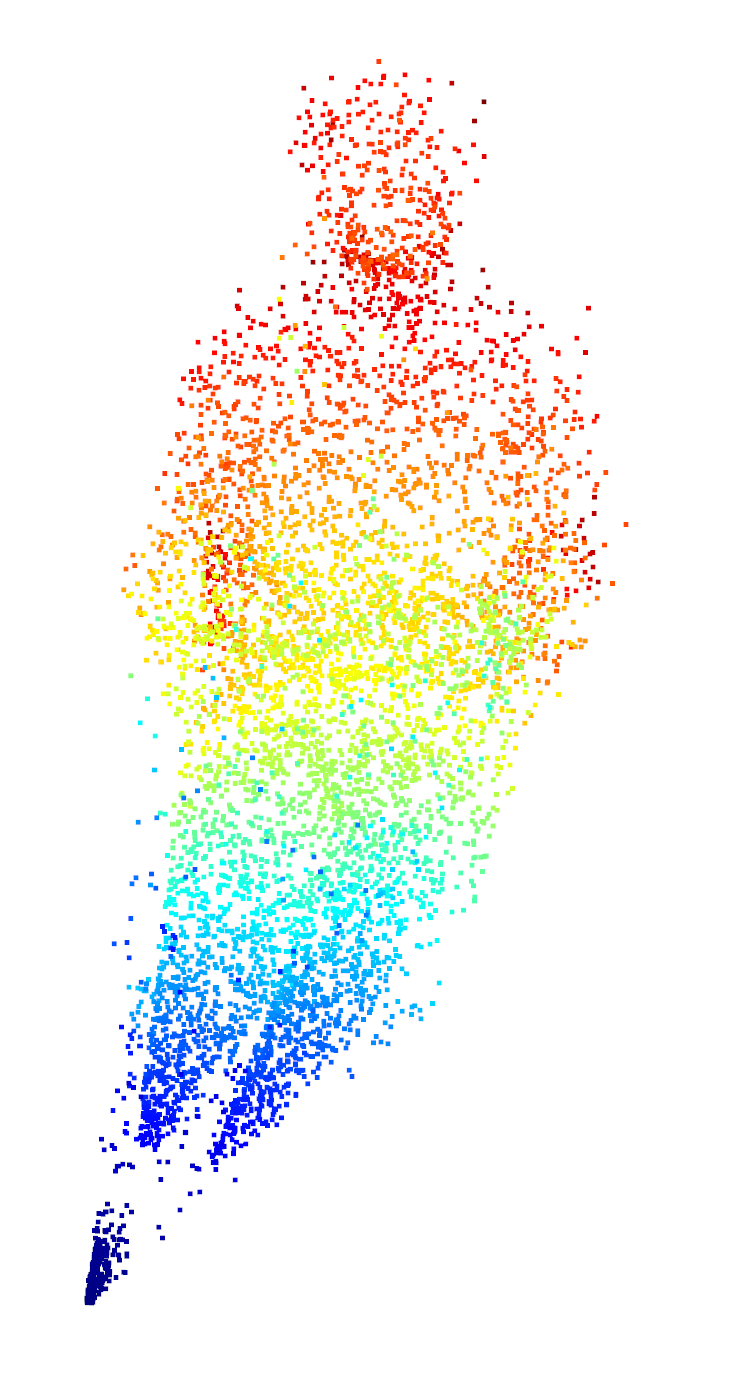}
		\subcaption{}
	\end{subfigure}
	\begin{subfigure}{0.1\textwidth}
		\includegraphics[height=60px, width=50px]{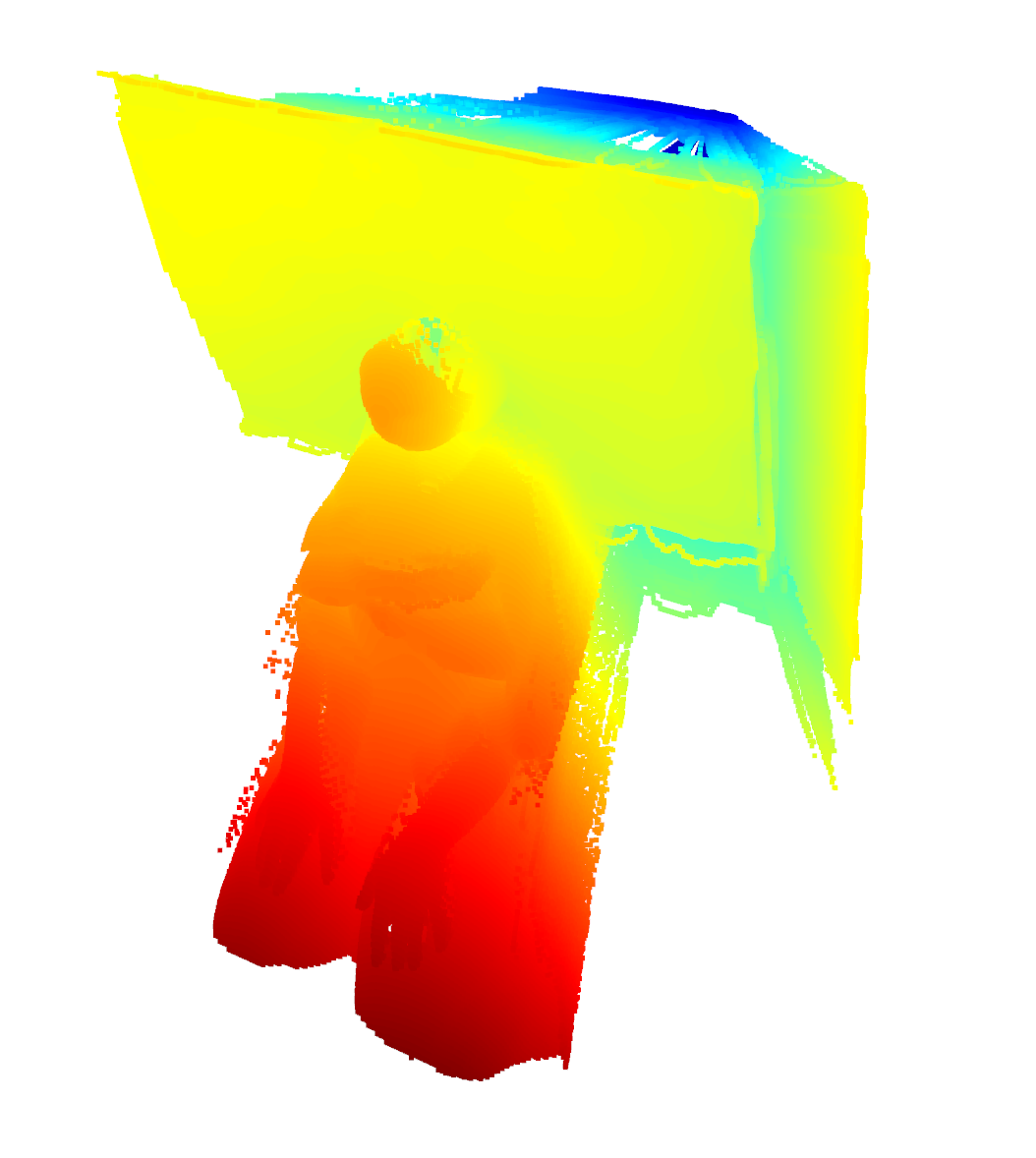}
		\subcaption{}
	\end{subfigure}
	\begin{subfigure}{0.1\textwidth}
		\includegraphics[height=60px, width=50px]{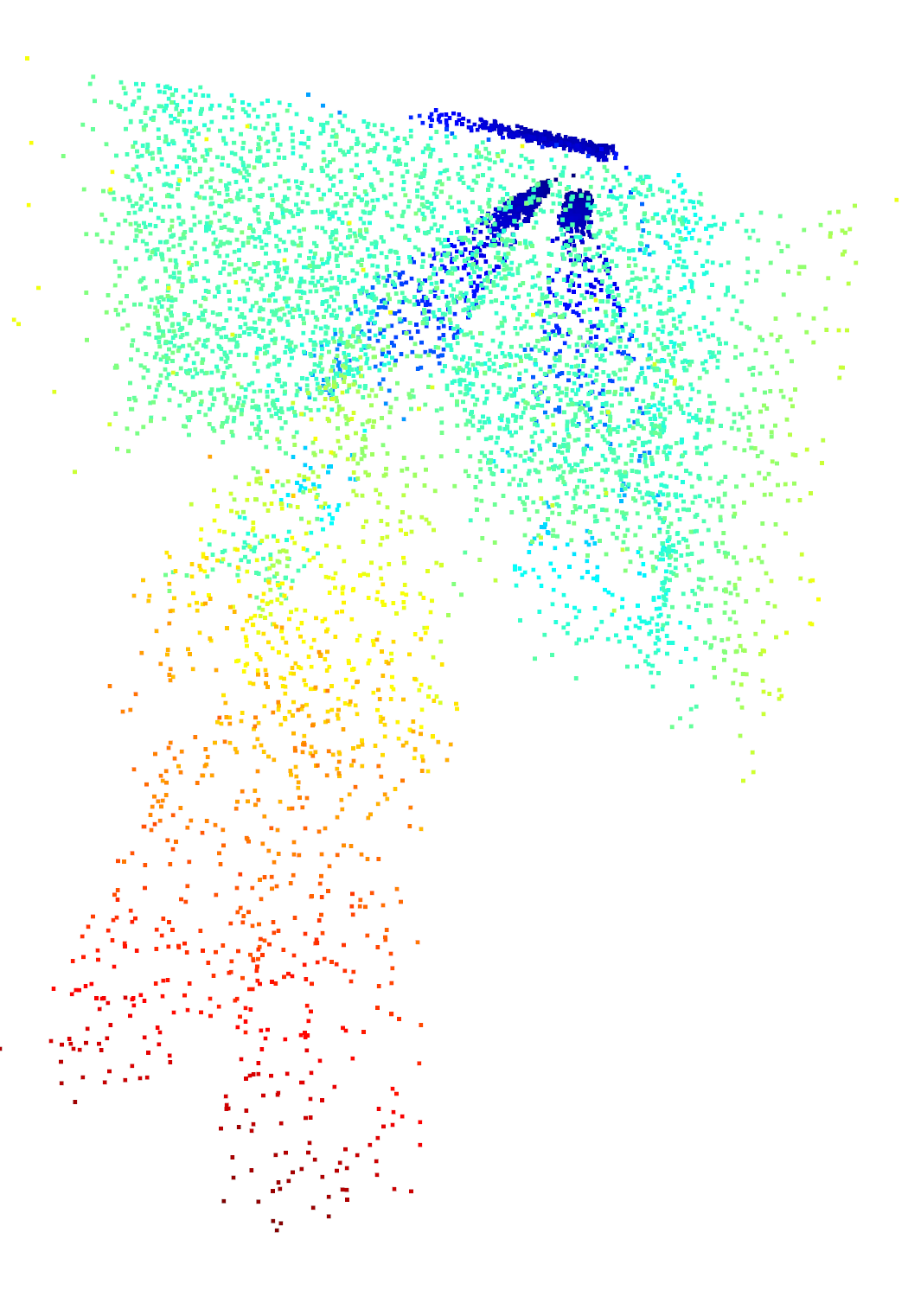}
		\subcaption{}
	\end{subfigure}
	\\
	\begin{subfigure}{0.1\textwidth}
		\includegraphics[height=60px, width=50px]{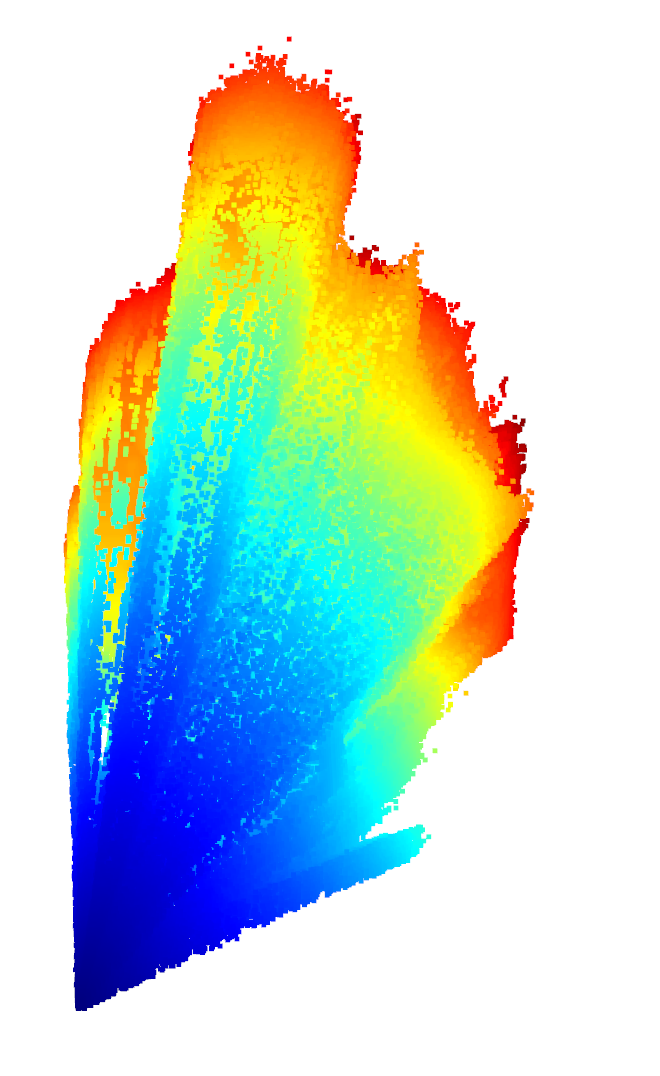}
		\subcaption{}
	\end{subfigure}
	\begin{subfigure}{0.1\textwidth}
		\includegraphics[height=60px, width=50px]{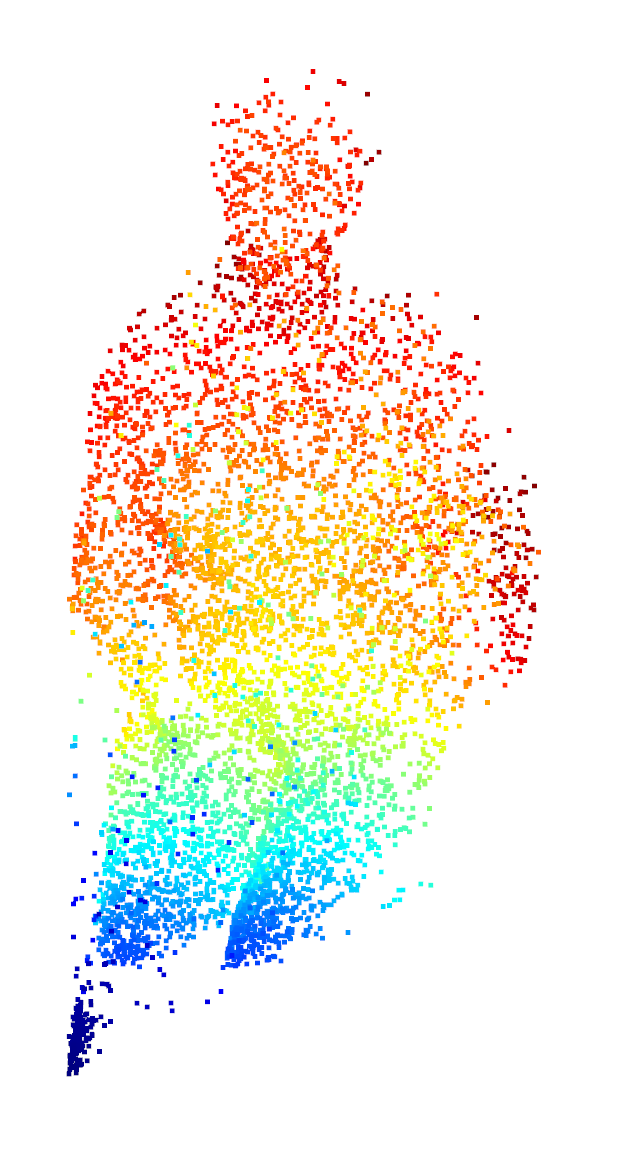}
		\subcaption{}
	\end{subfigure}
	\begin{subfigure}{0.1\textwidth}
		\includegraphics[height=60px, width=50px]{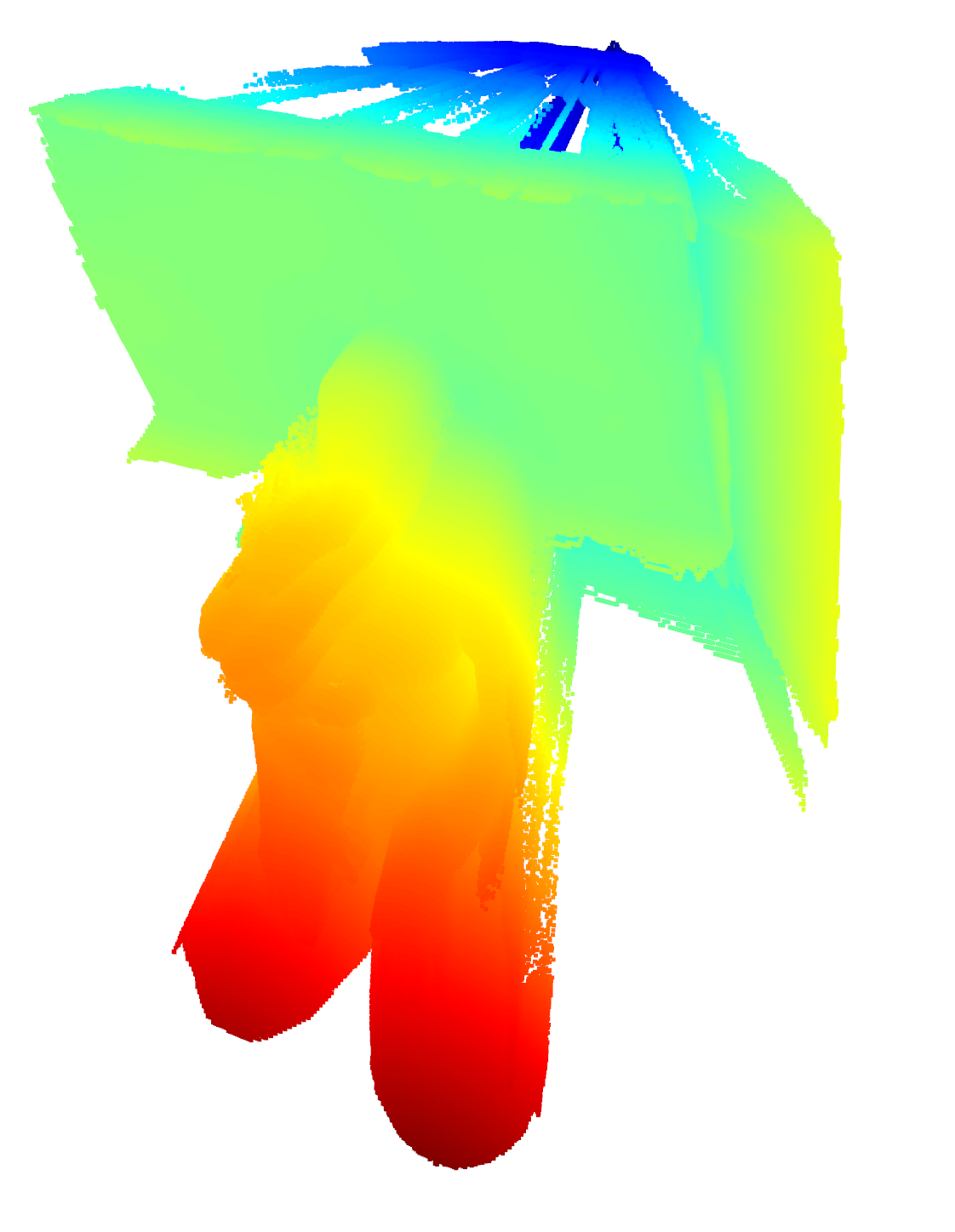}
		\subcaption{}
	\end{subfigure}
	\begin{subfigure}{0.1\textwidth}
		\includegraphics[height=60px, width=50px]{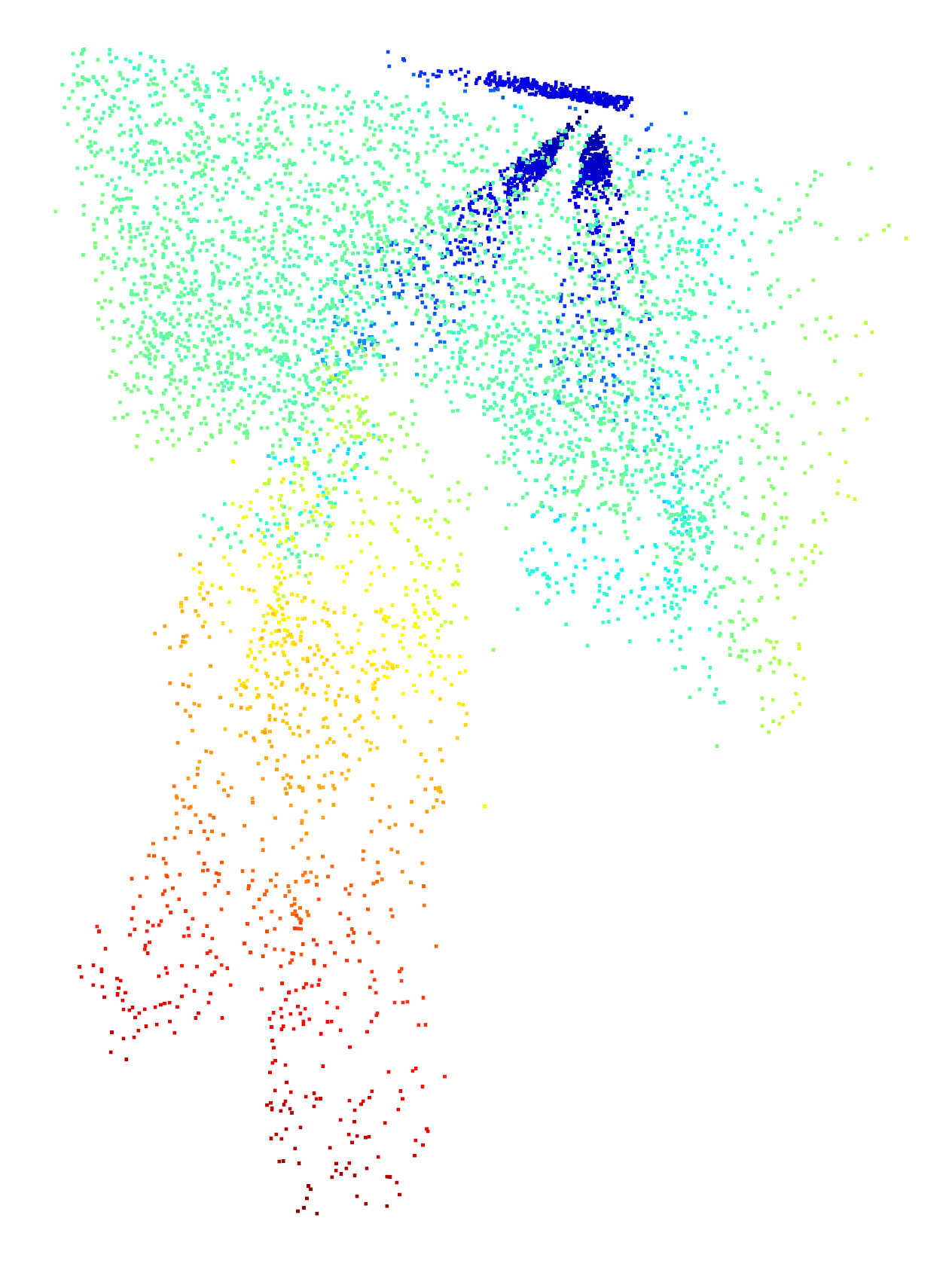}
		\subcaption{}
	\end{subfigure}
	\caption{\label{datasetC-examples2} Examples from the data created from Dataset-C. Top two rows: (a, g) Original depth, (b, h) Raw point cloud, (c, i) 512 sampled point cloud, (d, j) Synthetic depth, (e, k) Raw point cloud (f, l) 512 sampled point cloud. Bottom two row (m, q) Original depth PGM, (n, r) 6400 sampled PGM, (o, s) Synthetic PGM, (p, t) 6400 sampled PGM.}
\end{figure}

For all data in this dataset, depth-scale parameter was set to 500, depth-trunc was set to 1000 in Open3d. Figure \ref{datasetC-examples2} shows example data created from this dataset. Overall information about every data model of the datasets can be seen in Table \ref{table_dataset_info2}.

\begin{table}[htb!]
	\centering
	\caption{Data amount information for the datasets.}
	\begin{tabular}{lllll}
		\hline
		\textbf{Data} & \textbf{Train \& Validation} & \textbf{Test} & \textbf{Total} & \textbf{Network Input}\\ 
		\hline
		Dataset-A Frame (Original Depth) & 49417 & 16473 & 65890 & 32 x 512 x 3\\ 
		\hline
		Dataset-A Frame (Synthetic Depth) & 49330 & 16444 & 65774 & 32 x 512 x 3\\ 
		\hline
		Dataset-A PGM (Original Depth) & 955 & 319 & 1274 & 32 x 6400 x 3\\ 
		\hline
		Dataset-A PGM (Synthetic Depth) & 954 & 318 & 1272 & 32 x 6400 x 3\\ 
		\hline
		Dataset-B Frame (Original Depth) & 1443426 & 481143 & 1924569 & 32 x 512 x 3\\ 
		\hline
		Dataset-B Frame (Synthetic Depth) & 1443489 & 481164 & 1924653 & 32 x 512 x 3\\ 
		\hline
		Dataset-B PGM (Original Depth) & 56631 & 18878 & 75509 & 32 x 6400 x 3\\ 
		\hline
		Dataset-B PGM (Synthetic Depth) & 56636 & 18879 & 75515 & 32 x 6400 x 3\\ 
		\hline
		Dataset-B LSTM (Original Depth) & 55445 & 18482 & 73927 & 32 x 25 x 512\\ 
		\hline
		Dataset-B LSTM (Synthetic Depth) & 55417 & 18473 & 73890 & 32 x 25 x 512\\ 
		\hline
		Dataset-C Frame (Original Depth) & 1660229 & 553410 & 2213639 & 32 x 512 x 3\\ 
		\hline
		Dataset-C Frame (Synthetic Depth) & 1660230 & 553410 & 2213640 & 32 x 512 x 3\\ 
		\hline
		Dataset-C PGM (Original Depth) & 27226 & 9076 & 36302 & 32 x 6400 x 3\\ 
		\hline
		Dataset-C PGM (Synthetic Depth) & 22690 & 7564 & 30254 & 32 x 6400 x 3\\ 
		\hline
		Dataset-C LSTM (Original Depth) & 26936 & 8979 & 35915 & 32 x 30 x 512\\ 
		\hline
		Dataset-C LSTM (Synthetic Depth) & 28173 & 9391 & 37564 & 32 x 30 x 512\\ 
		\hline
	\end{tabular}
	\label{table_dataset_info2}
\end{table}

\subsection{Training}

Architecture of the PointNet used in the models is taken from the original Keras model \cite{pointnet-keras} with customized last layers. In frame based models, original architecture is used except from the last layer, which is adjusted to the class amount. In PGM models, last two layers (256 dense, Dropout, 128 dense, Dropout) were modified to four layers (4096, Dropout, 2048, Dropout, 1024, Dropout, 512, Dropout). Dropout layers were used with the 0.3 coefficient. In LSTM models, after an LSTM layer of size 256, 2048 dense, Dropout, 1024 dense, Dropout layers were used. Dropout layers were used with the 0.2 coefficient. ReLU activation function was used with the dense layers in all models. These layers were determined by observing various training experiments. Adam optimization \cite{kingma2014adam} was used for all networks with 0.0001 rate. All data from the datasets first shuffled before separating for training/validation/testing. To train the networks, Python 3.10.11 version with the Tensorflow-Keras 2.10 library was used. Computer that was used to train the models had these specs; AMD Ryzen 5 5600 CPU, Nvidia GeForce RTX 5060 TI 16 GB, 128 GB 3600 MHz DDR4 RAM. Overall training time for all networks lasted approximately for 100 days. With occasional adjustments and breaks, training process took 5 months. 

\underline{\textbf{Dataset-A (Real-time ASL Fingerspelling):}} For the Dataset-A, models mentioned in the training section were used. PGM PointNet architecture used in the Dataset-B and Dataset-C was also used in this dataset, because it was observed to perform well. Epoch amount and duration per epoch is given in the Table \ref{table-epoch-times-datasetA}. Training plots of the networks are given in the Figures \ref{datasetA-training-1}-\ref{datasetA-training-4}. 

\begin{table*}[!htb]
	\centering
	\caption{Epoch information for the Dataset-A.}
	\begin{tabular}{lll}
		\hline
		\textbf{Network} & \textbf{Epoch amount} & \textbf{Duration per epoch} \\ 
		\hline
		PointNet Frame (Original Depth) & 50 epoch & \textasciitilde50 seconds \\ 
		\hline
		PointNet Frame (Synthetic Depth) & 50 epoch & \textasciitilde52 seconds \\
		\hline
		PointNet PGM (Original Depth) & 100 epoch & \textasciitilde8 seconds \\
		\hline
		PointNet PGM (Synthetic Depth) & 100 epoch & \textasciitilde8 seconds \\
		\hline
	\end{tabular}
	\label{table-epoch-times-datasetA}
\end{table*}

\begin{table*}[!htb]
	\begin{tabularx}{\textwidth}{XX}
	\hspace{0.5cm}\includegraphics[width=0.40\textwidth]{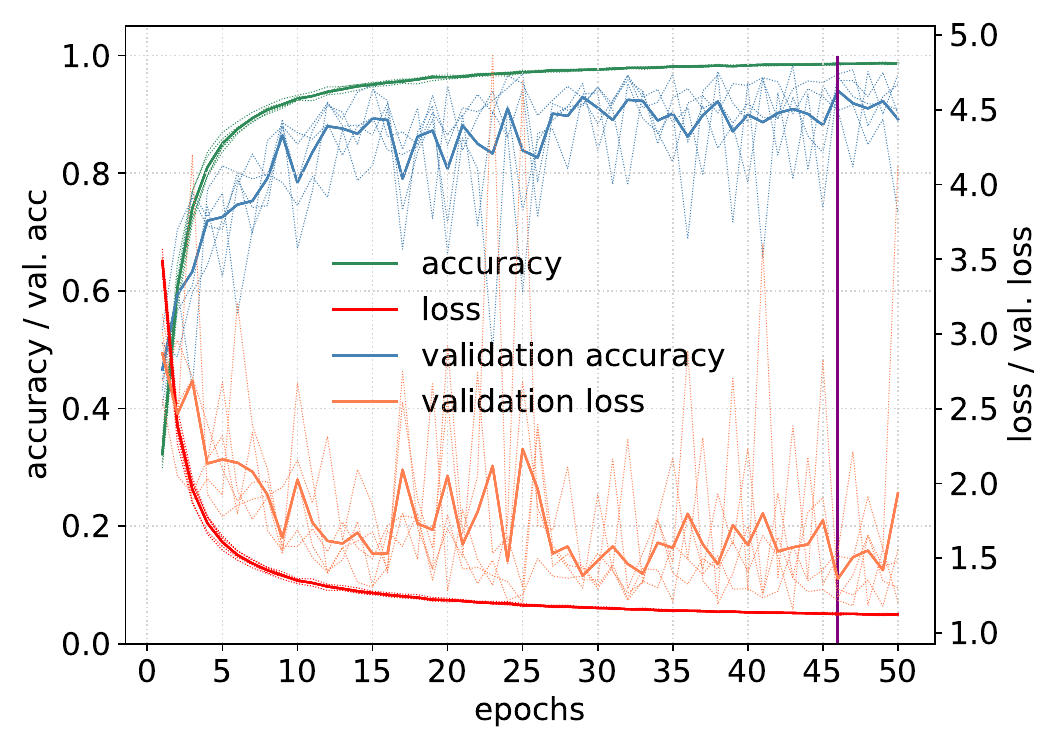} \captionof{figure}{Training for Dataset-A original depth frames.}\label{datasetA-training-1} &
	\hspace{0.5cm}\includegraphics[width=0.40\textwidth]{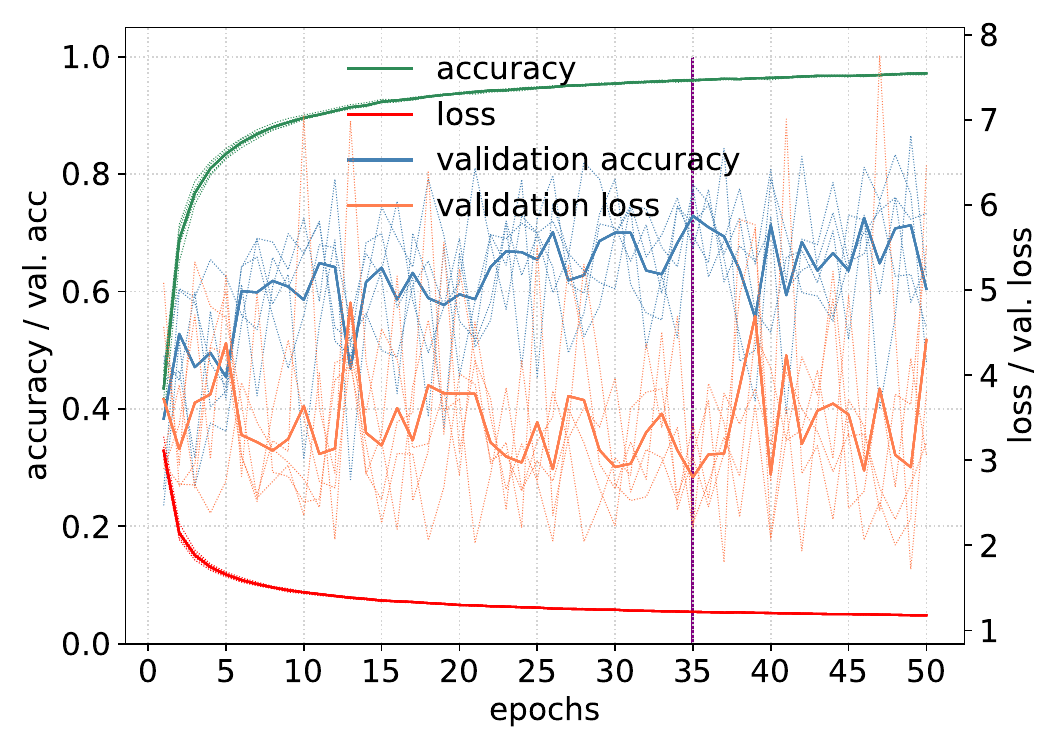} \captionof{figure}{Training for Dataset-A synthetic depth frames.}\label{datasetA-training-2}
	\\
	\hspace{0.5cm}\includegraphics[width=0.40\textwidth]{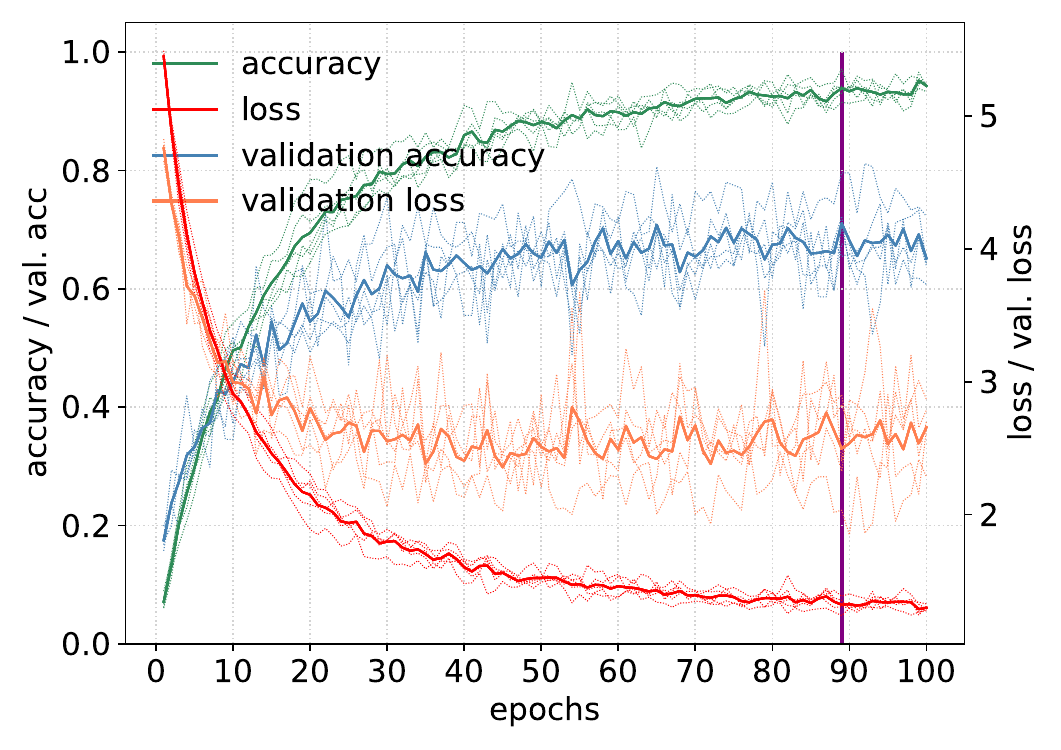} \captionof{figure}{Training for Dataset-A original PGM depth frames.}\label{datasetA-training-3} &
	\hspace{0.5cm}\includegraphics[width=0.40\textwidth]{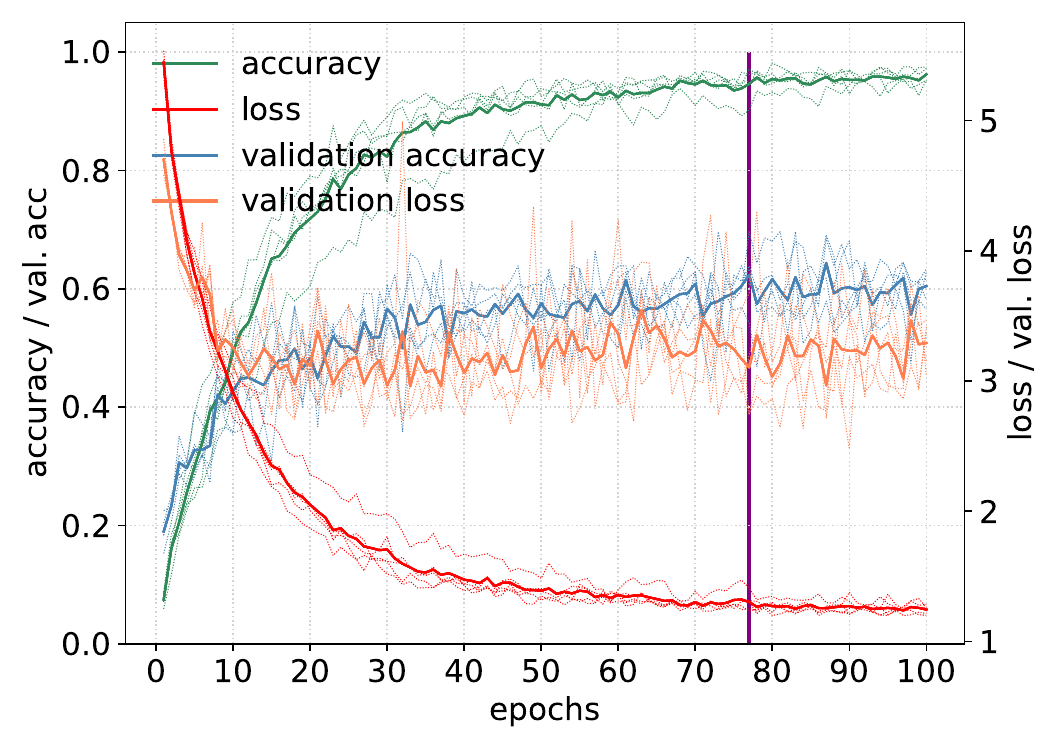}\captionof{figure}{Training for Dataset-A synthetic PGM depth frames.}\label{datasetA-training-4}
	\end{tabularx}
\end{table*}

\underline{\textbf{Dataset-B (KArSL):}} For the Dataset-B, frame based, PGM and LSTM models mentioned in the training section were used. Epoch amount and duration per epoch is given in the Table \ref{table-epoch-times-datasetB}. Training plots of the networks are given in the Figures \ref{datasetB-training-1}-\ref{datasetB-training-6}. 

\begin{table*}[!htb]
	\centering
	\caption{Epoch information for Dataset-B.}
	\begin{tabular}{lll}
		\hline
		\textbf{Network} & \textbf{Epoch amount} & \textbf{Duration per epoch}
		\\ 
		\hline
		PointNet Frame (Original Depth) & 240 epochs & \textasciitilde25 minutes
		\\ 
		\hline
		PointNet Frame (Synthetic Depth) & 50 epochs & \textasciitilde25 minutes
		\\
		\hline
		PointNet PGM (Original Depth) & 120 epochs & \textasciitilde8 minutes
		\\
		\hline
		PointNet PGM (Synthetic Depth) & 200 epochs & \textasciitilde8 minutes
		\\
		\hline
		LSTM (Original Depth) & 25 epochs & \textasciitilde15 seconds \\
		\hline
		LSTM (Synthetic Depth) & 25 epochs & \textasciitilde15 seconds \\
		\hline
	\end{tabular}
	\label{table-epoch-times-datasetB}
\end{table*}

\begin{table*}[!htb]
	\centering
	\begin{tabularx}{\textwidth}{XX}
		\hspace{0.5cm}\includegraphics[width=0.40\textwidth]{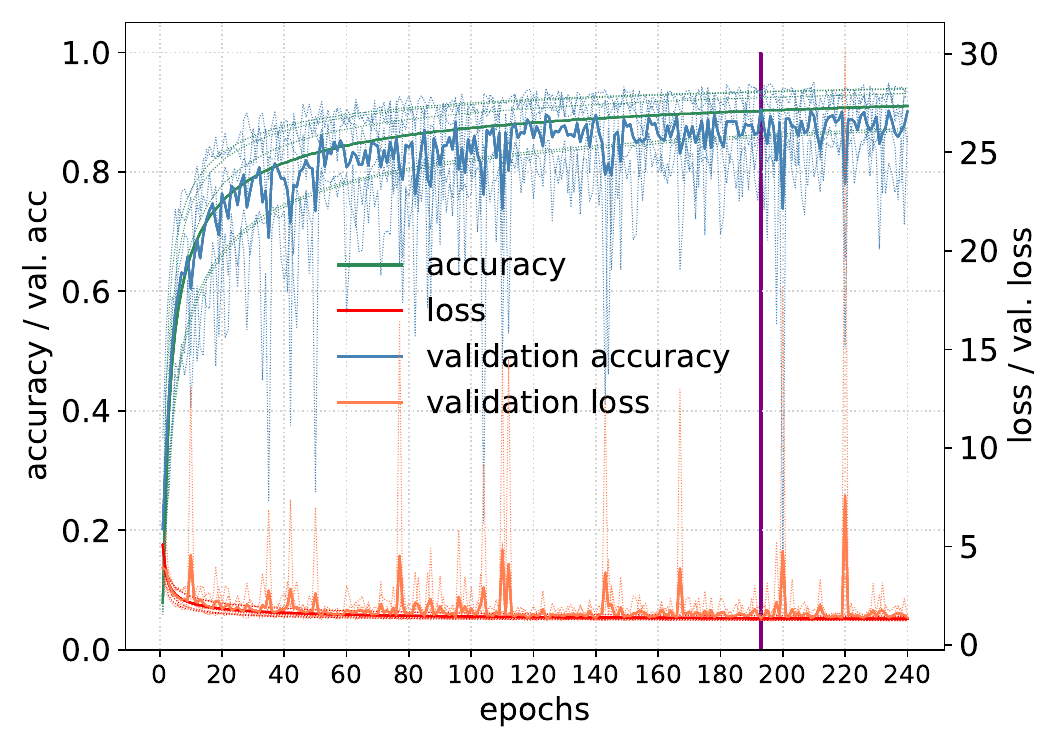} \captionof{figure}{Training for Dataset-B original depth frames.} \label{datasetB-training-1} & 
		\hspace{0.5cm}\includegraphics[width=0.40\textwidth]{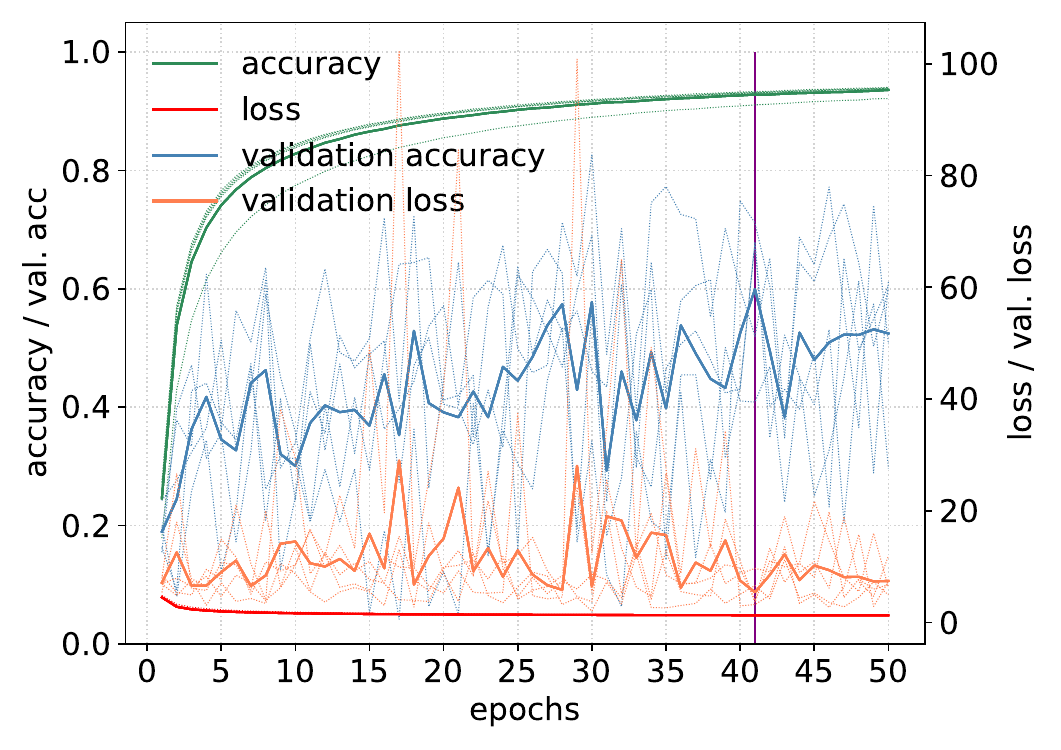}\captionof{figure}{Training for Dataset-b synthetic depth frames.} \label{datasetB-training-2}
		\\
		\hspace{0.5cm}\includegraphics[width=0.40\textwidth]{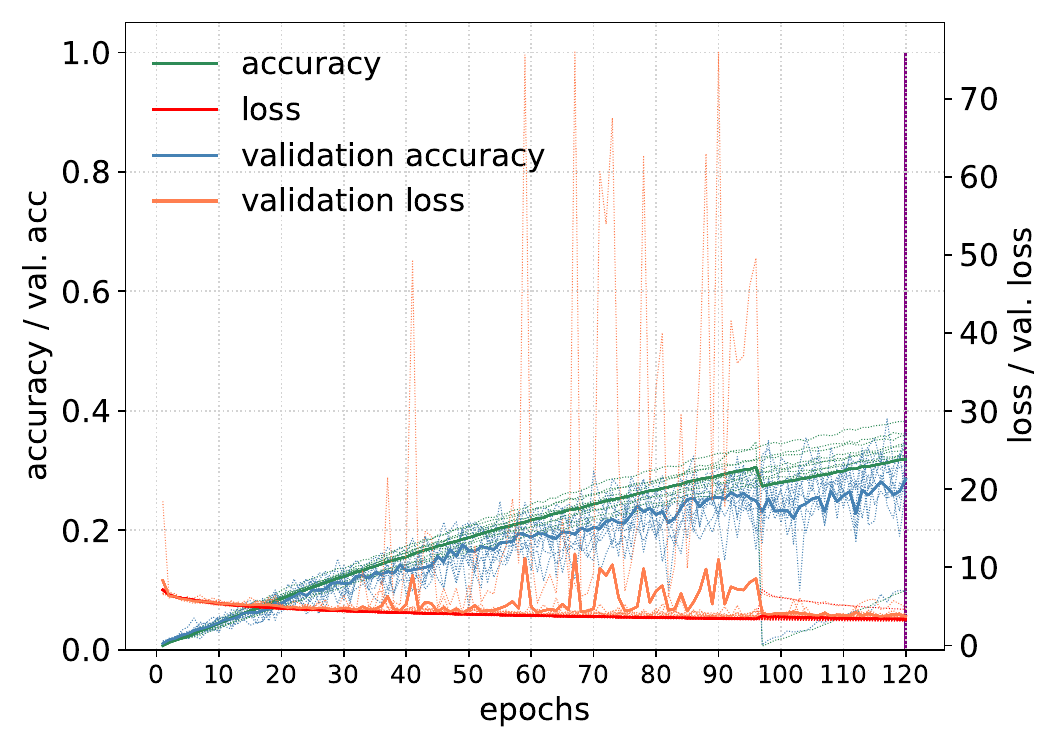}\captionof{figure}{Training for Dataset-B original depth frames PGM.} \label{datasetB-training-3} &
		\hspace{0.5cm}\includegraphics[width=0.40\textwidth]{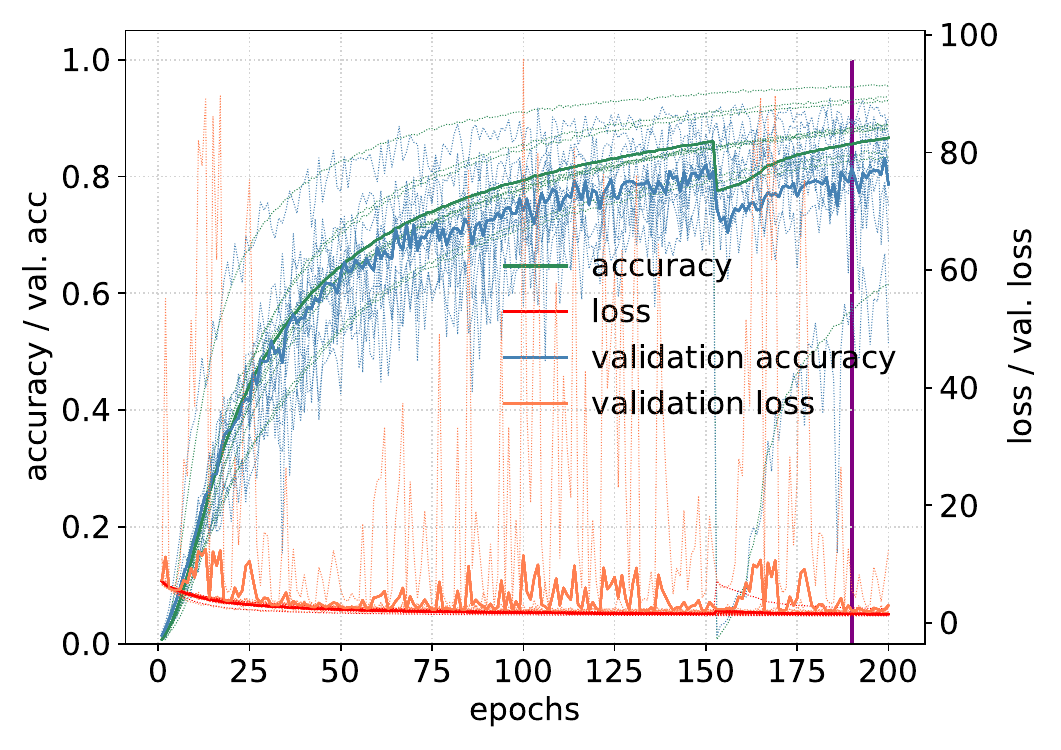}\captionof{figure}{Training for Dataset-B synthetic depth frames PGM.} \label{datasetB-training-4}
		\\
		\hspace{0.5cm}\includegraphics[width=0.40\textwidth]{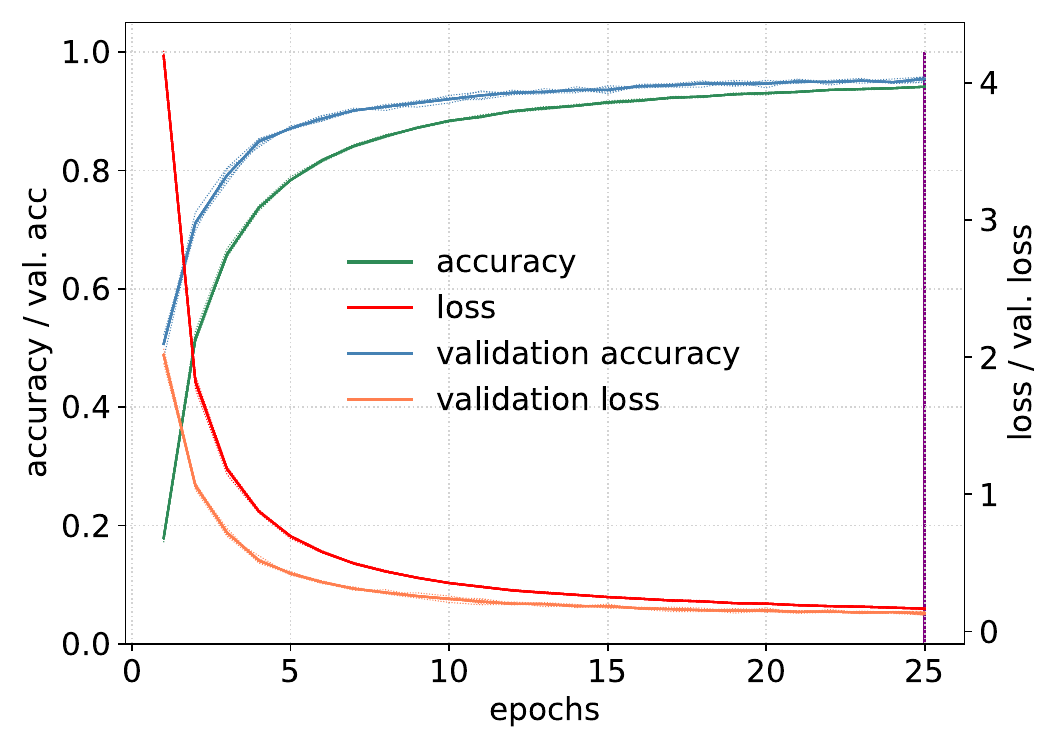}\captionof{figure}{Training for Dataset-B original depth frames LSTM.} \label{datasetB-training-5} &
		\hspace{0.5cm}\includegraphics[width=0.40\textwidth]{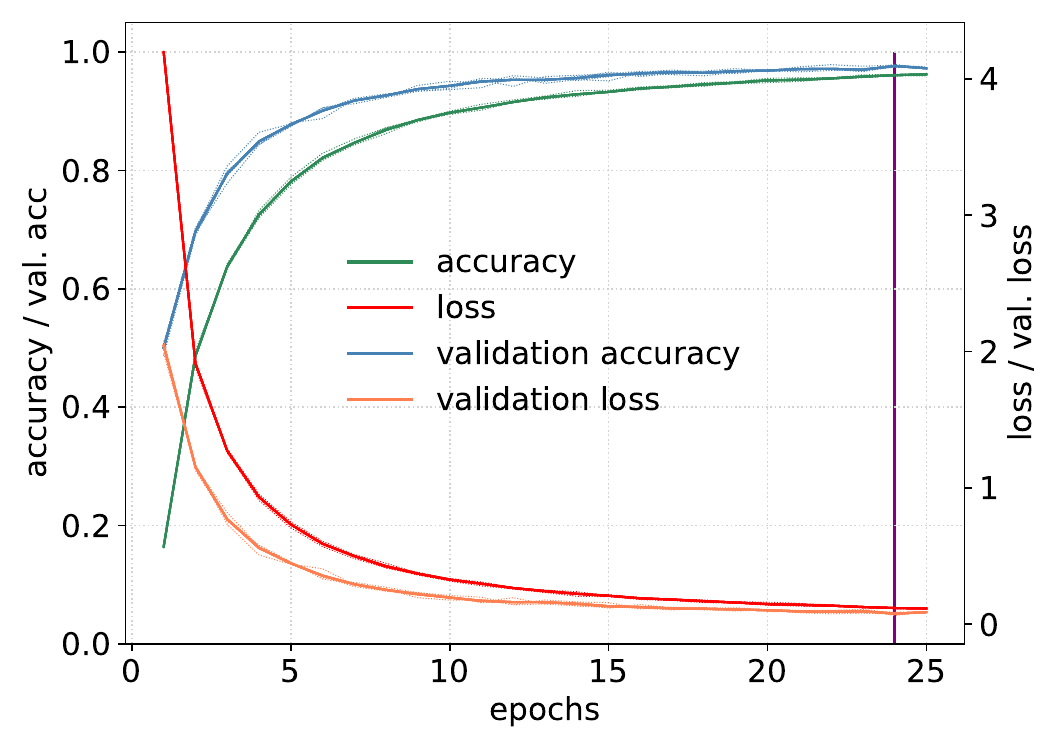}\captionof{figure}{Training for Dataset-B synthetic depth frames LSTM.} \label{datasetB-training-6}
	\end{tabularx}
\end{table*}

\underline{\textbf{Dataset-C (AUTSL):}} For the Dataset-C, same previous models mentioned in the training section is used. Epoch amount and duration per epoch is given in the Table \ref{table-epoch-times-datasetC}. Training plots of the networks are given in the Figures \ref{datasetC-training-1}-\ref{datasetC-training-5}. In some training figures, a few very high spikes of error were omitted to create a more readable figure. It can be seen from the training figures that, Dataset-C was more challenging for the networks to learn than the other datasets, with frequent oscillations in the validation loss.

\begin{table*}[!htb]
	\centering
	\caption{Epoch information for Dataset-C.}
	\begin{tabular}{lll}
		\hline
		\textbf{Network} & \textbf{Epoch amount} & \textbf{Duration per epoch}
		\\ 
		\hline
		PointNet Frame (Original Depth) & 200 epochs & \textasciitilde30 minutes
		\\ 
		\hline
		PointNet Frame (Synthetic Depth) & 250 epochs & \textasciitilde30 minutes
		\\
		\hline
		PointNet PGM (Original Depth) & 500 epochs & \textasciitilde3.5 minutes
		\\
		\hline
		PointNet PGM (Synthetic Depth) & 500 epochs & \textasciitilde3 minutes
		\\
		\hline
		LSTM (Original Depth) & 100 epochs & \textasciitilde7 seconds  \\
		\hline
		LSTM (Synthetic Depth) & 200 epochs & \textasciitilde7 seconds  \\
		\hline
	\end{tabular}
	\label{table-epoch-times-datasetC}
\end{table*}

\begin{table}[!htb]
	\centering
	\begin{tabularx}{\textwidth}{XX}
		\hspace{0.5cm}\includegraphics[width=0.40\textwidth]{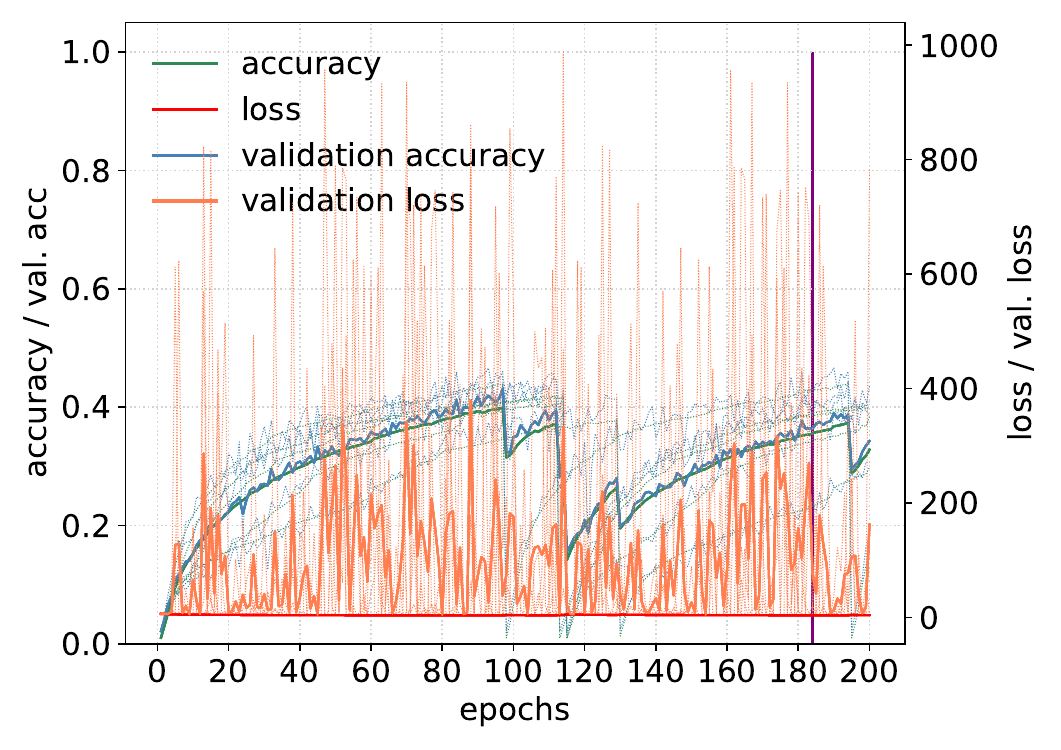}\captionof{figure}{Training for Dataset-C original depth frames.} \label{datasetC-training-1} &
		\hspace{0.5cm}\includegraphics[width=0.40\textwidth]{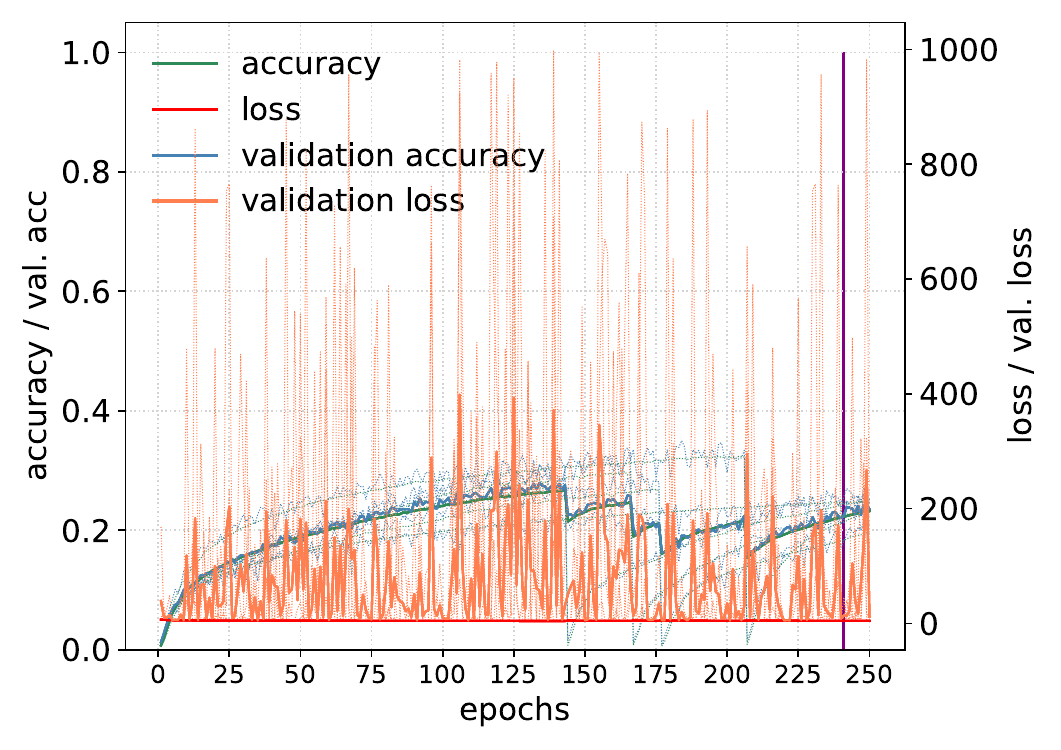}\captionof{figure}{Training for Dataset-C synthetic depth frames.} \label{datasetC-training-2}
		\\
		\hspace{0.5cm}\includegraphics[width=0.40\textwidth]{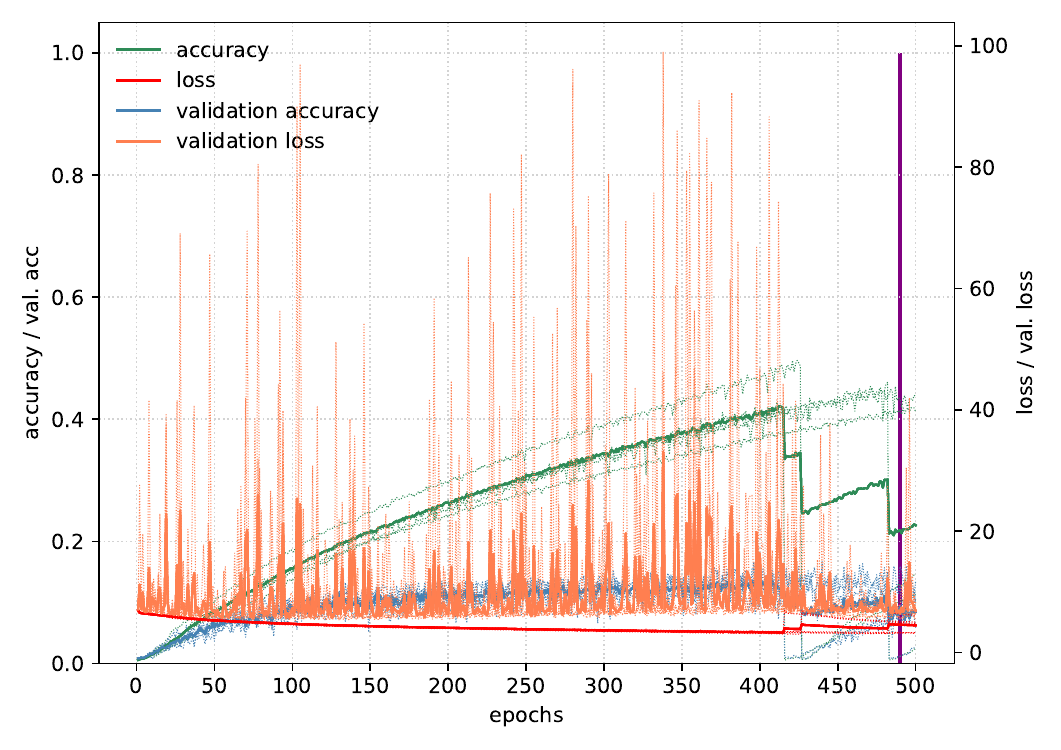}\captionof{figure}{Training for Dataset-C original depth frames PGM.} \label{datasetC-training-3} &
		\hspace{0.5cm}\includegraphics[width=0.40\textwidth]{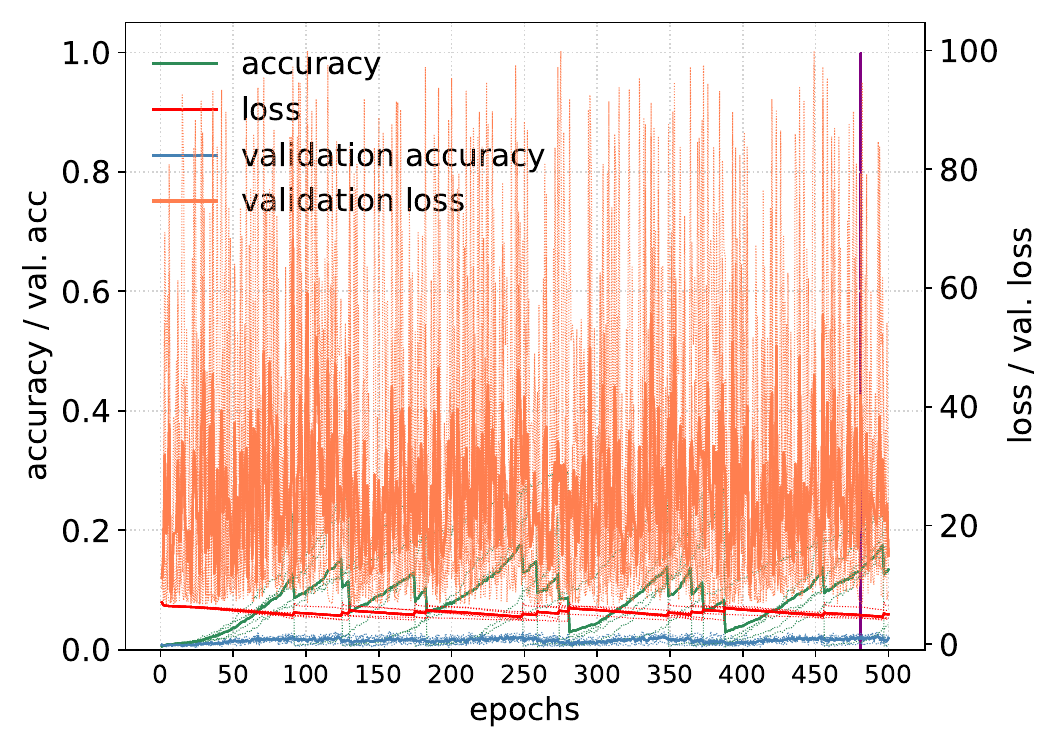}\captionof{figure}{Training for Dataset-C synthetic depth frames PGM.} \label{datasetC-training-4}
		\\
		\hspace{0.5cm}\includegraphics[width=0.40\textwidth]{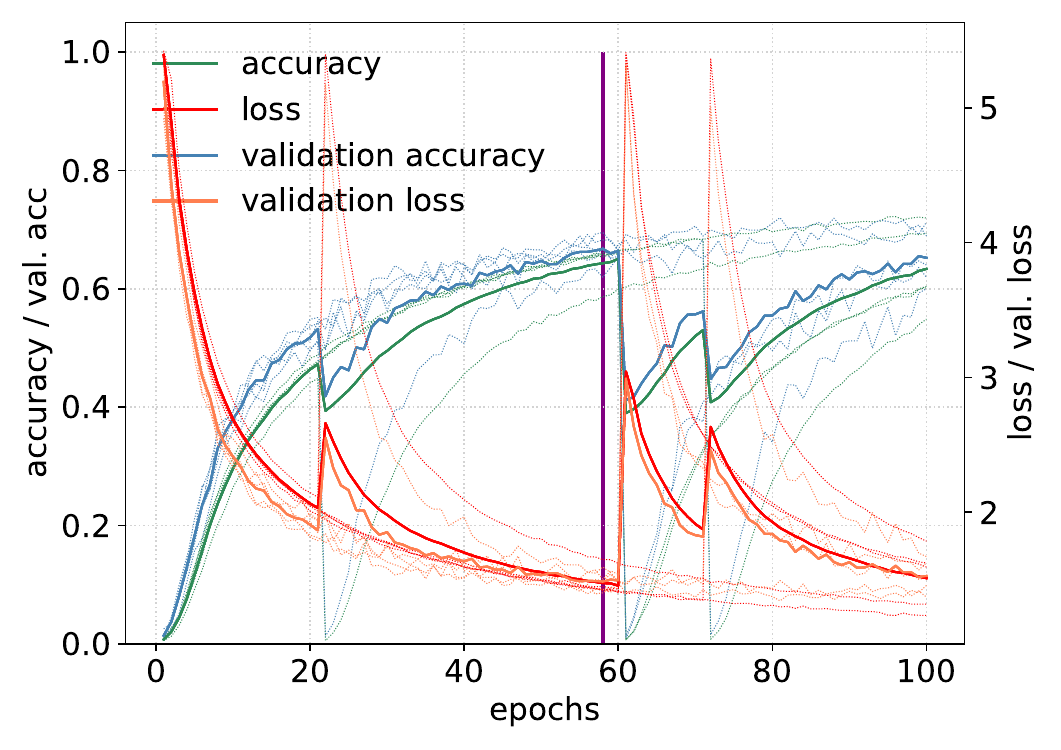}\captionof{figure}{Training for Dataset-C original depth frames LSTM.} \label{datasetC-training-5} &
		\hspace{0.5cm}\includegraphics[width=0.40\textwidth]{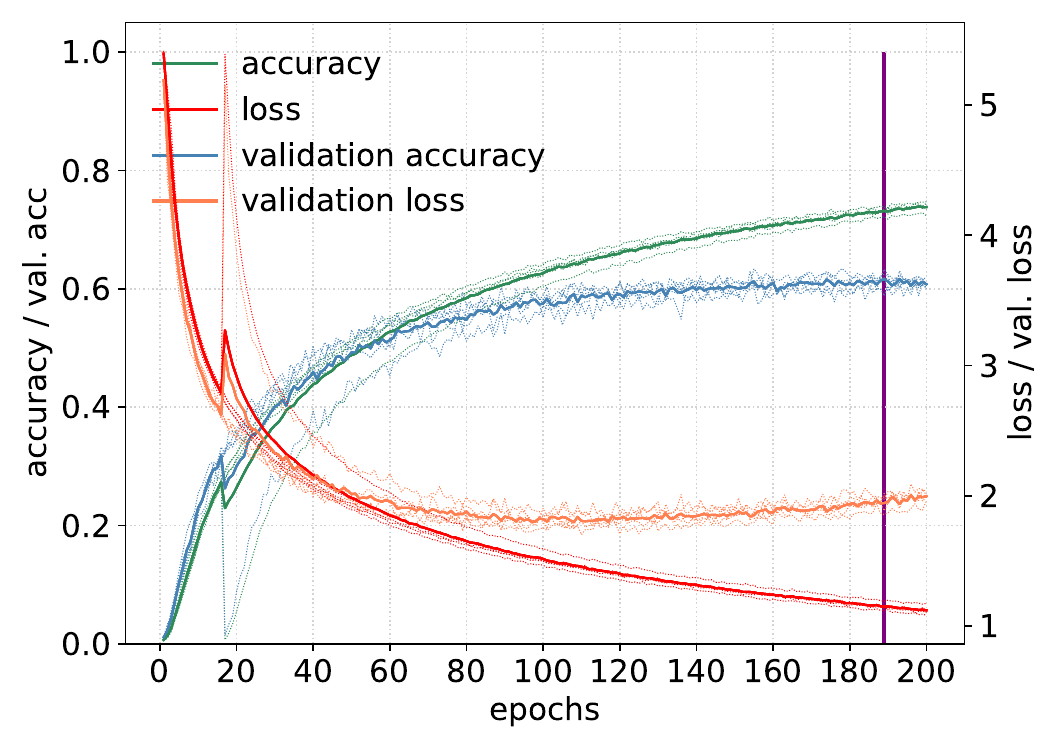}\captionof{figure}{Training for Dataset-C synthetic depth frames LSTM.} \label{datasetC-training-6}
	\end{tabularx}
\end{table}

\section{Results}

In this section, results of the all trained models are given with detailed explanation. During training, weights of the models were saved after each epoch for all models. Testing was done using the following method; for each model, all cross-validation models for all the epochs were tested using the test data. Test results of the same epoch from all cross-validation models were averaged and checked if the current epoch models achieved the best accuracy. The epoch with the most accuracy is marked with a bar on the training figures. Training figures show all cross-validation model performances with dashed line and average of all models with a solid line. Accuracy, recall, specifity, F1 scores are given for all models in Table \ref{table_overall_results}

\underline{\textbf{Dataset-A (Real-time ASL Fingerspelling):}} For the Dataset-A, best accuracy for the original depth frame based PointNet model was achieved with the 46. epoch out of 50 epochs. Best accuracy for the synthetic depth frame based PointNet was achieved with 35. epoch out of 50 epochs. For the original depth based PGM PointNet, best accuracy was achieved with 89. epoch out of 100 epochs. For the synthetic based PGM, best accuracy was achieved with 77. epoch out of 100 epochs. Confusion matrices for the Dataset-A are given in Figures \ref{datasetA-conf-1}-\ref{datasetA-conf-4}. Best performig model was the original depth frame based PointNet model with \%94.16 accuracy.

\begin{table*}[!htb]
	\centering
	\begin{tabularx}{\textwidth}{XX}
		\includegraphics[width=0.50\textwidth]{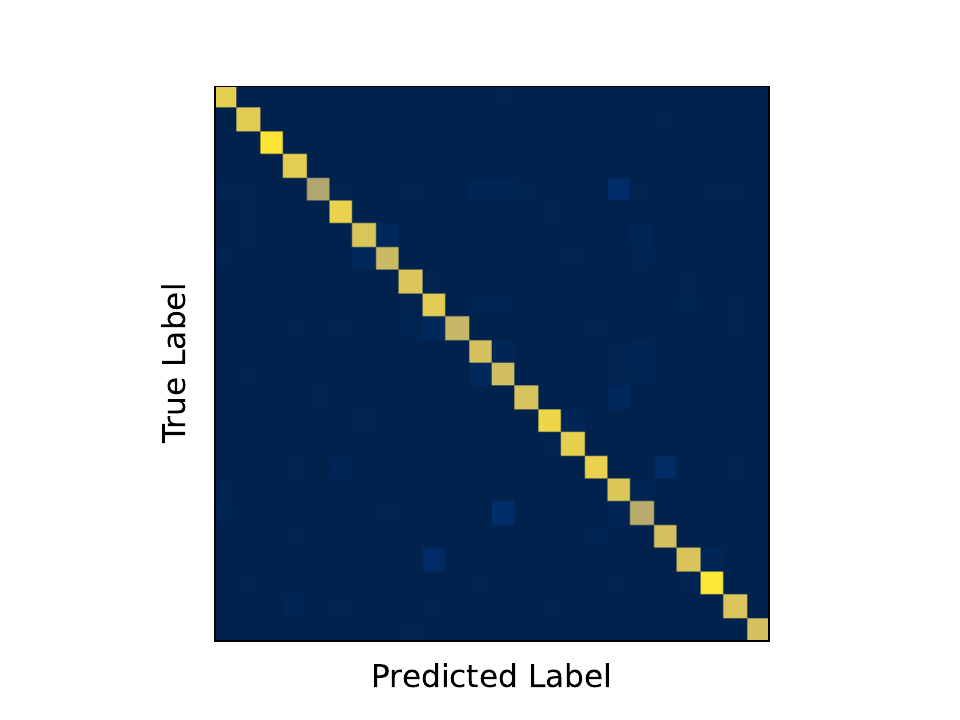}
		\captionof{figure}{Confusion matrix for Dataset-A original depth frames.}\label{datasetA-conf-1} & 
		\includegraphics[width=0.50\textwidth]{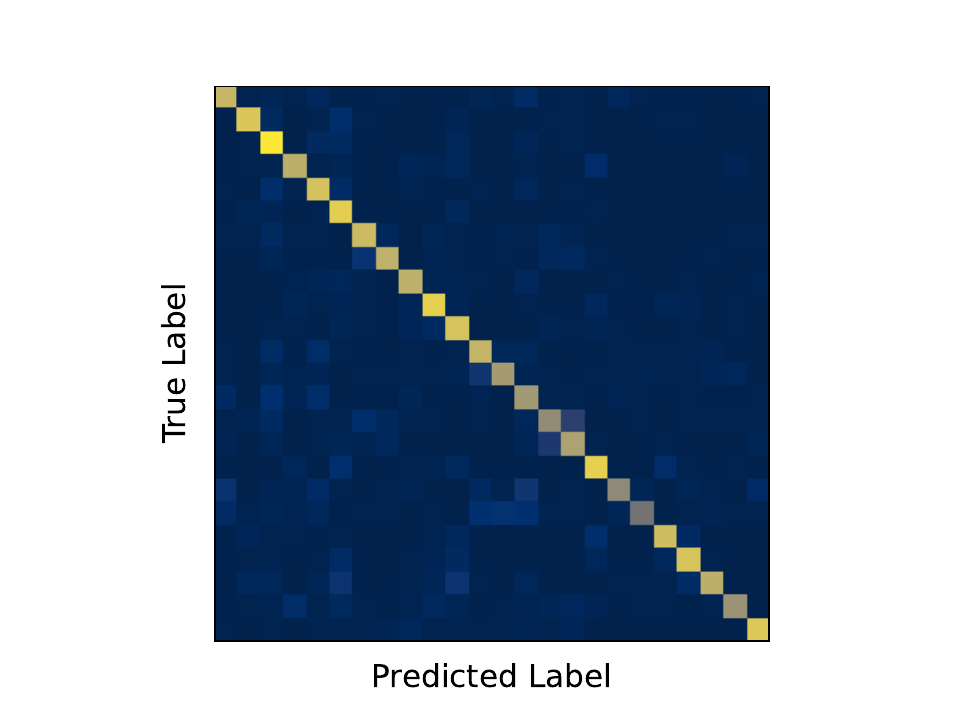} 
		\captionof{figure}{Confusion matrix for Dataset-A synthetic depth frames.}\label{datasetA-conf-2}
		\\
		\includegraphics[width=0.50\textwidth]{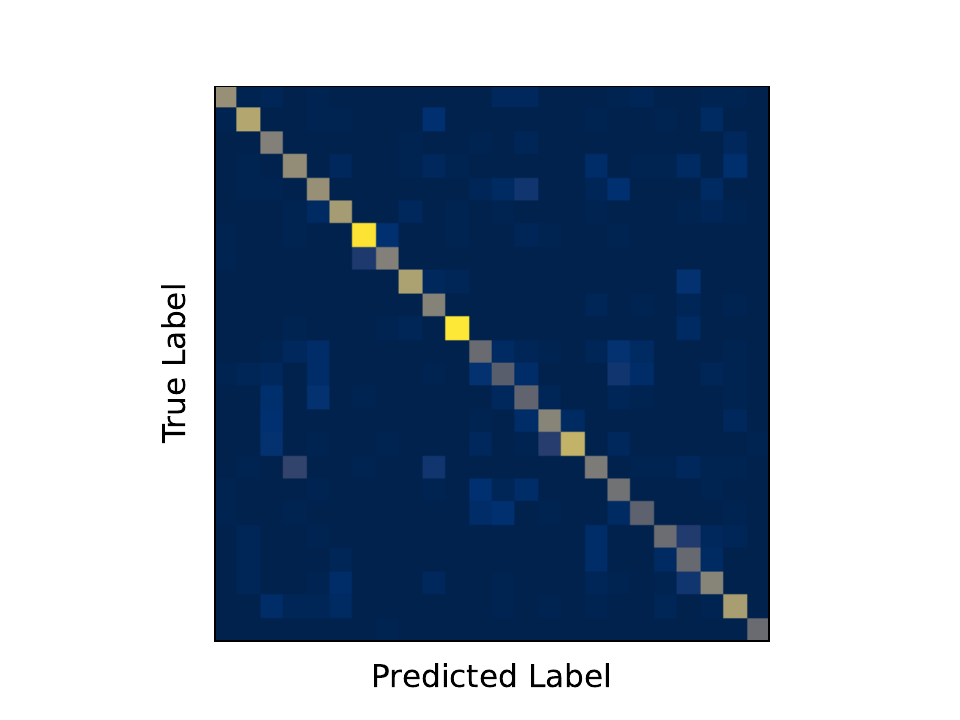}\captionof{figure}{Confusion matrix for Dataset-A original PGM depth frames.} \label{datasetA-conf-3} &
		\includegraphics[width=0.50\textwidth]{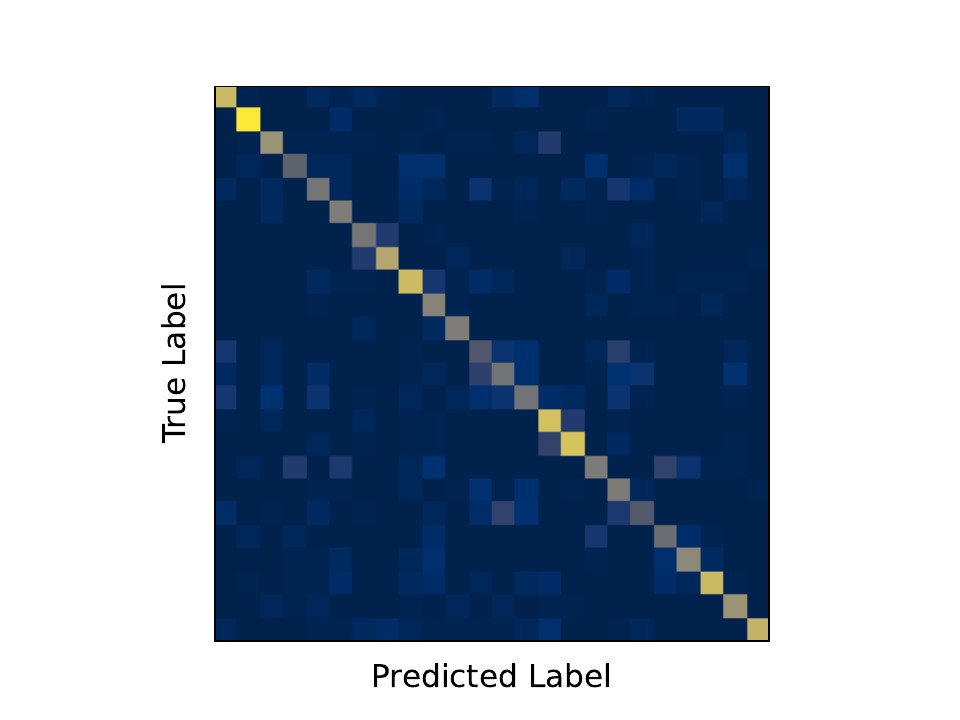} 
		\captionof{figure}{Confusion matrix for Dataset-A synthetic PGM depth frames.}\label{datasetA-conf-4}
		\\
	\end{tabularx}
\end{table*}

\underline{\textbf{Dataset-B (KArSL):}} For the Dataset-B, best accuracy for the original depth frame based PointNet model was achieved with the 193. epoch out of 240 epochs. Best accuracy for the synthetic depth frame based PointNet was achieved with 41. epoch out of 50 epochs. For the original depth based PGM PointNet model, best accuracy was achieved with 120. epoch out of 120 epochs. For the synthetic based PGM PointNet model, best accuracy was achieved with 190. epoch out of 200 epochs. In original depth based LSTM networks, best accuracy was achieved with 25. epoch out of 25 epochs. In synthetic based LSTM networks, best accuracy was achieved with 24. epoch out of 25 epochs. Confusion matrices for the Dataset-B are given in Figures \ref{datasetB-conf-1}-\ref{datasetB-conf-6}. Best performig model was the synthetic depth based LSTM model with \%97.56 accuracy.

\begin{table*}[!htb]
	\centering
	\begin{tabularx}{\textwidth}{XX}
		\includegraphics[width=0.50\textwidth]{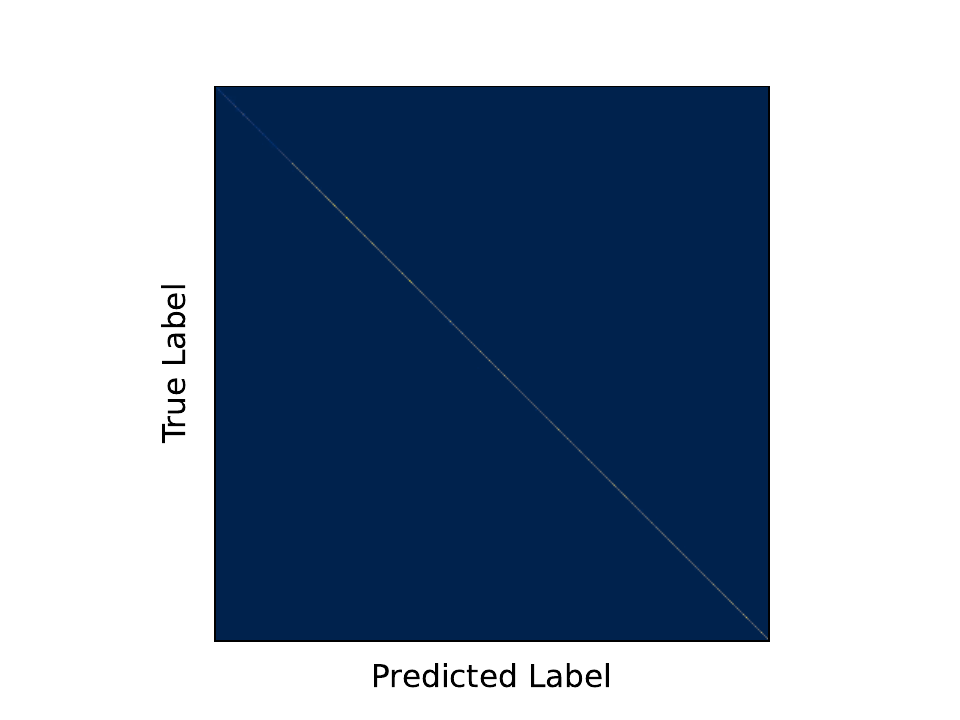} \captionof{figure}{Confusion matrix for Dataset-B original depth frames.} \label{datasetB-conf-1} & 
		\includegraphics[width=0.50\textwidth]{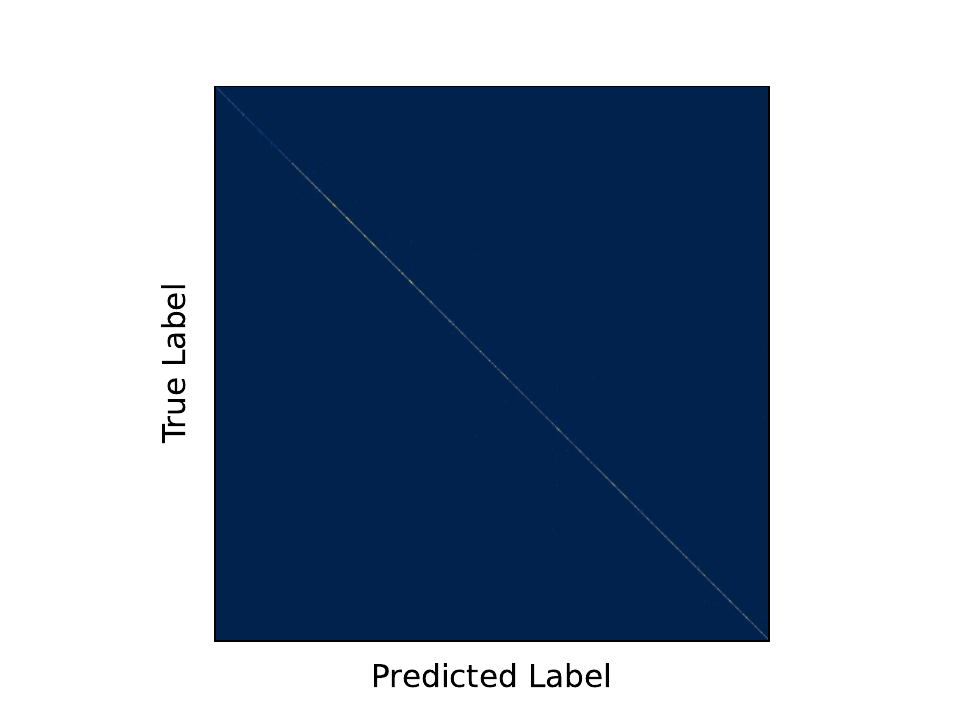}\captionof{figure}{Confusion matrix for Dataset-B synthetic depth frames.} \label{datasetB-conf-2}
		\\
		\includegraphics[width=0.50\textwidth]{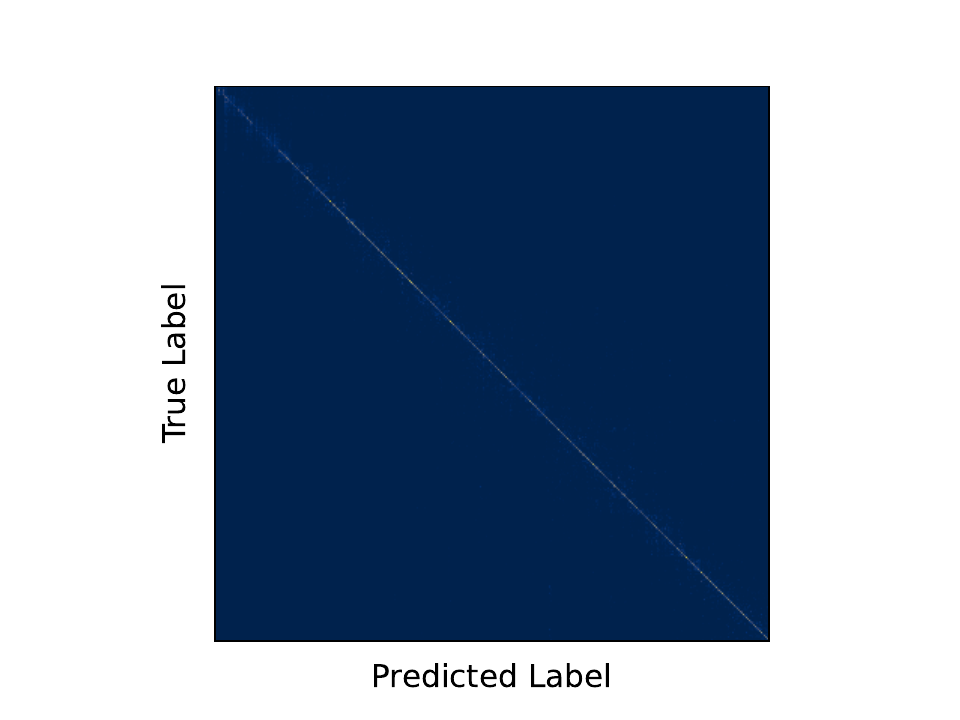}\captionof{figure}{Confusion matrix for Dataset-B original depth frames PGM.} \label{datasetB-conf-3} &
		\includegraphics[width=0.50\textwidth]{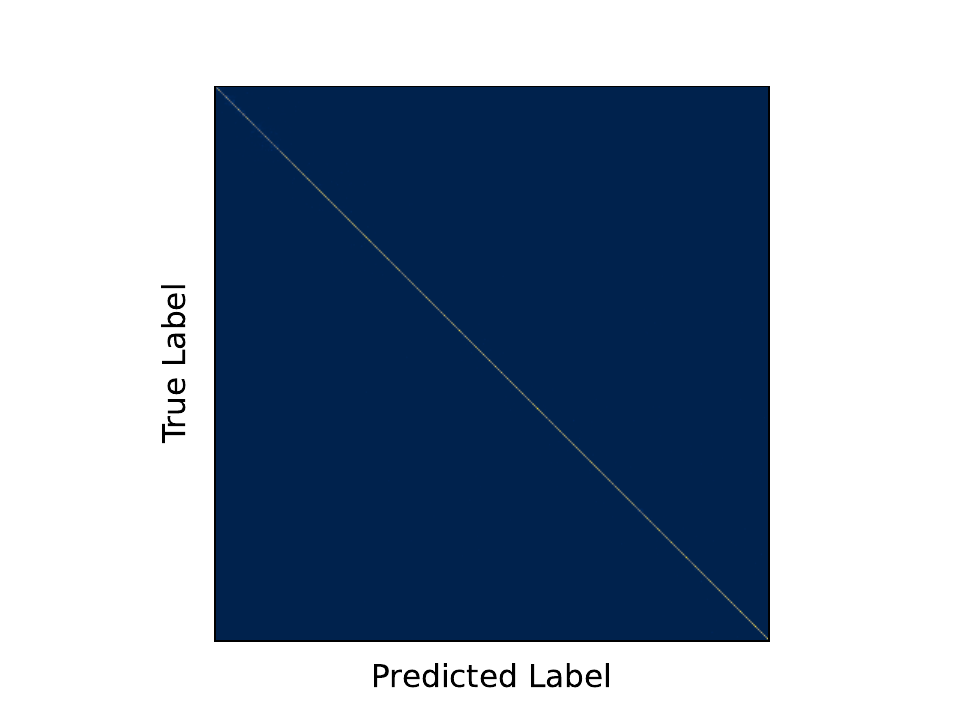}\captionof{figure}{Confusion matrix for Dataset-B synthetic depth frames PGM.} \label{datasetB-conf-4}
		\\
		\includegraphics[width=0.50\textwidth]{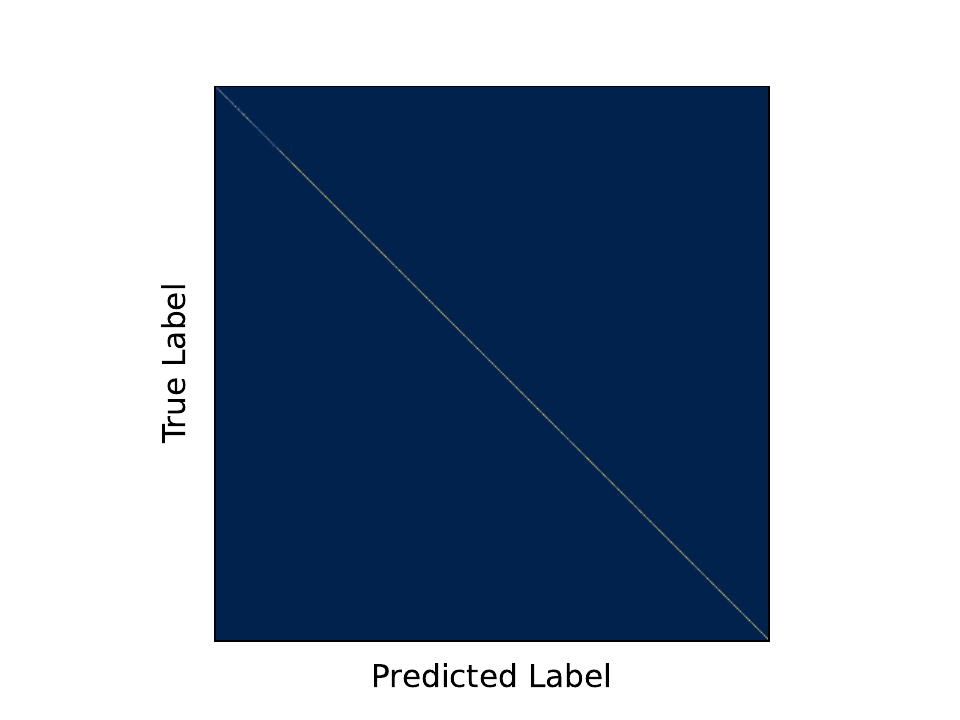}\captionof{figure}{Confusion matrix for Dataset-B original depth frames LSTM.} \label{datasetB-conf-5} &
		\includegraphics[width=0.50\textwidth]{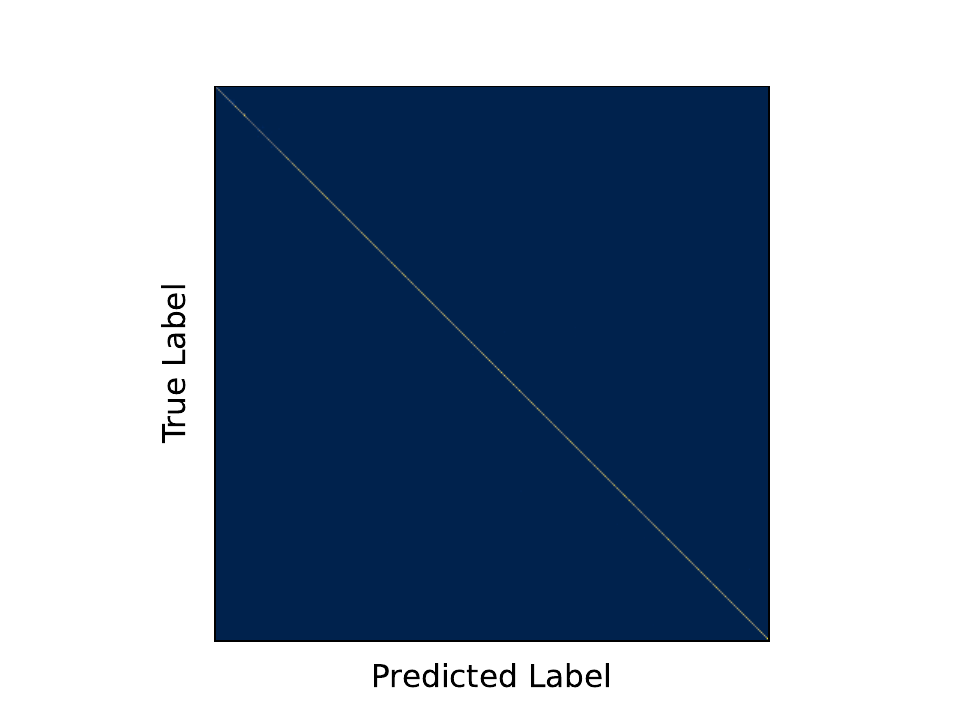}\captionof{figure}{Confusion matrix for Dataset-B synthetic depth frames LSTM.} \label{datasetB-conf-6}
	\end{tabularx}
\end{table*}

\underline{\textbf{Dataset-C (AUTSL):}} For the Dataset-C, best accuracy for the original depth frame based PointNet model was achieved with the 184. epoch out of 200 epochs. Best accuracy for the synthetic depth frame based PointNet was achieved with 241. epoch out of 250 epochs. For the original depth based PGM PointNet model, best accuracy was achieved with 490. epoch out of 500 epochs. For the synthetic based PGM PointNet model, best accuracy was achieved with 481. epoch out of 500 epochs. In original depth based LSTM networks, best accuracy was achieved with 58. epoch out of 100 epochs. In synthetic based LSTM networks, best accuracy was achieved with 189. epoch out of 200 epochs. Confusion matrices for the Dataset-C are given in Figures \ref{datasetC-conf-1}-\ref{datasetC-conf-5}. Synthetic depth based PGM model achieved an insignificant accuracy, for this reason confusion matrix of this model was not included. Best performig model was the original depth based LSTM model with \%68.43 accuracy.

\begin{table*}[!htb]
	\centering
	\caption{Overall results for all models.}
	\begin{tabular}{lllll}
		\hline
		\textbf{Model} & \textbf{Precision (Accuracy)} & \textbf{Recall (Sensitivity)} & \textbf{Specificity} & \textbf{F1} \\
		\hline
		Dataset-A Frame (Original Depth) & 94.16 &
		94.28 & 99.74 & 94.12 \\ 
		\hline
		Dataset-A Frame (Synthetic Depth) & 73.40 & 74.72 & 98.84 & 73.21 \\ 
		\hline
		Dataset-A PGM (Original Depth) & 70.59  & 70.97  & 98.72  & 70.28 \\ 
		\hline
		Dataset-A PGM (Synthetic Depth) & 61.25 & 61.49 & 98.31 & 61.05 \\ 
		\hline
		Dataset-B Frame (Original Depth) & 90.57  & 88.58 & 99.98 & 88.40 \\ 
		\hline
		Dataset-B Frame (Synthetic Depth) & 59.90 & 63.00 & 99.91 & 59.09 \\ 
		\hline
		Dataset-B PGM (Original Depth) & 32.11 & 31.89 & 99.86 & 31.39 \\ 
		\hline
		Dataset-B PGM (Synthetic Depth) & 86.81 & 86.67 & 99.97 & 86.62 \\ 
		\hline
		Dataset-B LSTM (Original Depth) & 95.19 & 94.74 & 99.99 & 94.63 \\ 
		\hline
		Dataset-B LSTM (Synthetic Depth) & 97.56  & 97.41 & 99.99 & 97.34 \\ 
		\hline
		Dataset-C Frame (Original Depth) & 44.28 & 50.04 & 99.75 & 45.88 \\ 
		\hline
		Dataset-C Frame (Synthetic Depth) & 29.23 & 34.16 & 99.68 & 30.39 \\ 
		\hline
		Dataset-C PGM (Original Depth) & 14.09 & 13.62 & 99.61 & 13.48 \\ 
		\hline
		Dataset-C LSTM (Original Depth) & 68.43 & 78.19 & 99.85 & 71.64 \\ 
		\hline
		Dataset-C LSTM (Synthetic Depth) & 61.61 & 65.72 & 99.82 & 62.68 \\ 
		\hline
	\end{tabular}
	\label{table_overall_results}
\end{table*}

\begin{table}[!htb]
	\centering
	\begin{tabularx}{\textwidth}{XX}
		\includegraphics[width=0.50\textwidth]{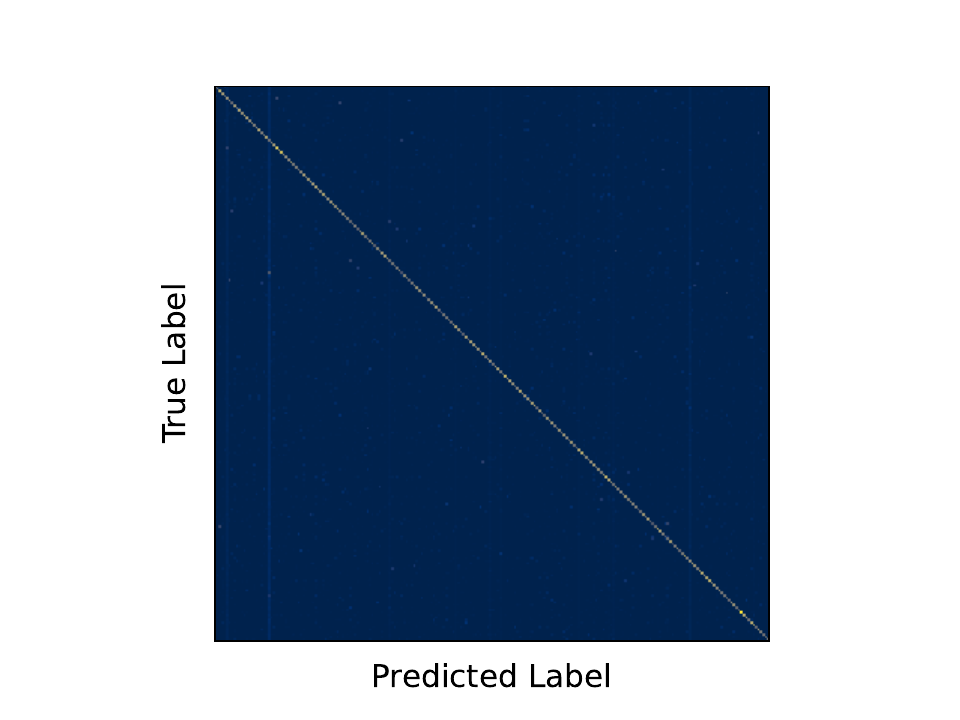} 
		\captionof{figure}{Confusion matrix for Dataset-C original depth frames.} \label{datasetC-conf-1} &
		\includegraphics[width=0.50\textwidth]{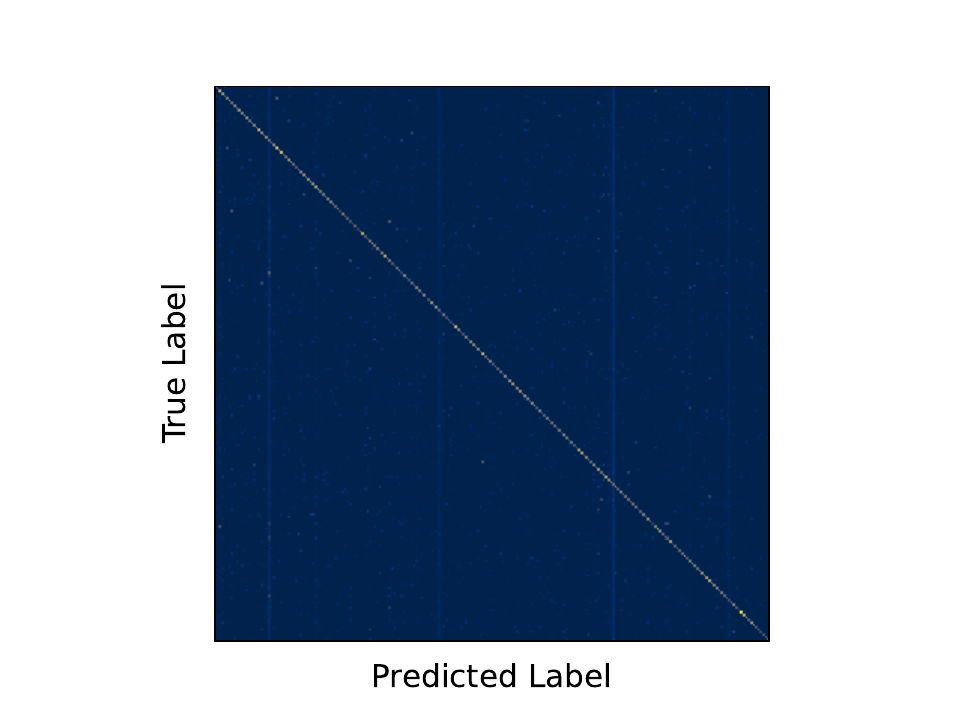} 
		\captionof{figure}{Confusion matrix for Dataset-C synthetic depth frames.} \label{datasetC-conf-2}
		\\
		\includegraphics[width=0.50\textwidth]{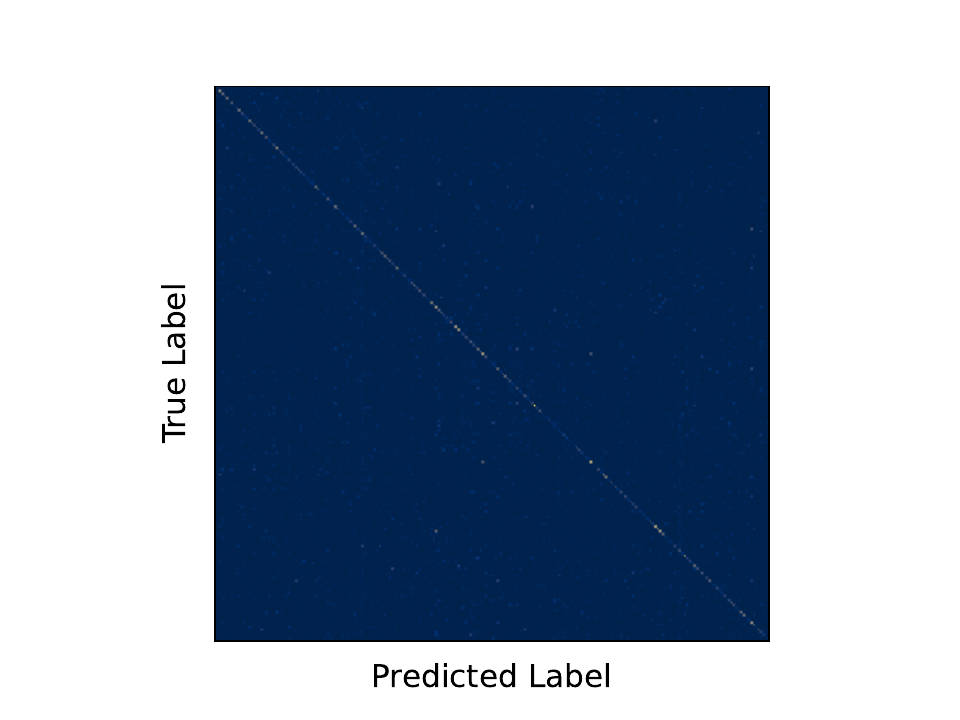}
		\captionof{figure}{Confusion matrix for Dataset-C original depth frames PGM.} \label{datasetC-conf-3} & 		\includegraphics[width=0.50\textwidth]{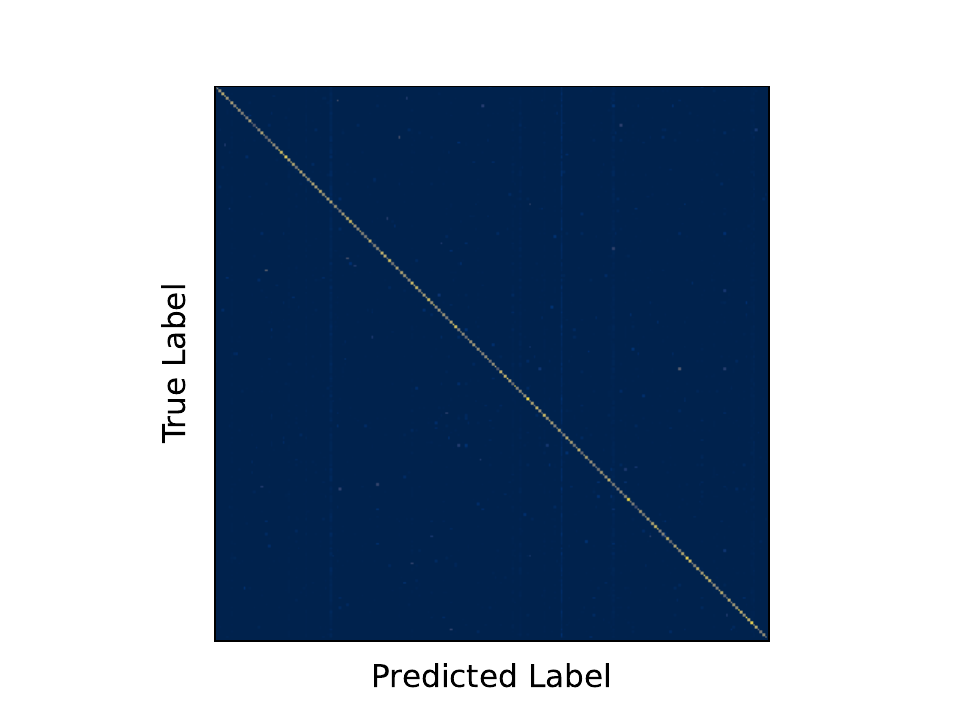} \captionof{figure}{Confusion matrix for Dataset-C original depth frames LSTM.} \label{datasetC-conf-4} 
		\\
		\multicolumn{2}{p{\linewidth}}{\centering \includegraphics[width=0.50\textwidth]{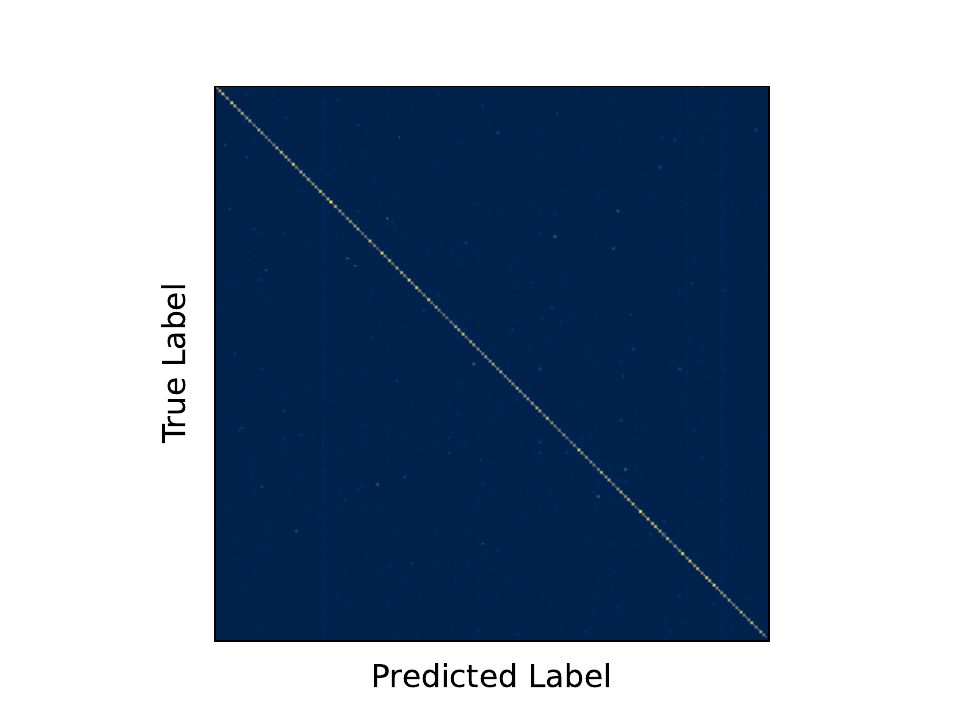}\captionof{figure}{Confusion matrix for Dataset-C synthetic depth frames LSTM.} \label{datasetC-conf-5}} \\
	\end{tabularx}
\end{table}

\section{Discussion}

In this work, synthetic and original depth image based point clouds were examined using different data models with PointNet networks. Three SLR datasets (Real-time ASL Fingerspelling, KArSL, AUTSL) were selected and results were presented. When training the models, raw data was used without arm/hand separation or processing.

In almost all neural network models, synthetic depth based data performed less successful. However, this difference does not show that they weren't completely useless, on the contrary they were close to the original data and showed potential. Also, in two extraordinary cases; synthetic based PGM model of the Dataset-B performed much better than the original based and synthetic based LSTM model performed slightly better than the original based model. Original based PGM network achieved \%32.11 accuracy whereas synthetic based PGM network achieved \%86.81 accuracy, original based LSTM network achieved \%95.19 accuracy, synthetic based LSTM network achieved \%97.56 accuracy. In some situations, synthetic data may be creating a distinctive positive effect that doesn't exist in the original data.

For the future works, point cloud data for SLR can be preprocessed to include only arm/hand regions. New SLR datasets can be added to comparison and new synthetic depth image generating methods can also be added to test and understand the extraordinary success observed in Dataset-B. Pre-trained models and codes can be accessed using \cite{github} link.

\clearpage

\bibliographystyle{IEEEtran}
\bibliography{references}  
\nocite{*}






\end{document}